\documentclass[runningheads]{llncs}

\usepackage{eccv}

\usepackage{eccvabbrv}

\usepackage{graphicx}
\usepackage{booktabs}

\usepackage[accsupp]{axessibility}  

\usepackage[pagebackref,breaklinks,colorlinks,citecolor=eccvblue]{hyperref}

\usepackage{orcidlink}

\begin{document}

\title{I3DGS: Improve 3D Gaussian Splatting from Multiple Dimensions} 

\titlerunning{Abbreviated paper title}

\author{Jinwei Lin\inst{1}\orcidlink{0000-0003-0558-6699}}

\authorrunning{F.~Author et al.}

\institute{Monash University, Clayton Victoria, Australia \and
Monash University Compuse, Malaysia 
\email{Jinwei.Lin@monash.edu}\\
\url{https://lin-jinwei.github.io/}
}

\maketitle

\begin{abstract}

3D Gaussian Splatting is a novel method for 3D view synthesis, which can gain an implicit neural learning rendering result than the traditional neural rendering technology but keep the more high-definition fast rendering speed. But it is still difficult to achieve a fast enough efficiency on 3D Gaussian Splatting for the practical applications. To Address this issue, we propose the I3DS, a synthetic model performance improvement evaluation solution and experiments test. From multiple and important levels or dimensions of the original 3D Gaussian Splatting, we made more than two thousand various kinds of experiments to test how the selected different items and components can make an impact on the training efficiency of the 3D Gaussian Splatting model. In this paper, we will share abundant and meaningful experiences and methods about how to improve the training, performance and the impacts caused by different items of the model. A special but normal Integer compression in base 95 and a floating-point compression in base 94 with ASCII encoding and decoding mechanism is presented. Many real and effective experiments and test results or phenomena will be recorded. After a series of reasonable fine-tuning, I3DS can gain excellent performance improvements than the previous one. The project code is available as open source.

\end{abstract}

\section{Introduction}

3D Gaussian plating \cite {kerbl20233d} is a novel implicit technology for 3D view synthesis or 3D volume rendering. Compared with the technologies that are based on the traditional original Neural Rendering (NeRF) \cite{mildenhall2021nerf} and traditional pure explicit rendering and non-learning 3D reconstruction technologies, 3D Gaussian plating can achieve higher resolution rendering and maintain an excellent processing speed. For these various rendering techniques, we propose a more widely recognized definition and classification, which we call the methods that are based on traditional 3D reconstruction techniques and theories and do not gain the main representation factors by machine learning, as explicit 3D rendering methods. Correspondingly, we refer to the methods that are based on novel 3D reconstruction techniques and theories and gain the main representation factors by machine learning as implicit 3D rendering methods. 

Before the rise of this technology, in the research area of implicit 3D representation, the technologies that were designed based on the NeRF were the dominant technologies and  governed this area for about three years.  In this development and evolution, Mip-nerf 360 \cite{barron2022mip} has achieved excellent results in high-resolution 3D representation, and Instant-NGP \cite{muller2022instant} has achieved outstanding results in fast real-time  3D representation. To design a novel real-time rendering method that can combine  high-resolution features such as the Mip-nerf 360 model and fast rendering such as Instant-NGP, the researchers proposed a 3D Gaussian Splatting model \cite {kerbl20233d}. In general, this study achieved most of the preset goals and targets. However, there are still some shortcomings, such as the lack of a direct process for the raw input data, not achieving near-real time at the application level on rendering, such as the Instant-NGP model, and insufficient detail on hyper-parameter fine-tuning. For each implicit 3D representation technology, the rendering speed, or more specifically, the learning and training speed, is one of the most important factors for its practical application. Therefore, to improve the rendering speed or performance of the original 3D Gaussian Splatting, we proposed I3DS, which improves 3D Gaussian Splatting. 

The main idea of the original 3D Gaussian Splatting model \cite {kerbl20233d} is to use  3D Gaussian splitting to perform rasterization in implicit neural rendering. 3D Gaussian Splatting is a novel method that was proposed in the 3D Gaussian Splatting model \cite {kerbl20233d}; however, it is a traditional explicit 3D representation technique  that is used in color representation \cite{mueller1999splatting} and image processing \cite{neophytou2005gpu}. Therefore, the rasterization of color items in a 3D scene is the main function of 3D Gaussian Splatting technology. This is also why we select the color as our first analyzed dimension that is used to test and  improve the rendering speed of the original 3D Gaussian plating mode. More interesting and meaningful, we had another novel idea that the color may not dominate the localization in the training of the rendering model, which means some  special components such as color that can be neglected and not included in the learning and training process, but needs to be added back after the learning. We call this idea training added back. The meaningful result is that we really made the learning without the color and gained a great improvement in training and gained a light-blur basic 3D model shape without the color, but it was difficult to add the color back. Further details are presented in the following sections. The second way to test and analyze the function caused by color is to analyze the role played by color in the model design and representation learning mechanism. In the 3D Gaussian Splatting model \cite {kerbl20233d}, color is represented by a spherical harmonic function and spherical harmonic coefficient. Spherical harmonics can be used to represent and render special surfaces \cite{max1988spherical}  and lighting representations in 3D modeling or 3D scene rendering \cite{green2003spherical}. Hence, we attempted to test and update the spherical harmonic architecture in the original 3D Gaussian Splatting \cite {kerbl20233d}, and some of them achieved satisfactory results. Further analysis is presented below.

As an implicit neural rendering method, the process of training and learning is also a good dimension for improving the analysis and optimization. In the learning and model design of the original 3D Gaussian Splatting model \cite {kerbl20233d}, we changed and tested various updates, such as optimizing the composition parameters, changing the learning rate, hyper-parameter fine-tuning, and changing the methods that handle the 3D Gaussian points. Changing the design and architecture of the learning method of the original 3D Gaussian Splatting model \cite {kerbl20233d} can lead to more direct and prominent improvements in performance, which is also a major part of the experiments in this study. During this procedure, we also found many different and meaningful model features that contain significant research value for the original 3D Gaussian Splatting model. To improve the training speed of the original 3D Gaussian Splatting model, we proposed a special but normal integer compression in base 95 and a floating-point compression in base 94 with ASCII encoding and decoding mechanisms. This data compression encoding technique is used in the CUDA coding part of the original 3D Gaussian Splatting model. The main components of rasterization learning and forward and backward training are matrix data. We also conducted a test for matrix data compression; however, there were other obstacles when applying it in practical applications. More details about testing, changing, and improving the rendering and training speed of the original 3D Gaussian Splatting model \cite {kerbl20233d} will be presented in the following sections.

\section{Background and Related Work}

Trace to its source, the technology of the original 3D Gaussian Splatting model \cite {kerbl20233d} belongs to the research area of 3D representation and 3D reconstruction. The core technique of the 3D Gaussian Splatting model is the 3D Gaussian Splatting theory, which can be used in both the traditional explicit and implicit radiance fields. However, our study focuses more on the implicit radiance field with machine learning methods; therefore, only some related explicit techniques and implicit technologies will be discussed and analyzed here.

\subsection{Neural Radiance Field}

The Neural Radiance Field is a top-hot technology in the research area of the implicit radiance field \cite{mildenhall2021nerf} in these three years, but it was surpassed by the original 3D Gaussian Splatting model \cite {kerbl20233d} in research hot concern in 2023. Although the original NeRF is a good method for implicit 3D scene representation and reconstruction, many related technologies and theories were inspired by and based on it, for example, Mip-nerf as a multi-scale representation model \cite{barron2021mip}, Nerf in the wild as a wild scene reconstruction model \cite{martin2021nerf} and Block-nerf as a rendering model for large scalable scenes \cite{tancik2022block}. As an implicit rendering method based on learning and training, NeRF has made a great contribution to the development and evaluation of novel 3D representations and 3D reconstructions, which can be used as a prior successful development example for the research area and as a clue for the current 3D representation methods that are based on the original 3D Gaussian Splatting model. There are many existing research ideas of NeRF that can be used in designing or improving the research that is developed on the original 3D Gaussian Splatting model, following similar research approaches. For example,  researching the application of 3D Gaussian Splatting rendering in the reconstruction of the human head or face \cite{hong2022headnerf},  researching the implementation of dynamic rendering \cite{pumarola2021d} with 3D Gaussian Splatting, and researching approaches to improve the rendering speed \cite{garbin2021fastnerf} when the design is based on 3D Gaussian Splatting. The encoding and decoding compression idea was also inspired by the Instant-NGP \cite{muller2022instant}.

\subsection{Explicit and Implicit Research}

The research of 3D representation is usually divided into two categories, one is the explicit 3D representation, the other is the implicit 3D representation. Explicit 3D representation means using the direct representation of the 3D structure voxel \cite{zhang2001efficient}, such as the items of location, rotation, opacity, light or color to represent the structure, shape and color. While the implicit 3D representation means using the indirect representation of the 3D structure voxel \cite{sundermeyer2018implicit}, such as distance functions, fields, radiance fields to represent the structure, shape and color. Note that, whether using machine learning methods should not be a dogmatical criteria of whether a representation is explicit or implicit. Machine learning should be considered as a method that can be used in explicit or implicit 3D representations. In 1996, researchers began the work of using the learning methods for 3D representations \cite{beymer1996image}.

Analogously, the research of the radiance field is also usually divided into two categories, one is the explicit radiance fields \cite{fridovich2023k}, the other is the implicit radiance fields \cite{oechsle2021unisurf}. The significant feature of the explicit radiance fields is that all of the compositions of the radiance field can be calculated directly from the some explicit and certain formulas. Correspondingly, the main feature of the implicit radiance fields is that some of the compositions of the radiance field are indirectly gained from implicit calculations like deep learning. 

\begin{equation}  
\label{eq1}
\left\{  
\begin{array}{lr}  
R_{explicit}(x, y, z, \theta, \phi) = PointsData[(x, y, z)] \nabla f(\theta, \phi)  \\  
R_{implicit}(x, y, z, \theta, \phi) = F_{learning}(x, y, z, \theta, \phi) \to F_{\Theta}\to(RGB\sigma)    \\
\end{array}  
\right.  
\end{equation} 

The research on the explicit radiance field has been going on for decades \cite{nash1988extension}, while the research on the implicit radiance field has emerged in recent years. As shown in Equation \ref{eq1}, the input parameters of the main function type of the explicit radiance field $R_{explicit}$ will be handled and calculated with the explicit and certain functions. The parameters $(x, y, z)$ present the space location coordinates of the volume points. The parameters $(\theta, \phi)$ present the 2D viewing direction information like rotation and normal vector. The basic processing of the explicit radiance field tries to calculate all of the values of the components of the radiance field before rendering, which means making a pre-calculated dataset or data architecture to save these data first. For example, the parameter $PointsData$ likes a database that stores the space location coordinates of volume points. Operator $\nabla$ means other special operations for some specific parameters $(\theta, \phi)$. As for implicit radiance field, neural radiance field \cite{mildenhall2021nerf} is one of the most classic representatives,  with the same input parameter $(x, y, z, \theta, \phi)$, all of the parameters will be indirectly gained by the training model or neural network function $F_{learning}$, which is the MLP function $F_{\Theta}$ for the original NeRF \cite{mildenhall2021nerf}, and the final output is the calculated color $RGB$ and volume density $\sigma$.

As the development of radiance fields, the explicit and implicit technologies of radiance fields have been merged in more applications. For instance, the 3D Gaussian Splatting is a merged technology combining the traditional explicit technology and novel deep learning methods.  It may be a significant trend to get novel development or design in the future in this research area.

\subsection{3D Gaussian Splatting}

Analyzed as a pure technology, 3D Gaussian Splatting is a traditional explicit volume or  points cloud rendering technique \cite{mueller1999splatting}. The first successful related result about the 3D Gaussian Splatting technique can be traced back to EWA (elliptical weighted average) splatting \cite{zwicker2001ewa}, which has referenced the Heckbert's EWA texture filter in 1989 \cite{heckbert1989fundamentals}. In that time, 3D Gaussian Splatting is used to generate the high resolution surface rendering \cite{botsch2005high}. As the development, the 3D Gaussian Splatting technique has proved its advantages in aspects of 3D surface \cite{botsch2005high} or volume rendering \cite{zhang2014splatting}. The main rendering processing of current 3D Gaussian Splatting model \cite{kerbl20233d} is similar as the referenced initial EWA volume splatting, using the elliptical or spherical base equation to simulate the procedure of splatting to make the rendering, followed by project the rendered result from 3D space to 2D plane. As the technology burst of the current 3D Gaussian Splatting model \cite{kerbl20233d}, 3D Gaussian Splatting has been a new dominated technique than the NeRF \cite{mildenhall2021nerf}. Like the technological revolution boom sparked by NeRF, a similar novel research and application revolution boom sparked by 3D Gaussian Splatting is developing as mentioned above: e.g.: the dynamic 3D Gaussian Splatting \cite{wu20234d}, the human 3D Gaussian Splatting generation \cite{liu2023humangaussian}, and the multi-scalable 3D Gaussian Splatting \cite{yan2023multi} etc. History is sometimes similar.

\subsection{Spherical Harmonics Algorithm}

Spherical Harmonics \cite{muller2006spherical} is the core of rendering process of the 3D Gaussian Splatting model \cite{kerbl20233d}. Therefore it is necessary to understand the basic theories of Spherical Harmonics before the change and optimization. Laplace's equation $\nabla^{2}f(\vec{r}) = 0$ \cite{shortley1938numerical}, which is also known as the potential equation, the harmonic equation, belongs to the second order partial differential equation family \cite{da2002second}. The source of the Spherical Harmonics equation can be seen as the solution procedure and result of the representation of the solution of Laplace's equation in three-dimensional (3D) spherical coordinate system \cite{shen2009modeling}.

First, as shown in Equation \ref{eq2}, using the $(x, y, z) = (rsin\theta cos\varphi, rsin\theta sin\varphi, rcos\theta )$ to replace the $(x, y, z)$ in the 3D Cartesian coordinate system $\frac{\partial^{2}f}{\partial x^{2}} + \frac{\partial^{2}f}{\partial y^{2}} + \frac{\partial^{2}f}{\partial z^{2}} = 0$ and transfer the equation to 3D spherical coordinate system. Second, make the separation of variables and gain the expression as shown in Equation \ref{eq3}.

\begin{equation}  
	\label{eq2}
	\frac{1}{r^{2}}\frac{\partial}{\partial r}(r^{2}\frac{\partial f}{\partial r}) 
	+ \frac{1}{r^{2}sin\theta} \frac{\partial}{\partial \theta}(sin\theta \frac{\partial f}{\partial \theta})
	+ \frac{1}{r^{2}sin^{2}\theta} \frac{\partial^{2} f}{\partial \varphi^{2}} = 0
\end{equation}

\begin{equation}  
	\label{eq3}
	\left\{  
	\begin{array}{lr}  
		f(r, \theta, \varphi ) = R(r)\Theta(\theta)\Phi(\varphi)   \\
		\frac{1}{R}\frac{\partial}{\partial r}(r^{2}\frac{\partial R}{\partial r}) = \lambda   \\
		\frac{\partial^{2}\Phi}{\partial \varphi^{2}} = -m^{2}\Phi, m\in \mathbb{N}   \\
		sin\theta\frac{\partial}{\partial \theta}(sin\theta\frac{\partial \Theta}{\partial \theta}) = (m^{2}-\lambda sin^{2}\theta)\Theta   \\
	\end{array}  
	\right.  
\end{equation}

Considering the $\theta$, let $x$ be equal to $cos\theta$, let $\Theta \to P(x)$, let $\lambda $ be equal to $l(l+1)$, get the l-th order Associated Legendre Equation \cite{virchenko2001generalized}, the result is shown as Equation \ref{eq4}. The solution of the equation is called the Associated Legendre Function, which is shown as Equation \ref{eq5}. $L=l$ when $k$ is even number while $L=l-1$ when $k$ is odd number.

\begin{equation}  
	\label{eq4}
	\frac{\partial}{\partial x}[(1-x^{2})\frac{\partial P(x)}{\partial x}] + [l(l+1)-\frac{m^{2}}{1-x^{2}}]P(x) = 0, x\in \mathbb{R}, \left | x \right | \le 1 
\end{equation}

\begin{equation}  
	\label{eq5}
	P_{l} = \sum_{k=0}^{\frac{L}{2}}(-1)^{k}\frac{(2(l-k))!}{2^{l}k!(l-k)!(l-2k)!}x^{l-2k} 
\end{equation}

Next, make use of the inherent features of Legendre Equation, such as the features of orthonormality, Rodrigues Equation, parity, generation, and recursion, etc, to make the calculation, as shown in Equation \ref{eq6}. There are many others recursion relationships \cite{lundberg1988recursion} of Legendre Equation, here only some of them were selected. With the feature of recursion, we can calculate the following expression and relationship of $P_{l}(x)$.

\begin{equation}  
	\label{eq6}
	\left\{  
	\begin{array}{lr}  
		Orthonormality\to \int_{-1}^{1}\sqrt{\frac{2m+1}{2}}P_{m}(x) \sqrt{\frac{2n+1}{2}}P_{n}(x)dx=\delta_{mn}   \\
		Rodrigues \to P_{l}(x)=\frac{1}{2^{l}l!}\frac{\partial^{l}}{\partial x^{l}}(x^2-1)^{l}   \\
		Parity\to P_{l}(-x)=(-1)^{l}P_{l}(x)   \\
		Generation\to \frac{1}{\sqrt{1-2rx+r^{2}}} = \sum_{l=0}^{\infty}P_{l}(x)r^{l}\Leftarrow r<1   \\
		Recursion1\to P'_{l+1}(x)-P'_{l-1}(x)=(2l+1)P_{l}(x)   \\  
		Recursion2\to P_0(x)=1,P_1(x)=x   \\  
		Recursion3\to (l+1)P_{l+1}(x)=(2l+1)xP_{l}(x)-lP_{l-1}(x),l\in \mathbb{N}^{+}    \\  
	\end{array}  
	\right.  
\end{equation}

Then, let $x=cos\theta$ and substitute it back into the Legendre Equation, can get the corresponding expressions of Spherical Harmonics, as shown in Equation \ref{eq7}. Here, the $l$ represents the degree of Spherical Harmonics, and $m$ represents the order of Spherical Harmonics. These two parameters are significant for the representation of color with Spherical Harmonics. The solution of the Legendre Equation which is about $\varphi$ is represented as $\Phi_{m}(\varphi)$.

\begin{equation}  
	\label{eq7}
	\left\{  
	\begin{array}{lr}  
		P_{l}^{m}=(1-x^{2})^{\frac{m}{2}}\frac{\partial^{m}}{\partial x^{m}}P_{l}    \\
		P_{1}^{1}=sin\theta, P_{2}^{1}=\frac{3}{2}sin2\theta, P_{2}^{2}=\frac{3}{2}(1-cos2\theta)  \\ 
		P_{1}^{-1}=-\frac{1}{2}sin\theta, P_{2}^{-1}=-\frac{1}{4}sin2\theta   \\
		\Phi_{m}(\varphi) = A_{m}cosm\varphi+B_{m}sinm\varphi   \\
	\end{array}  
	\right.  
\end{equation} 

Finally, combine the intrinsic solutions of $\theta$ and $\varphi$ and make the normalization, get the Spherical Harmonics function, as shown in Equation \ref{eq8}.

\begin{equation}  
	\label{eq8}
	Y^m_l(\theta, \varphi) = \left\{  
	\begin{array}{lr}  
		\sqrt{2} K^{m}_{l}cos(m\varphi)P^{m}_{l}(cos\theta) \gets (m > 0)   \\  
		\sqrt{2} K^{m}_{l}sin(-m\varphi)P^{-m}_{l}(cos\theta) \gets (m < 0)   \\
		K^{0}_{l}P^{0}_{l}(cos\theta) \gets (m = 0)   \\
	\end{array}  
	\right.  
\end{equation}

Here, $K_{l}^{m}$ is a special brief expression and represents the scaling factor, which is used to make the normalization calculation, as shown in Equation \ref{eq9}.

\begin{equation}  
	\label{eq9}
	K_{l}^{m}=\sqrt{\frac{2l+1}{4\pi}\cdot \frac{(l-\left|m\right|)!}{(l+\left|m\right|)!}}, l\in \mathbb{Z}^{+}, m\in \mathbb{Z}, \left|m\right|\le l   
\end{equation}


\section{Methodology and Evaluation}

In this section, we will start the principle analysis and test experiments results evaluations to discuss the performances of various improvements.

\subsection{Color}

The analysis and processing of the component color is always considered as a significant item of 3D representation in its history of development. Due to that in the current 3D Gaussian Splatting model \cite{kerbl20233d}, using the orders and degree of Spherical Harmonics function to represent the color is an important task, we made the corresponding explorations to test the different impacts caused by the different kinds of compositions of a RGB color or single channel gray-scale color.

\begin{table}[h]
	\caption{ Comparison of Impacts by Different Color Split Components.}
	\label{tb1}
	\centering
	\begin{tabular}{  c |c | c | c | c | c  }
		\hline
		\textbf{it} & \textbf{origin} & \textbf{red} & \textbf{green} & \textbf{blue} & \textbf{gray}
		\\ \hline
		ri & \begin{minipage}[b]{0.18\columnwidth}
			\centering
			\raisebox{-.2\height}{\includegraphics[width=\linewidth]{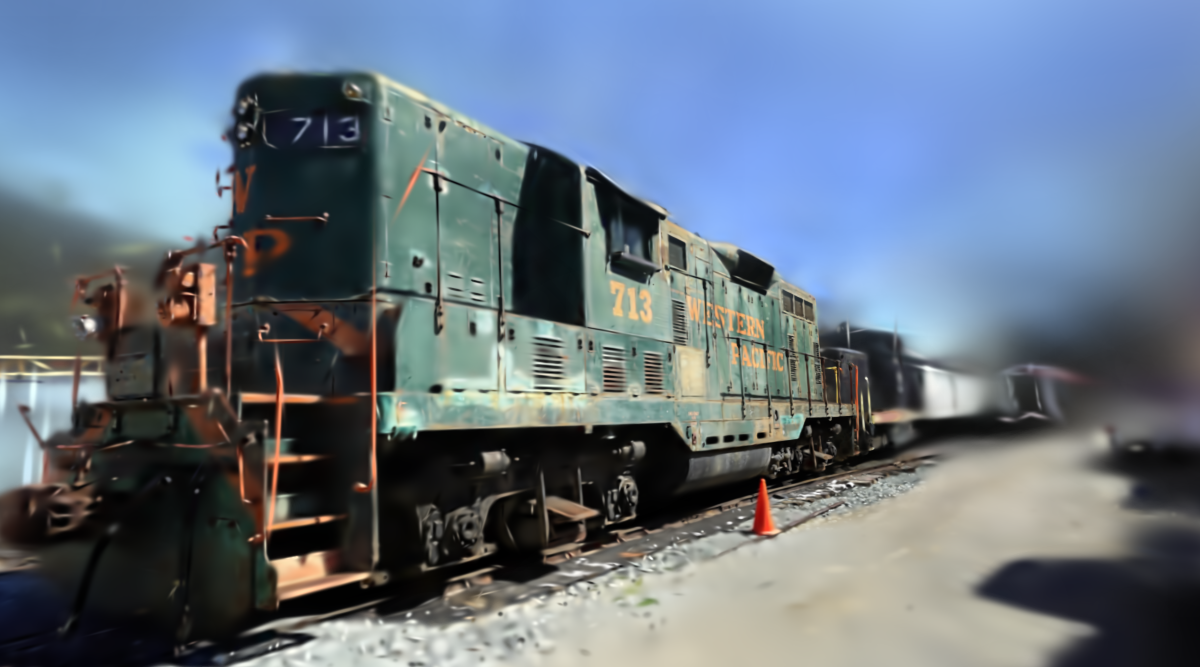}}
		\end{minipage}
		& \begin{minipage}[b]{0.18\columnwidth}
			\centering
			\raisebox{-.2\height}{\includegraphics[width=\linewidth]{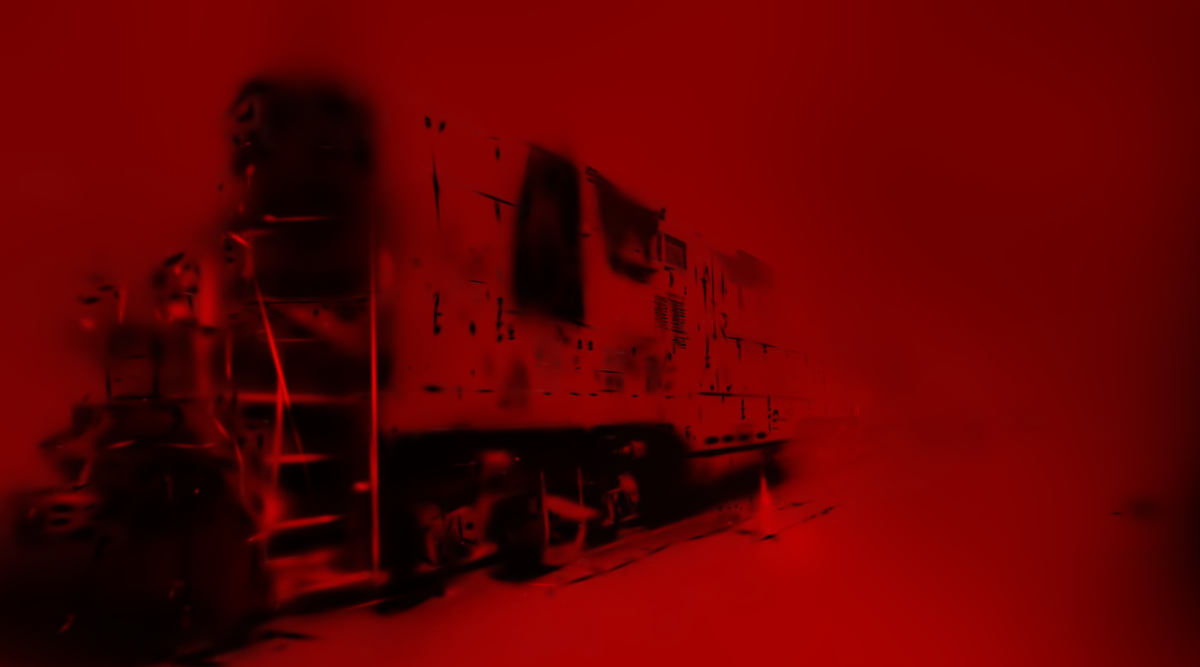}}
		\end{minipage}
		& \begin{minipage}[b]{0.18\columnwidth}
			\centering
			\raisebox{-.2\height}{\includegraphics[width=\linewidth]{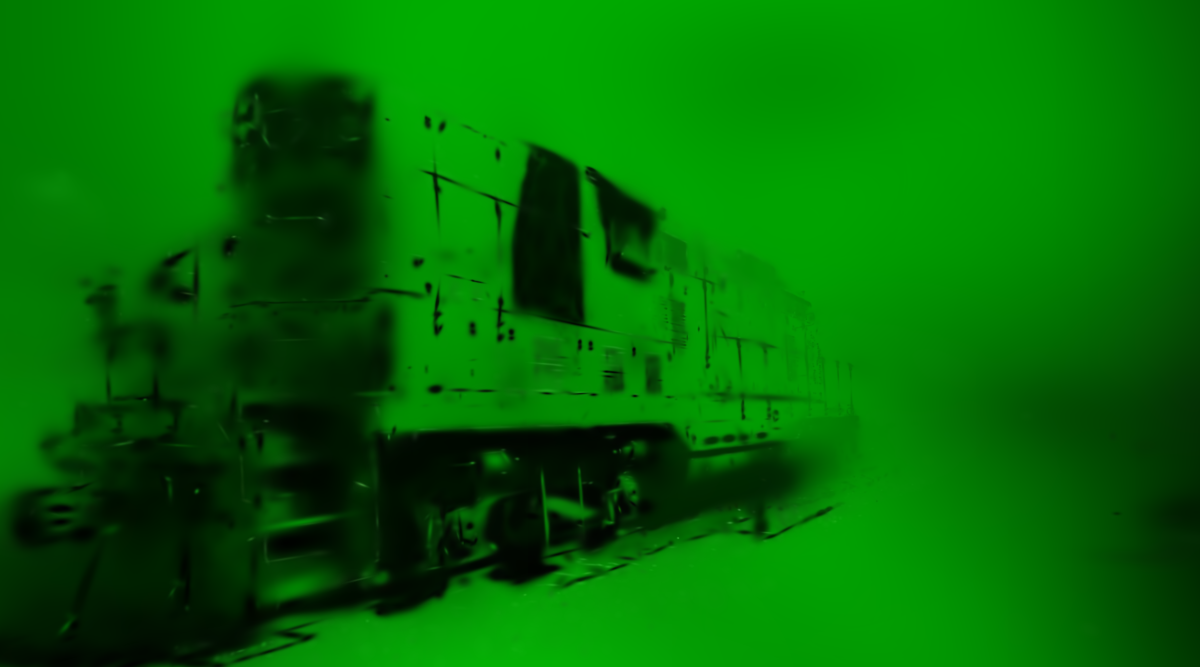}}
		\end{minipage} 
		& \begin{minipage}[b]{0.18\columnwidth}
			\centering
			\raisebox{-.2\height}{\includegraphics[width=\linewidth]{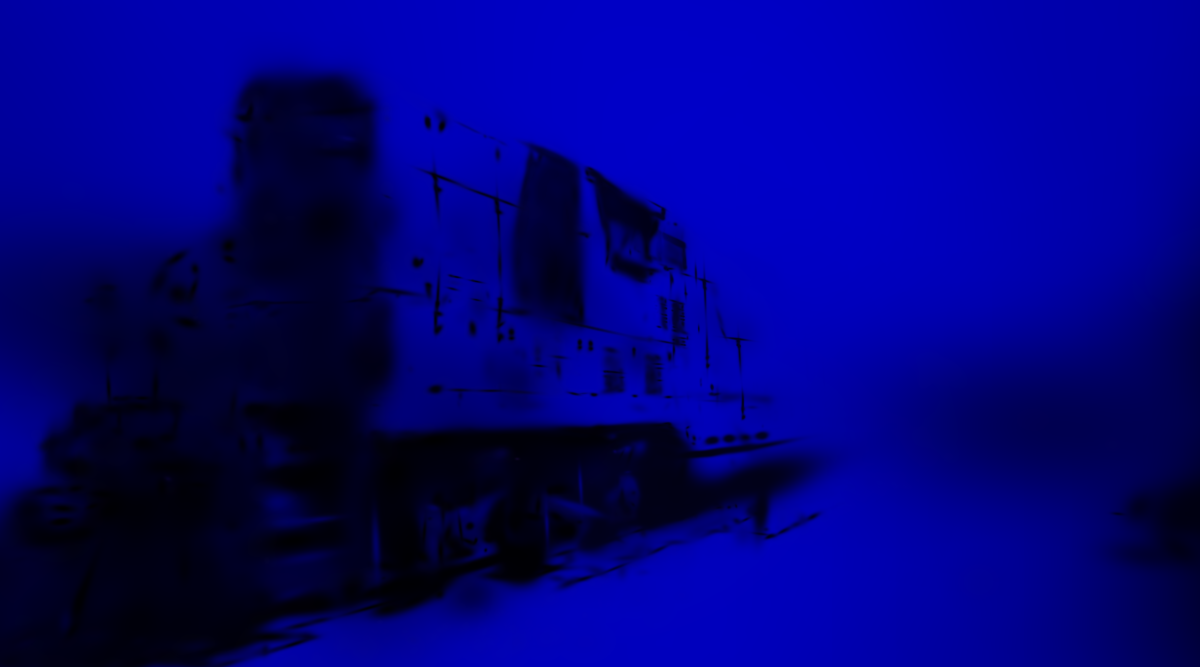}}
		\end{minipage} 
		& \begin{minipage}[b]{0.18\columnwidth}
			\centering
			\raisebox{-.2\height}{\includegraphics[width=\linewidth]{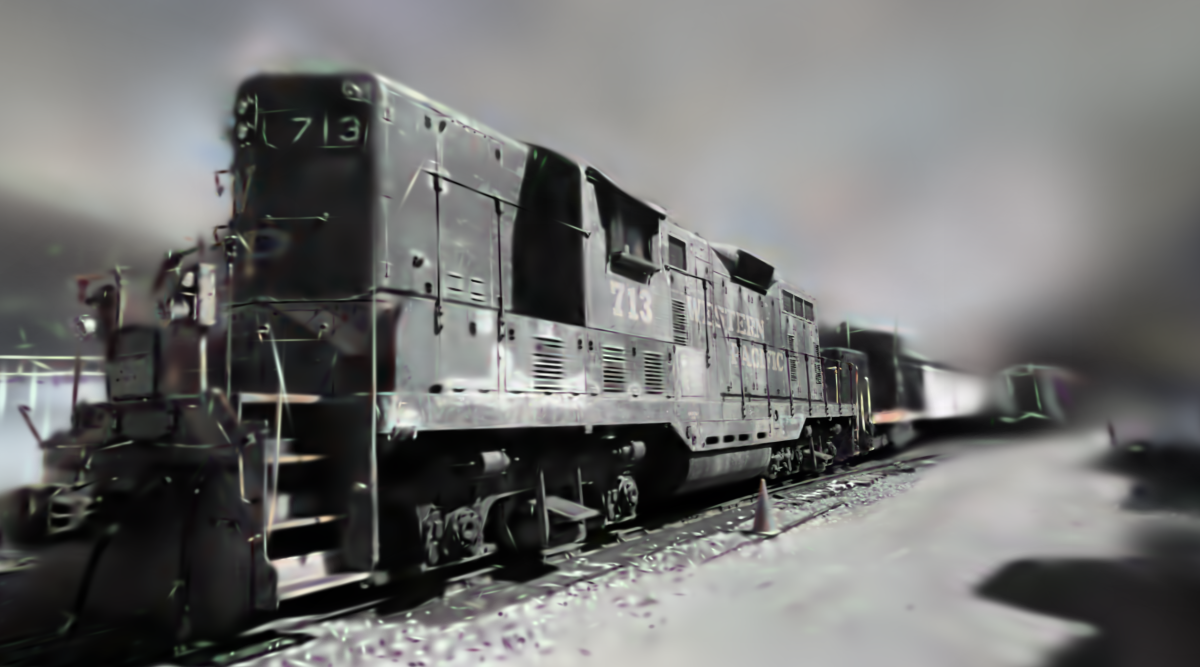}}
		\end{minipage} 
		
		\\ \hline
		
		ls & \begin{minipage}[b]{0.18\columnwidth}
			\centering
			\raisebox{-.2\height}{\includegraphics[width=\linewidth]{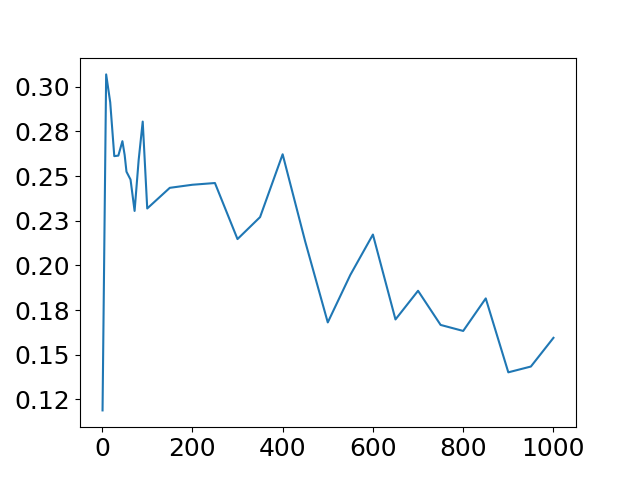}}
		\end{minipage}
		& \begin{minipage}[b]{0.18\columnwidth}
			\centering
			\raisebox{-.2\height}{\includegraphics[width=\linewidth]{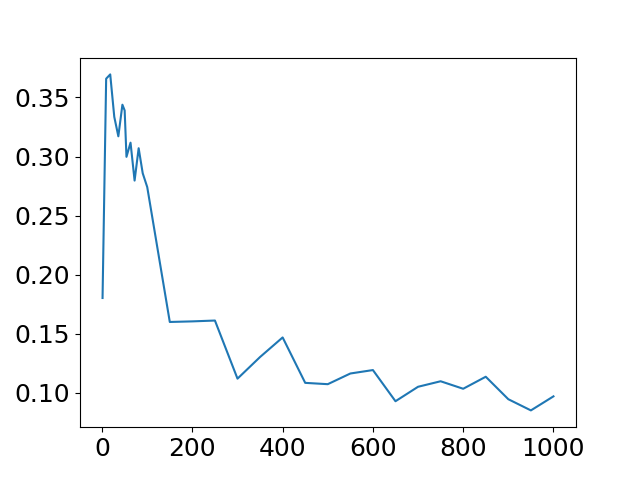}}
		\end{minipage}
		& \begin{minipage}[b]{0.18\columnwidth}
			\centering
			\raisebox{-.2\height}{\includegraphics[width=\linewidth]{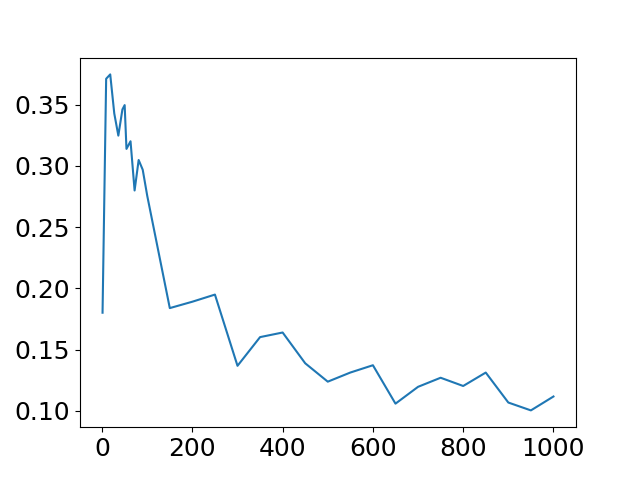}}
		\end{minipage} 
		& \begin{minipage}[b]{0.18\columnwidth}
			\centering
			\raisebox{-.2\height}{\includegraphics[width=\linewidth]{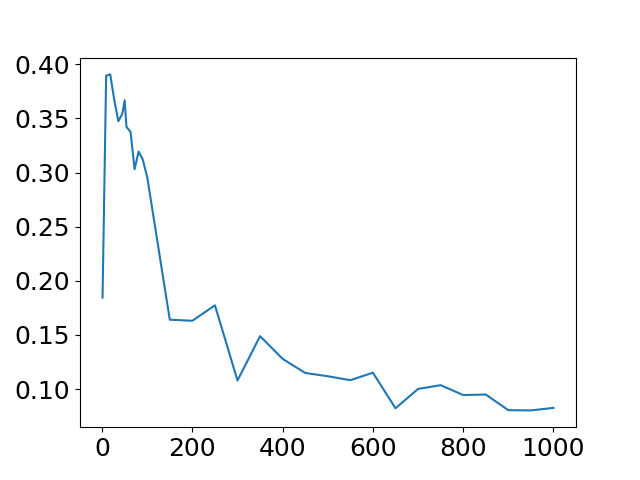}}
		\end{minipage} 
		& \begin{minipage}[b]{0.18\columnwidth}
			\centering
			\raisebox{-.2\height}{\includegraphics[width=\linewidth]{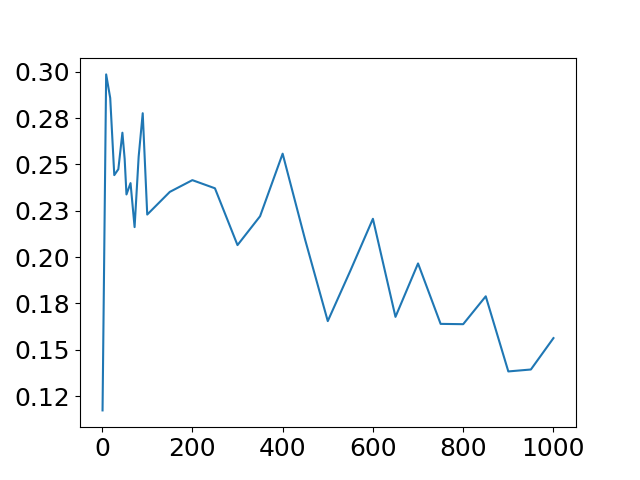}}
		\end{minipage} 

		\\ \hline
		
		l1 &  0.10339
		& 0.05882
		& 0.06358 
		& 0.06102
		& 0.10287
		
		\\ \hline
		
		l2 &  0.07707
		& 0.05295
		& 0.05455 
		& 0.05148
		& 0.07448
		
		\\ \hline
		
		p1 & 17.2078
		& 32.07301
		& 28.72609 
		& 36.24833
		& 17.3931
		
		\\ \hline
		
		p2 & 18.85038
		& 33.30938
		& 30.24411
		& 37.50113
		& 19.23794
		
		\\ \hline
		
		pt & 9.38438
		& 6.72812
		& 7.2
		& 6.34688
		& 11.92812
		
		\\ \hline
		
		pc & 29.13134
		& 19.83754
		& 20.54045
		& 20.1017
		& 31.12973
		
		\\ \hline
	\end{tabular}
\end{table}

As shown in Table \ref{tb1}, $it$ represents the different kinds of the rendering items. The parameter $ri$ represents the final viewing image of different rendering results. The parameter $ls$ represents the record and changing of the real time learning loss as the training is going. $l1$ represents the final test loss of the evaluation. $l2$ represents the final train loss of the evaluation. $p1$ represents the final test Peak Signal to Noise Ratio (PSNR) \cite{hore2010image} value of the evaluation. $p2$ represents the final train PSNR value of the evaluation. The parameter $pt$ in seconds represents the pure CPU and GPU processing time. The parameter $pc$ in seconds represents the performance counter time, including the processing waiting time. All of the experiment results of this paper were gained from a laptop with a Nvidia laptop-version 4090 GPU and an Intel i9-13900HX CPU. Omit the same variables explanation when others used the same parameters in this paper.

According to the Equation \ref{eq8}, different kinds of color will be represented by different degree $l$ and order $m$ of the Spherical Harmonics function. Due to different color  split single channels having different main color components, the Spherical Harmonics parameters used to represent them are different, which causes the different rendering result as shown in Table \ref{tb1}. The training steps for each record is 1000, 31 times of losses will be recorded during the training. As shown in Table \ref{tb1}, it will be faster in training and has a better rate of convergence in training loss, and has a higher rendering quality, if only one color channel data of input images is selected. The reason for this kind of phenomenon  is mainly about the reduction of partial rendering information of input data. We found that it only changes the color of the input data from RGB to gray-scale, it still remains the three channels from the image data to be processed, which leads the time of simple gray-scale image for rendering is similar with the normal RGB images.  Subsequently, we changed the code of the model and used the pure one channel gray-scale images for the evaluation, the final speed has not been improved to a good level, due to more time needed to be used in transformation processing of from RGB images to gray-scale images.

Subsequently, we considered the tool of color component in the current 3D Gaussian Splatting model \cite{kerbl20233d}. It is often that the time used to make the learning and training of the color is usually one of the longests, for most of the 3D representations that are based on deep learning. Actually we can consider the color component as a fixed component and omit it in the learning process. Only adding back the pre-calculated color component when the learning finishes. This idea we call the adding back idea, which means omitting some components that are not very primary important and not must be gained by learning but need more time to be gained in learning, and adding them back after the learning.

\begin{table}[h]
	\caption{Comparison of Impacts without All of the Color Components.}
	\label{tb2}
	\centering
	\begin{tabular}{  c | c | c | c | c | c   }
		\hline
		\textbf{it}  & \textbf{origin} & \textbf{truck} & \textbf{truck} & \textbf{truck} & \textbf{truck} 
		\\ \hline
		ri & \begin{minipage}[b]{0.18\columnwidth}
			\centering
			\raisebox{-.2\height}{\includegraphics[width=\linewidth]{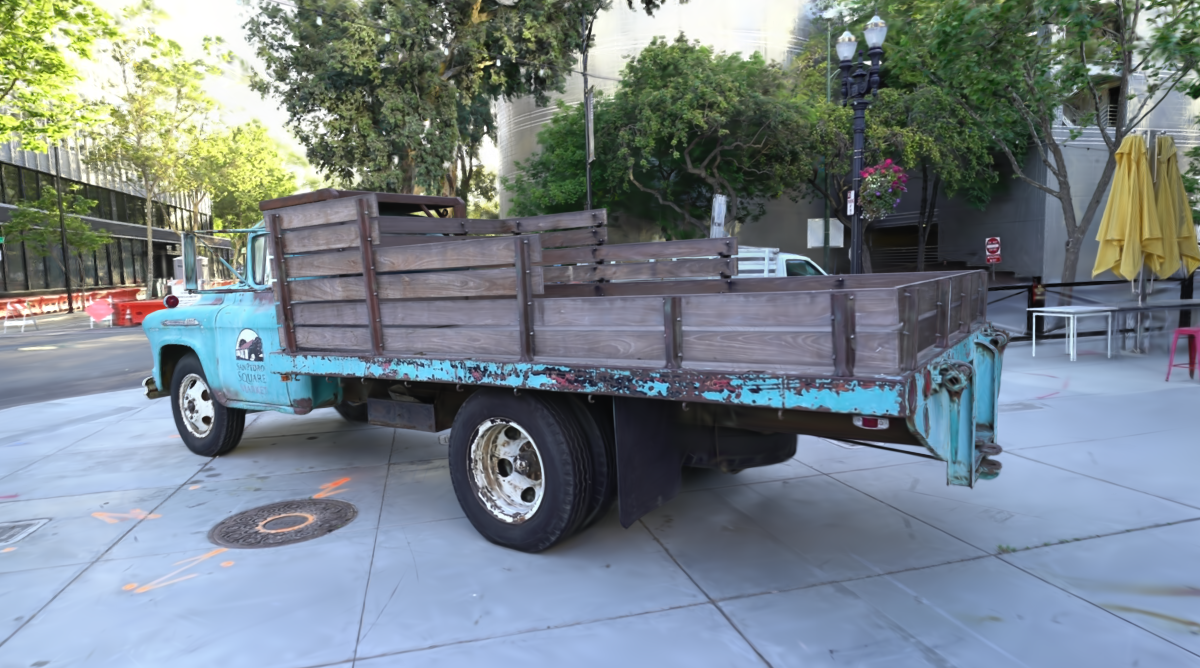}}
		\end{minipage}
		& \begin{minipage}[b]{0.18\columnwidth}
			\centering
			\raisebox{-.2\height}{\includegraphics[width=\linewidth]{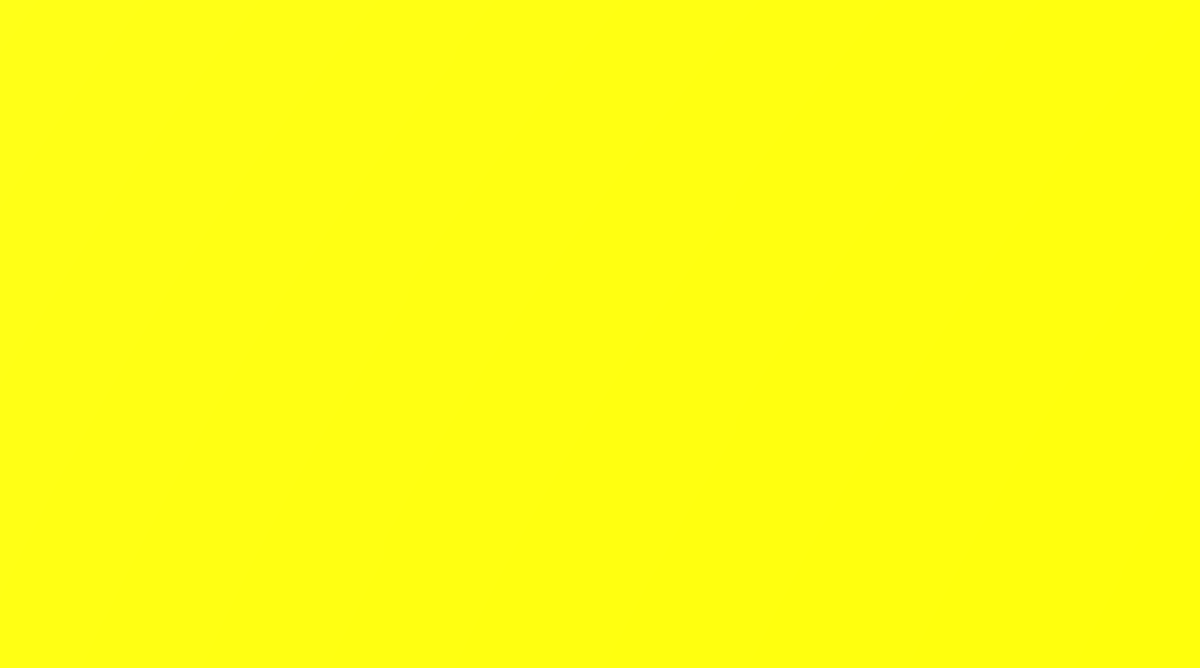}}
		\end{minipage}
		& \begin{minipage}[b]{0.18\columnwidth}
			\centering
			\raisebox{-.2\height}{\includegraphics[width=\linewidth]{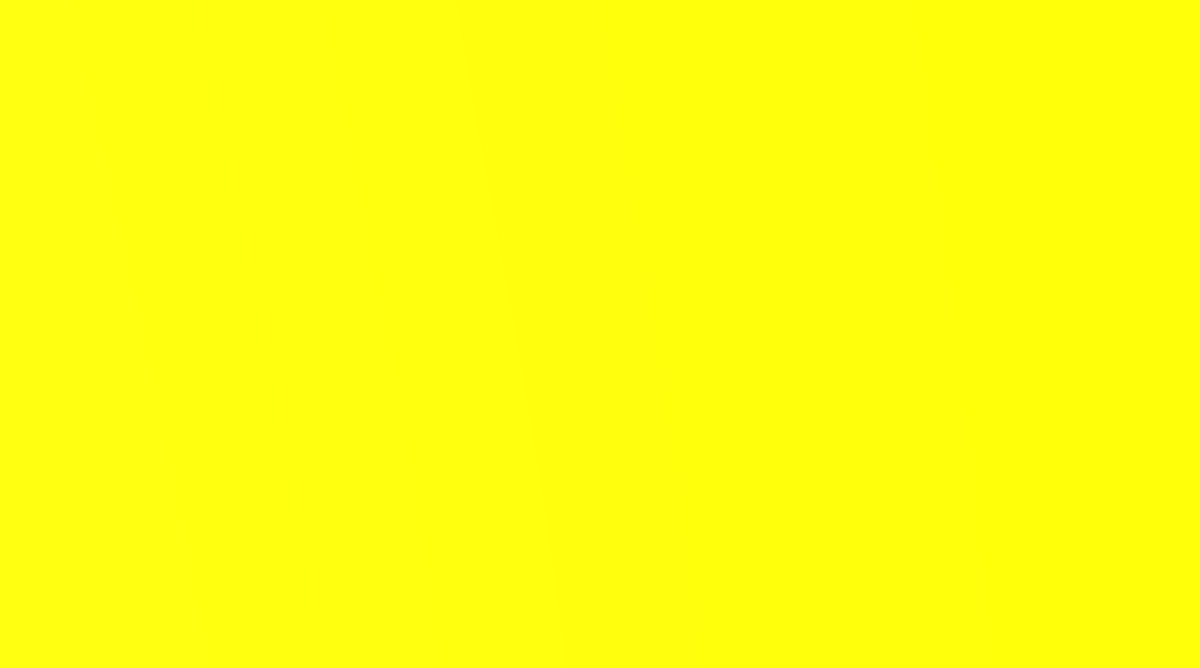}}
		\end{minipage} 
		& \begin{minipage}[b]{0.18\columnwidth}
			\centering
			\raisebox{-.2\height}{\includegraphics[width=\linewidth]{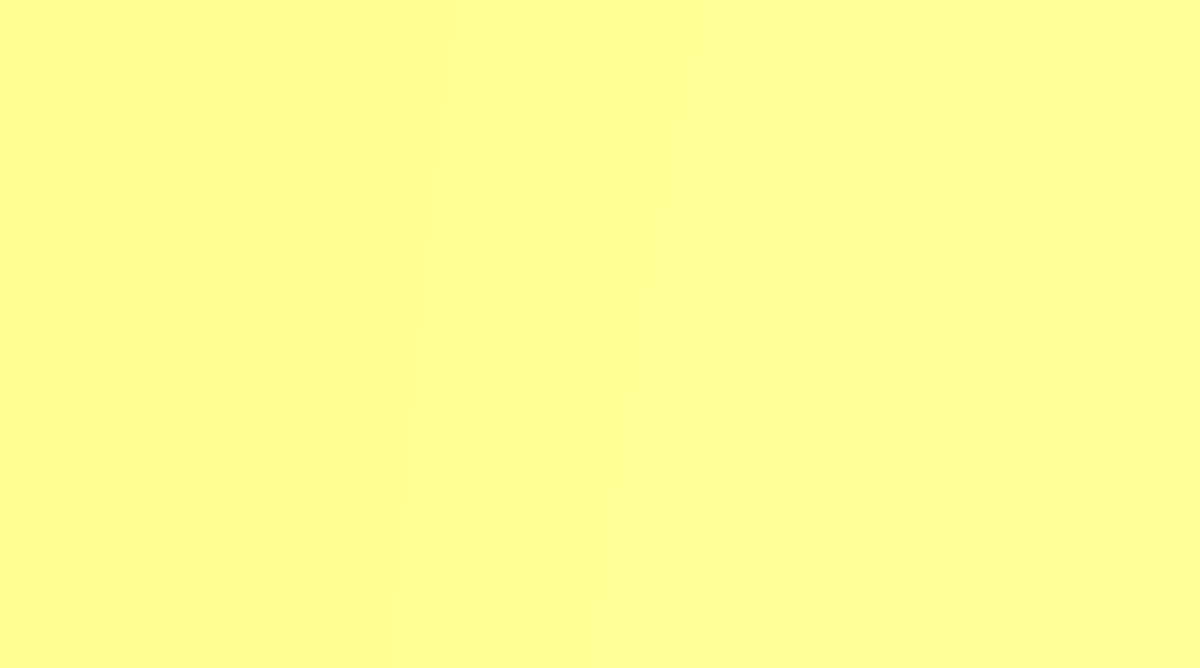}}
		\end{minipage} 
		& \begin{minipage}[b]{0.18\columnwidth}
			\centering
			\raisebox{-.2\height}{\includegraphics[width=\linewidth]{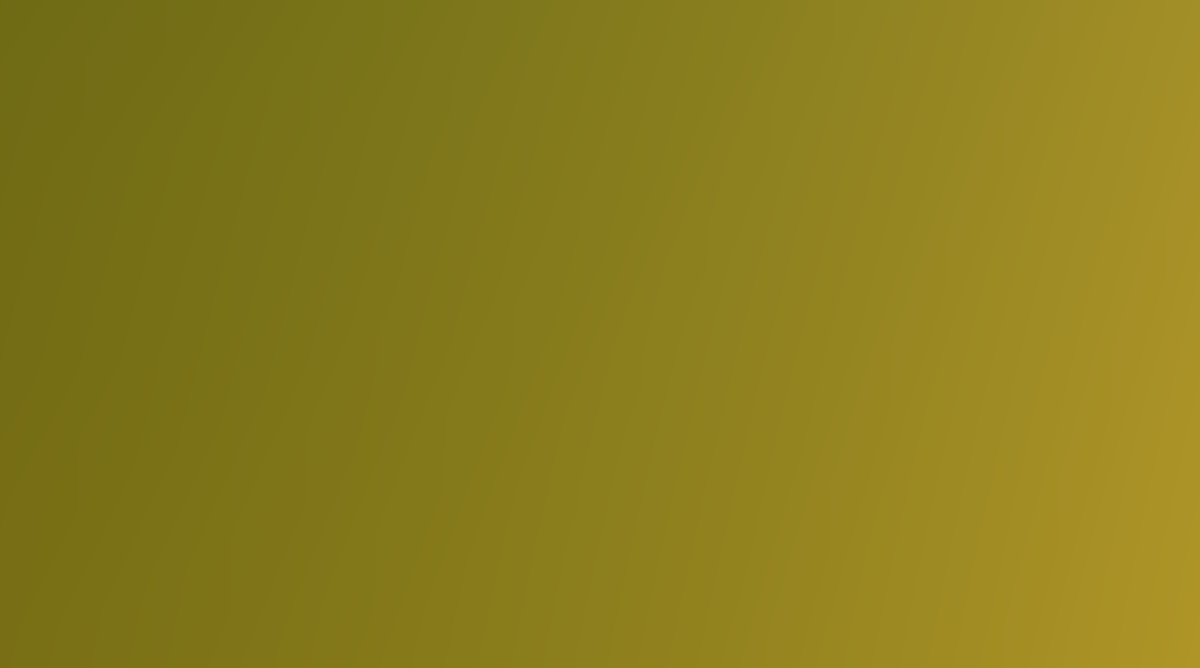}}
		\end{minipage} 
		
		\\ \hline
		
		ls & \begin{minipage}[b]{0.18\columnwidth}
			\centering
			\raisebox{-.2\height}{\includegraphics[width=\linewidth]{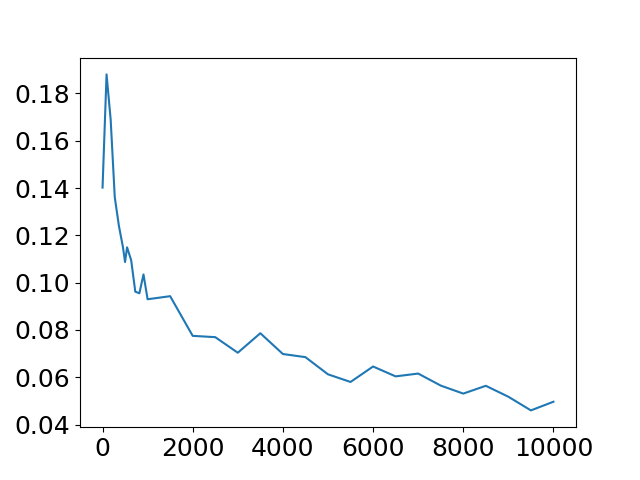}}
		\end{minipage}
		& \begin{minipage}[b]{0.18\columnwidth}
			\centering
			\raisebox{-.2\height}{\includegraphics[width=\linewidth]{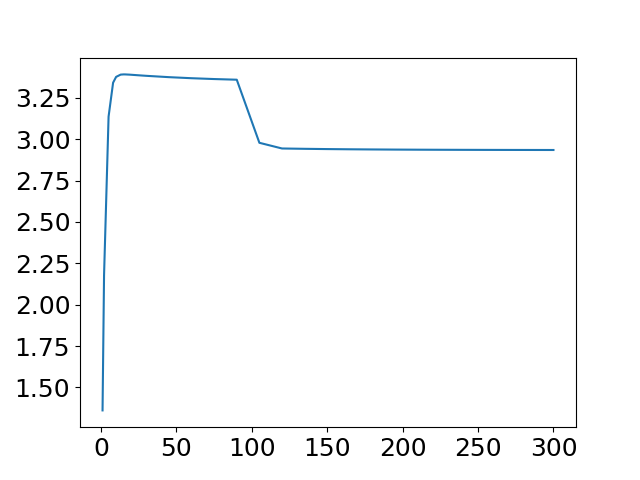}}
		\end{minipage}
		& \begin{minipage}[b]{0.18\columnwidth}
			\centering
			\raisebox{-.2\height}{\includegraphics[width=\linewidth]{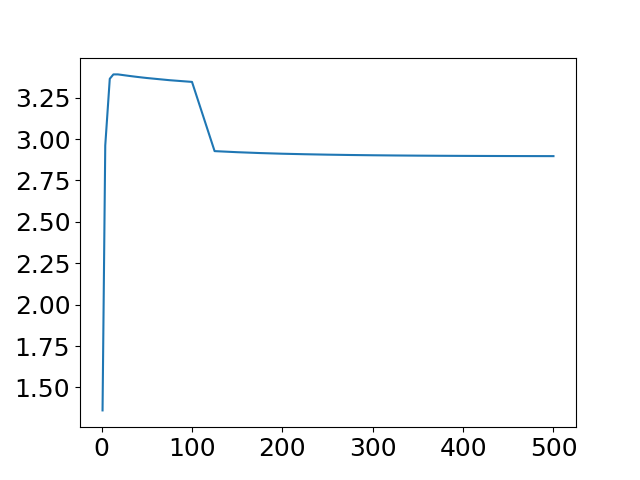}}
		\end{minipage} 
		& \begin{minipage}[b]{0.18\columnwidth}
			\centering
			\raisebox{-.2\height}{\includegraphics[width=\linewidth]{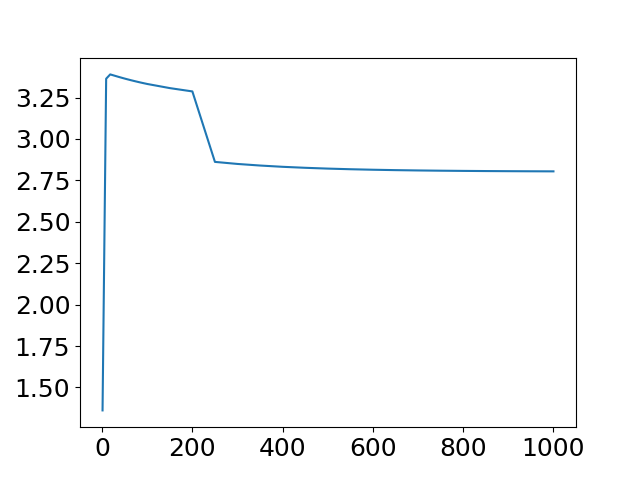}}
		\end{minipage} 
		& \begin{minipage}[b]{0.18\columnwidth}
			\centering
			\raisebox{-.2\height}{\includegraphics[width=\linewidth]{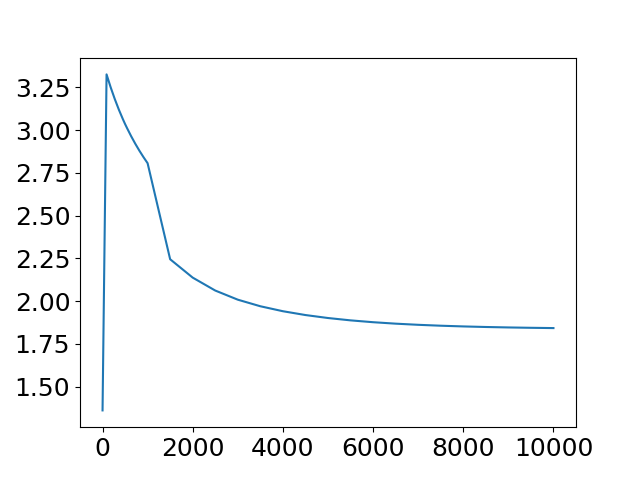}}
		\end{minipage}

		\\ \hline
		
		l1 &  0.03664
		& 2.93507
		& 2.89677
		& 2.80446
		& 1.84268
		
		\\ \hline
		
		pt & 87.40234
		& 0.45312
		& 0.21562
		& 0.44062
		& 7.44688
		
		\\ \hline
		
		pc & 318.55518
		& 1.31667
		& 1.51708
		& 2.79653
		& 60.63582
		
		\\ \hline
		
		st & 10000
		& 300
		& 500
		& 1000
		& 10000
		
		\\ \hline
	\end{tabular}
\end{table}

Based on the idea of adding back, we made an experiment to remove all of the color components in the model architecture level of the 3D Gaussian Splatting model \cite{kerbl20233d} and tested the improvement of the learning without all of the color components, which has gained a series of results as shown in Table \ref{tb2}. The invigorating thing is without the learning of all of the color components, which means the training process will nor include the calculation of the degree $l$ and order $m$ of the Spherical Harmonics function as shown in Equation \ref{eq8} and Equation \ref{eq9}, the training speed will has a 5-8 times improvement approximately.

However, adding the color back is a difficult task. We made many explorations and found a feasible method to add the color components back but could not achieve a satisfactory result. We used the feature functions $torch.save()$ and $torch.load()$ of Pytorch to save and load the pre-calculated features of the 3D Gaussian Splatting model respectively. It was not easy to align the location $x, y, z$ coordinates data with the other pre-calculated features data, especially the color data. But we found the time used to reload the pre-calculated features data is enough fast to be omitted, which means this idea is still promising.

\begin{table}[h]
	\caption{ Comparison of Impacts by Different Max Spherical Harmonics Degree.}
	\label{tb3}
	\centering
	\begin{tabular}{  c |c | c | c | c | c  }
		\hline
		\textbf{it} & \textbf{degree=0} & \textbf{degree=1} & \textbf{degree=2} & \textbf{degree=3} & \textbf{change code}
		\\ \hline
		ri & \begin{minipage}[b]{0.18\columnwidth}
			\centering
			\raisebox{-.2\height}{\includegraphics[width=\linewidth]{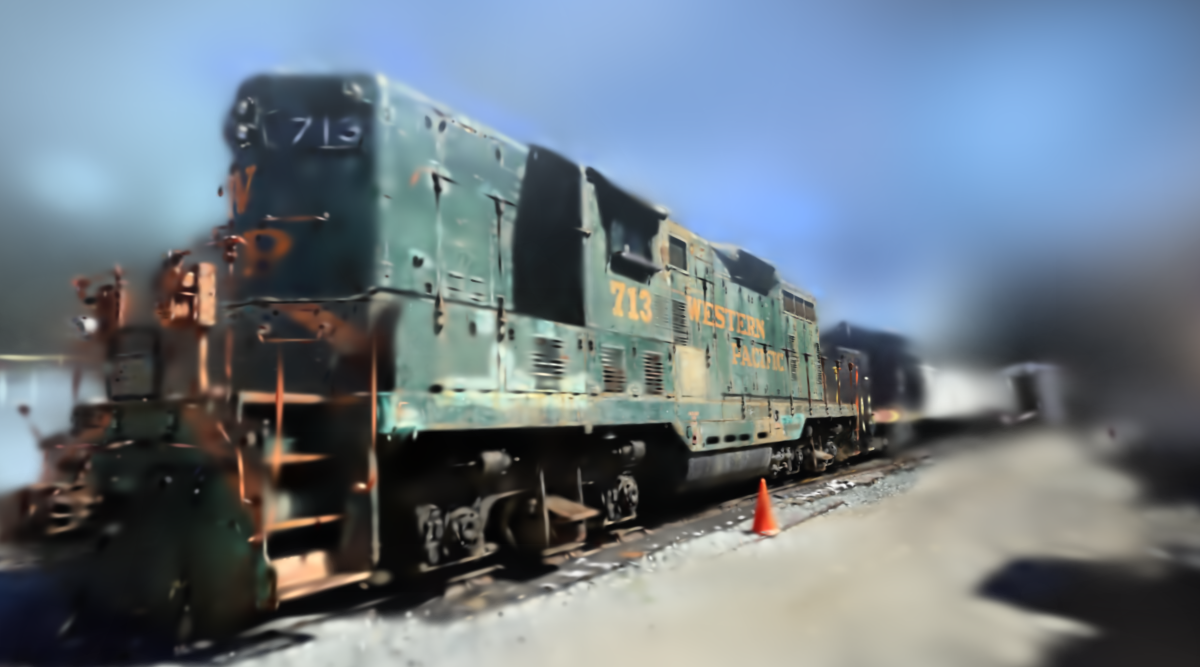}}
		\end{minipage}
		& \begin{minipage}[b]{0.18\columnwidth}
			\centering
			\raisebox{-.2\height}{\includegraphics[width=\linewidth]{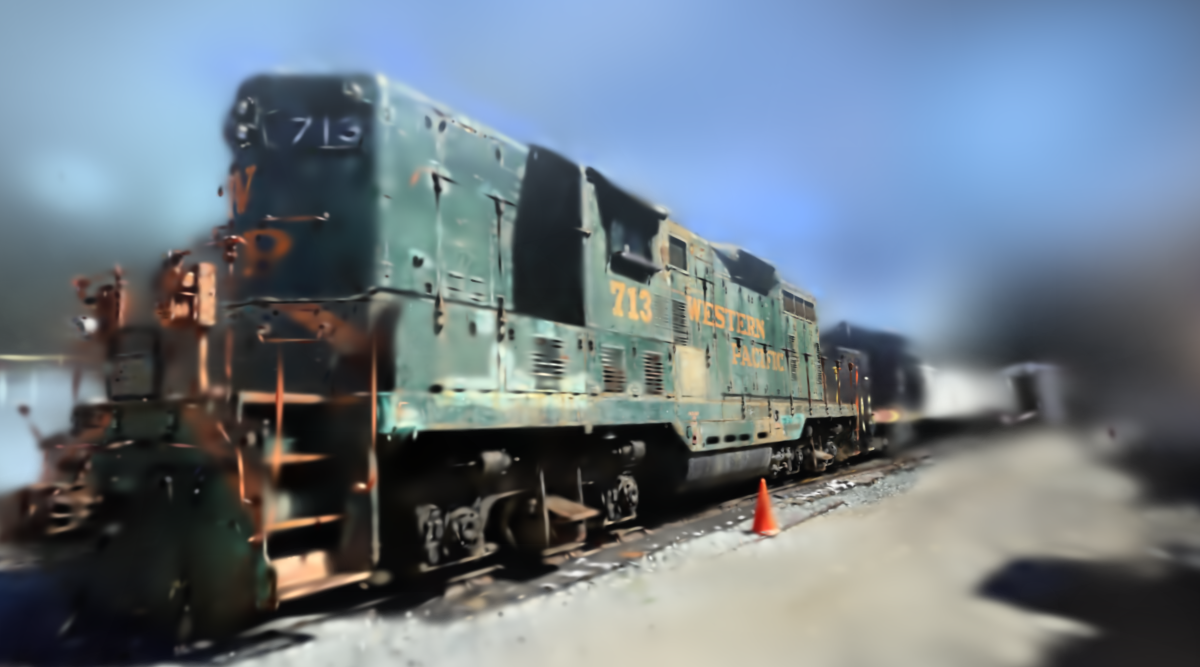}}
		\end{minipage}
		& \begin{minipage}[b]{0.18\columnwidth}
			\centering
			\raisebox{-.2\height}{\includegraphics[width=\linewidth]{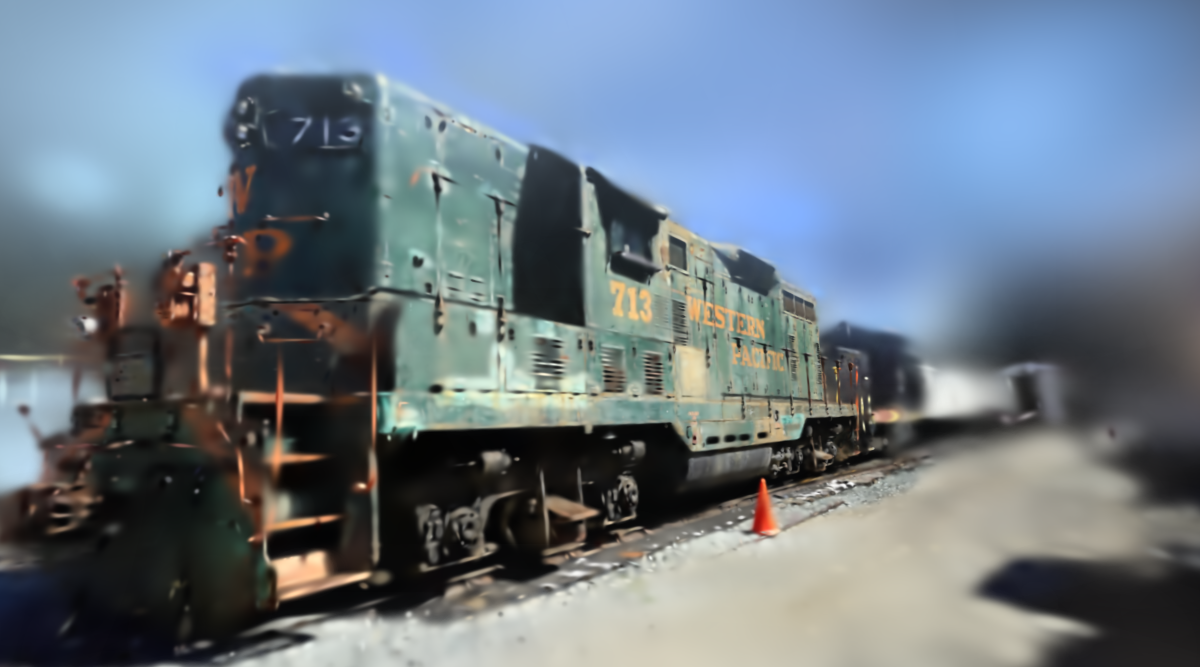}}
		\end{minipage} 
		& \begin{minipage}[b]{0.18\columnwidth}
			\centering
			\raisebox{-.2\height}{\includegraphics[width=\linewidth]{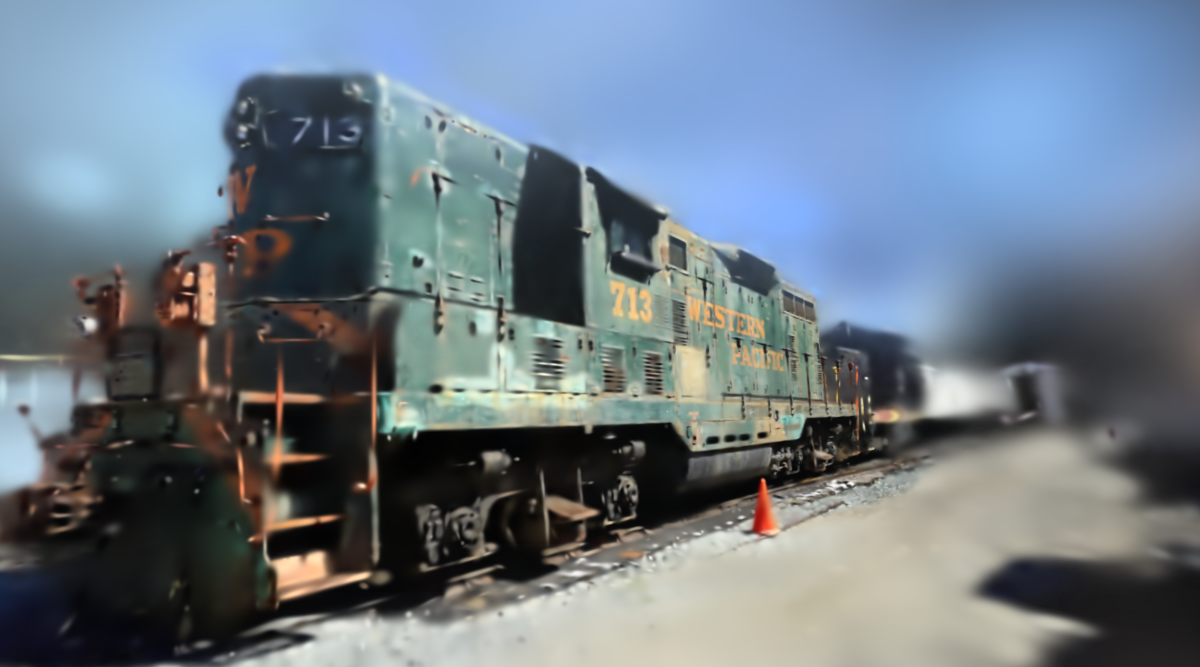}}
		\end{minipage} 
		& \begin{minipage}[b]{0.18\columnwidth}
			\centering
			\raisebox{-.2\height}{\includegraphics[width=\linewidth]{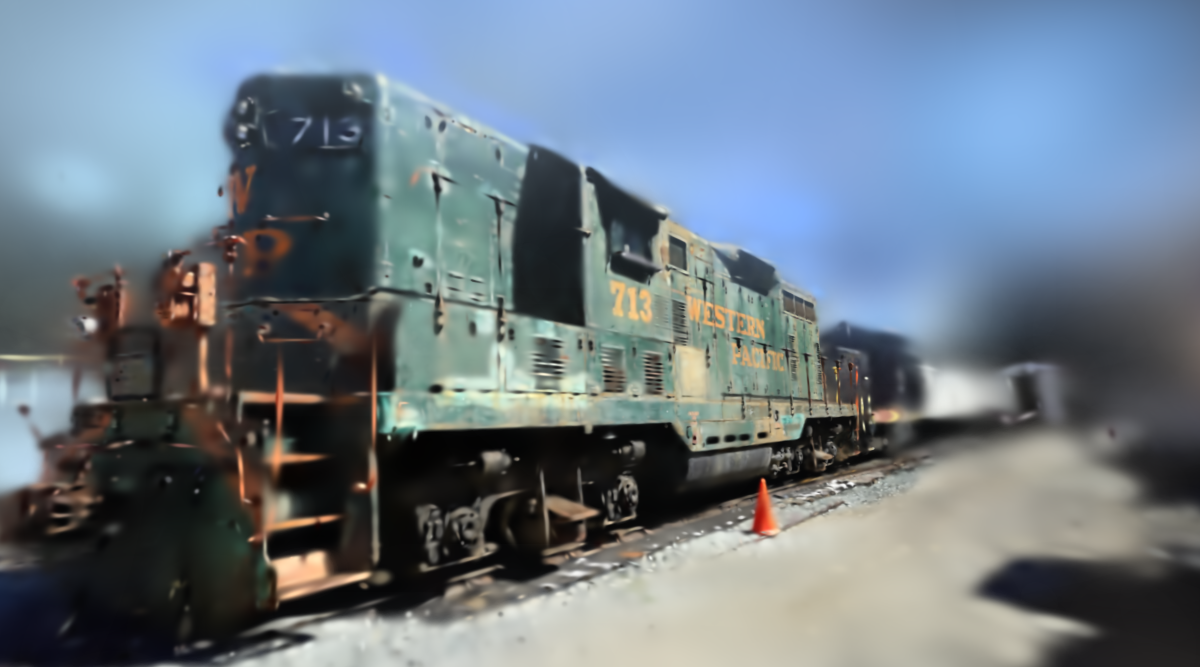}}
		\end{minipage} 
		
		\\ \hline
		
		ls & \begin{minipage}[b]{0.18\columnwidth}
			\centering
			\raisebox{-.2\height}{\includegraphics[width=\linewidth]{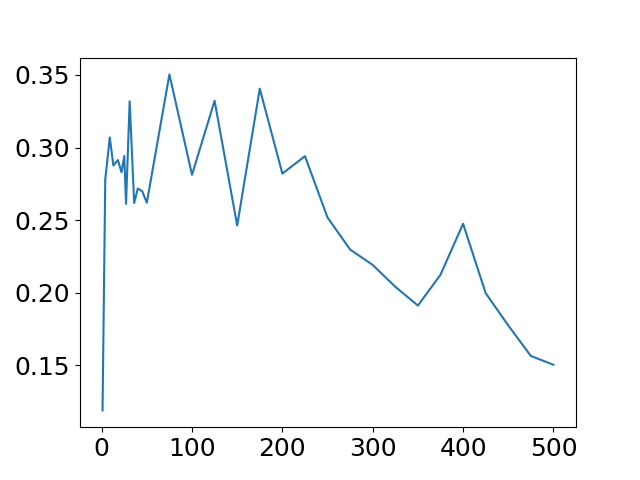}}
		\end{minipage}
		& \begin{minipage}[b]{0.18\columnwidth}
			\centering
			\raisebox{-.2\height}{\includegraphics[width=\linewidth]{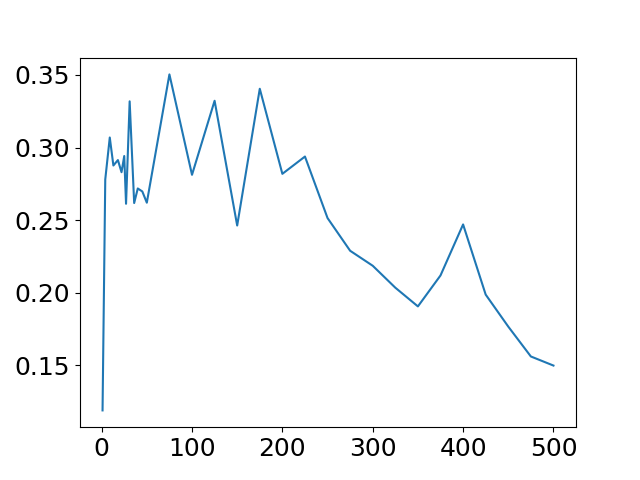}}
		\end{minipage}
		& \begin{minipage}[b]{0.18\columnwidth}
			\centering
			\raisebox{-.2\height}{\includegraphics[width=\linewidth]{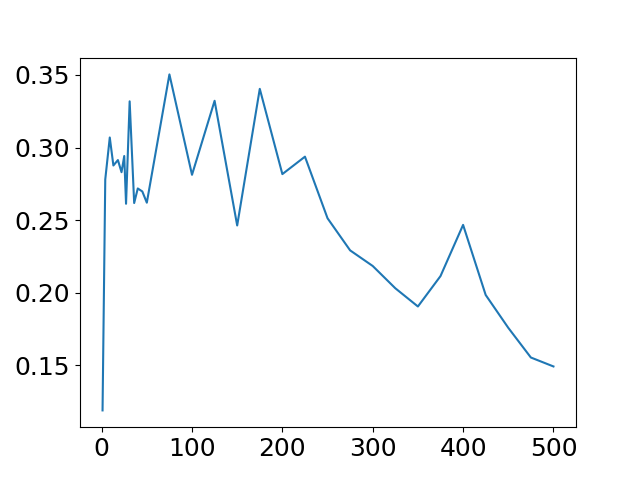}}
		\end{minipage} 
		& \begin{minipage}[b]{0.18\columnwidth}
			\centering
			\raisebox{-.2\height}{\includegraphics[width=\linewidth]{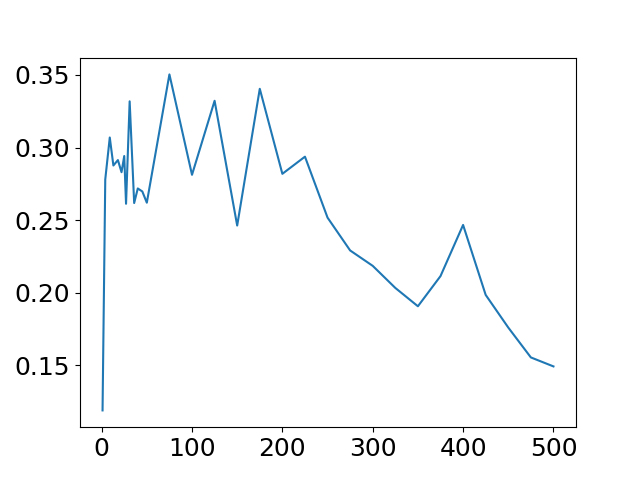}}
		\end{minipage} 
		& \begin{minipage}[b]{0.18\columnwidth}
			\centering
			\raisebox{-.2\height}{\includegraphics[width=\linewidth]{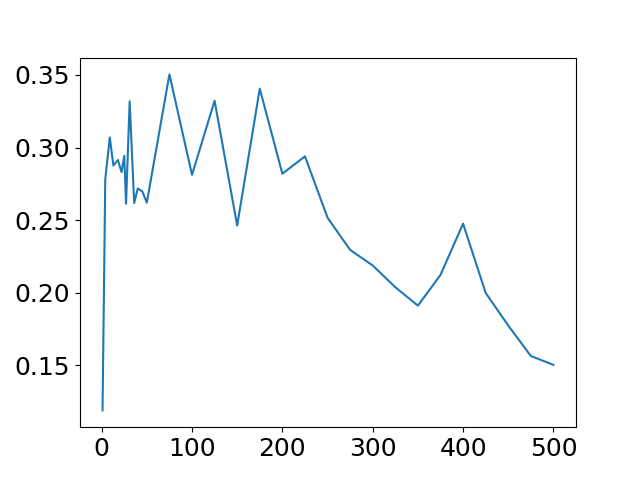}}
		\end{minipage} 
		
		\\ \hline
		
		l1 &  0.12422
		& 0.1237
		& 0.12296
		& 0.12306
		& 0.12421
		
		\\ \hline
		
		l2 &  0.10005
		& 0.09964
		& 0.09865
		& 0.09902
		& 0.09997

		\\ \hline
		
		p1 & 15.89995
		& 15.93068
		& 15.97068
		& 15.96469
		& 15.9007
		
		\\ \hline
		
		p2 & 17.22667
		& 17.25812
		& 17.32866
		& 17.30231
		& 17.22774
		
		\\ \hline
		
		pt & 5.30625
		& 4.94688
		& 5.38125
		& 6.59688
		& 4.7625
		
		\\ \hline
		
		pc & 13.80117
		& 14.42404
		& 15.23902
		& 16.47812
		& 13.74742
		
		\\ \hline
	\end{tabular}
\end{table}

Therefore, a better method to improve the training speed of the 3D Gaussian Splatting model \cite{kerbl20233d}, meanwhile keep the color rendering, is reducing the influence of Spherical Harmonics function. According to Equation \ref{eq8}, using more degree parameter $l$ and order parameter $m$ to combine more Spherical Harmonics coefficients can represent more abundant and detailed or exquisite color, but not the more will be the better. First, without other improvement strategies, we only change the max degree $self.sh\_degree$ to $0, 1, 2, 3$ and tested each performances. The default value is 3. The experiment results are shown in Table \ref{tb3}. Second, we changed the model code to delete some superfluous operations that are relative to the max degree in the model and set it as 0, we tested and gained a better result. Compared with the default one, directly changing the max degree will gain $16.25\%$ improvement. With the specific changing code, will gain $16.57\%$ improvement, but gain more improvement in CPU-GPU process time.

As shown in Table \ref{tb3}, setting the max degree of Spherical Harmonics coefficients as 0 will not make an explicit impact for the quality of the final rendering result. Actually, with enough experiments, we found that in the original code of 3D Gaussian Splatting model \cite{kerbl20233d}, the features, that are represented by the Spherical Harmonics coefficients which have degree that is more than 0,  of the color is unchanged during the all training, which means it will be not very useful if more than 0 degree Spherical Harmonics coefficients are used to represent the color. The following content will analyze and discuss more components of the model which are not very useful but can be omitted to improve the training speed of the model. We also applied this improvement strategy in other datasets and gained the similar conclusions. Each experiment record in this paper is the average of the values of 5 experiments that have the some criteria and conditions.

\begin{table}[h]
	\caption{ Comparison of Impacts Caused by Different Kinds of Backgrounds.}
	\label{tb4}
	\centering
	\begin{tabular}{  c | c | c | c | c | c | c  }
		\hline
		\textbf{it} & \textbf{white} & \textbf{black} & \textbf{random} & \textbf{0.1} & \textbf{0.2} & \textbf{0.3}
		\\ \hline
		ri & \begin{minipage}[b]{0.15\columnwidth}
			\centering
			\raisebox{-.2\height}{\includegraphics[width=\linewidth]{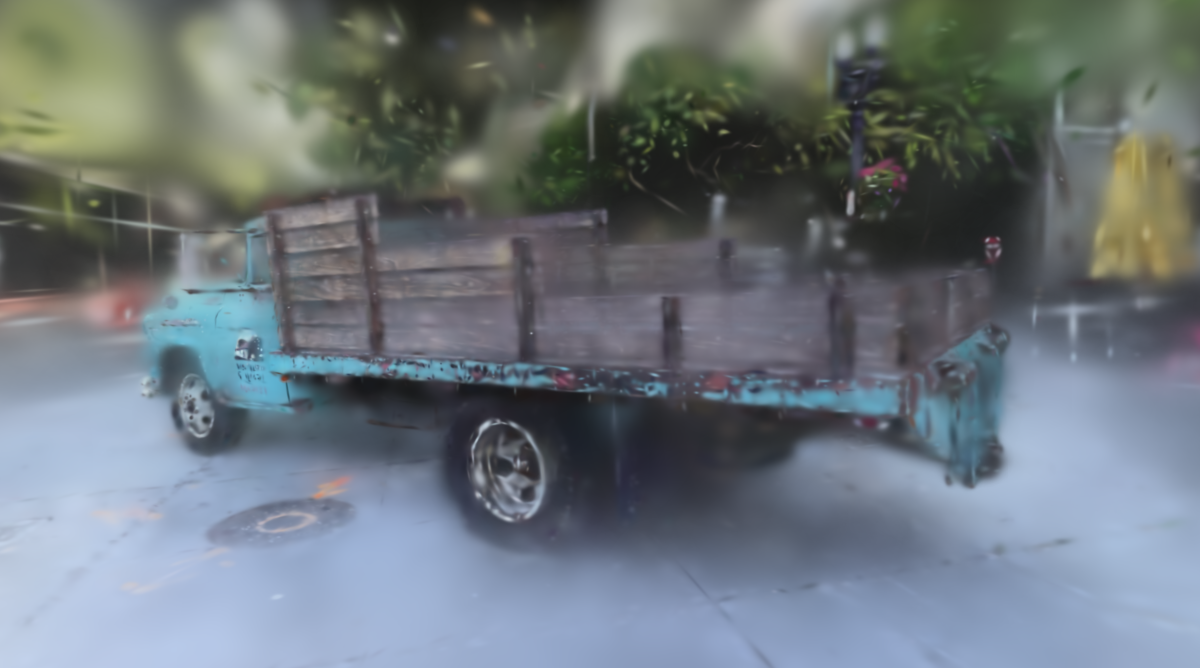}}
		\end{minipage}
		& \begin{minipage}[b]{0.15\columnwidth}
			\centering
			\raisebox{-.2\height}{\includegraphics[width=\linewidth]{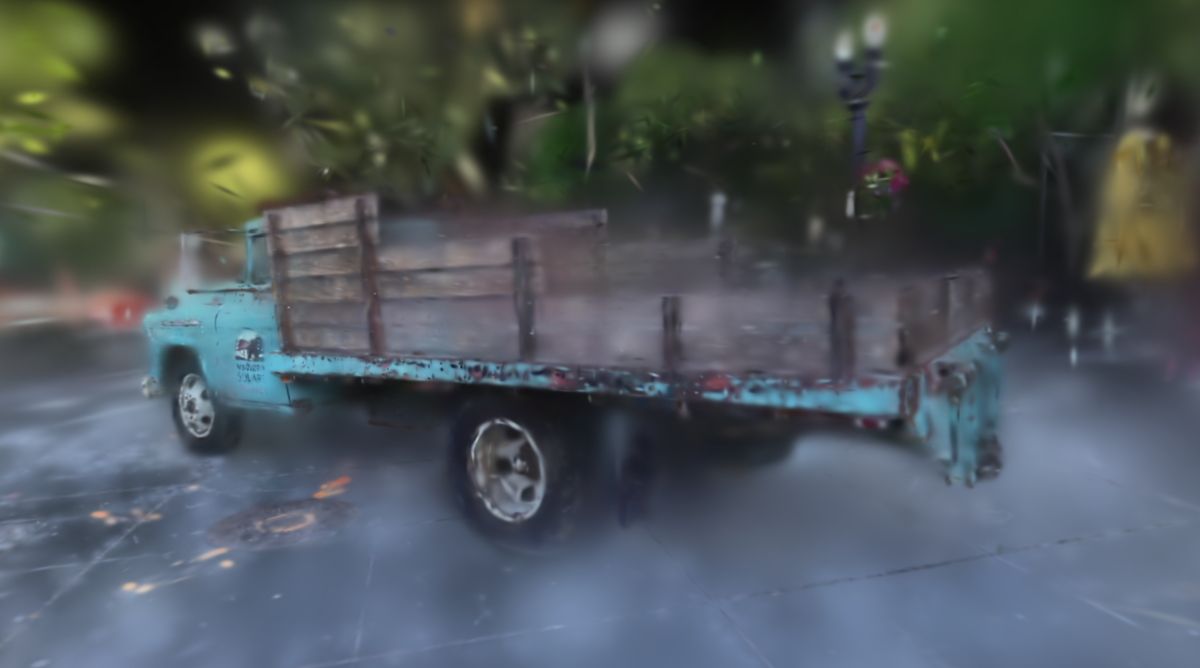}}
		\end{minipage}
		& \begin{minipage}[b]{0.15\columnwidth}
			\centering
			\raisebox{-.2\height}{\includegraphics[width=\linewidth]{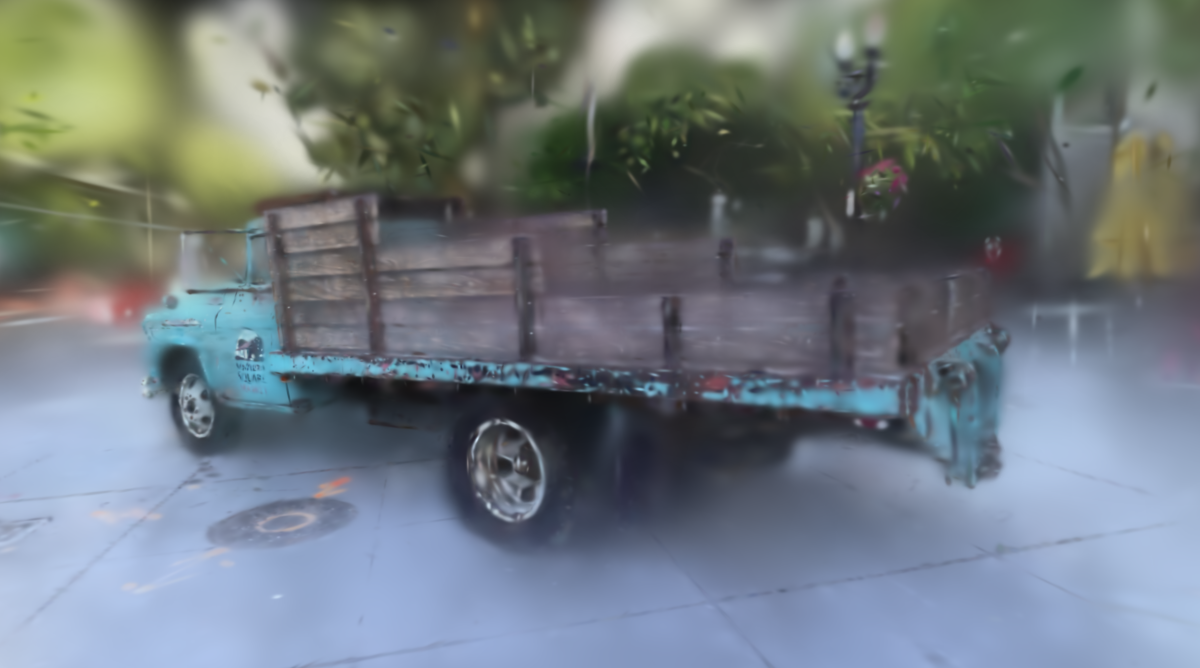}}
		\end{minipage} 
		& \begin{minipage}[b]{0.15\columnwidth}
			\centering
			\raisebox{-.2\height}{\includegraphics[width=\linewidth]{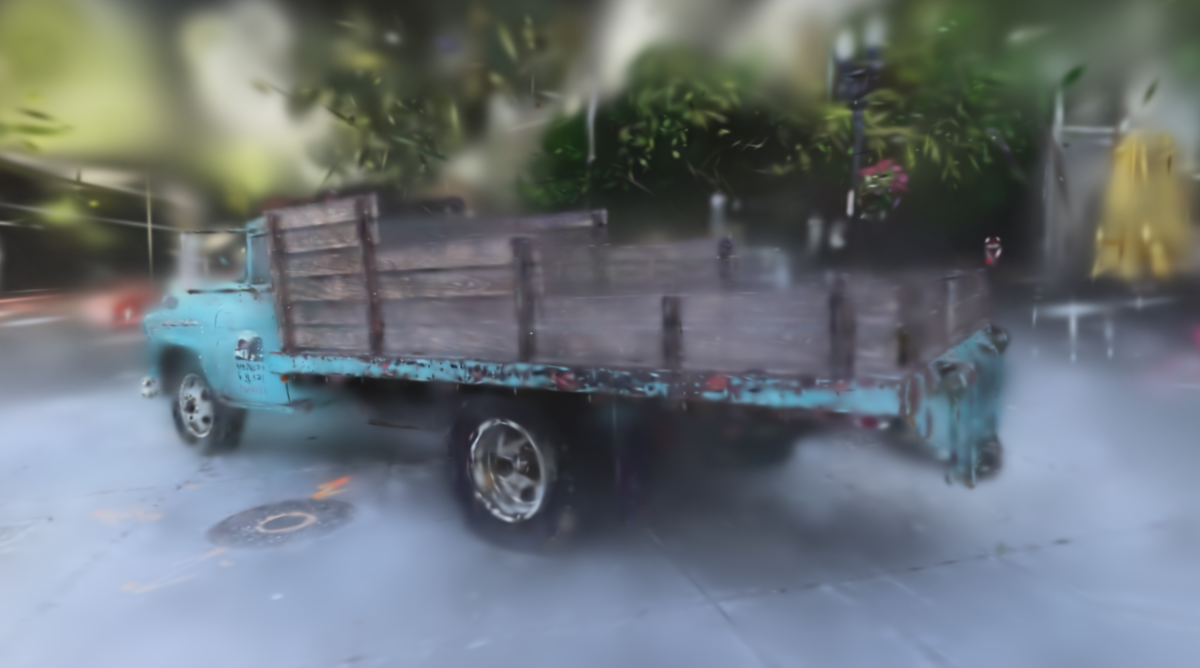}}
		\end{minipage} 
		& \begin{minipage}[b]{0.15\columnwidth}
			\centering
			\raisebox{-.2\height}{\includegraphics[width=\linewidth]{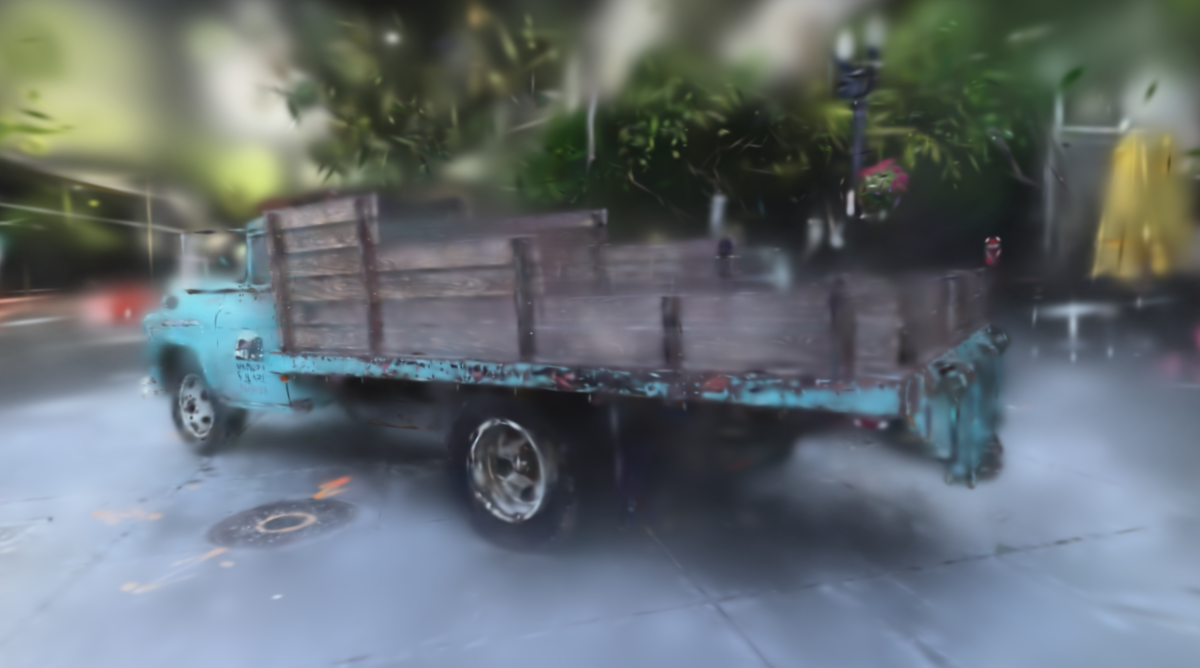}}
		\end{minipage} 
		& \begin{minipage}[b]{0.15\columnwidth}
			\centering
			\raisebox{-.2\height}{\includegraphics[width=\linewidth]{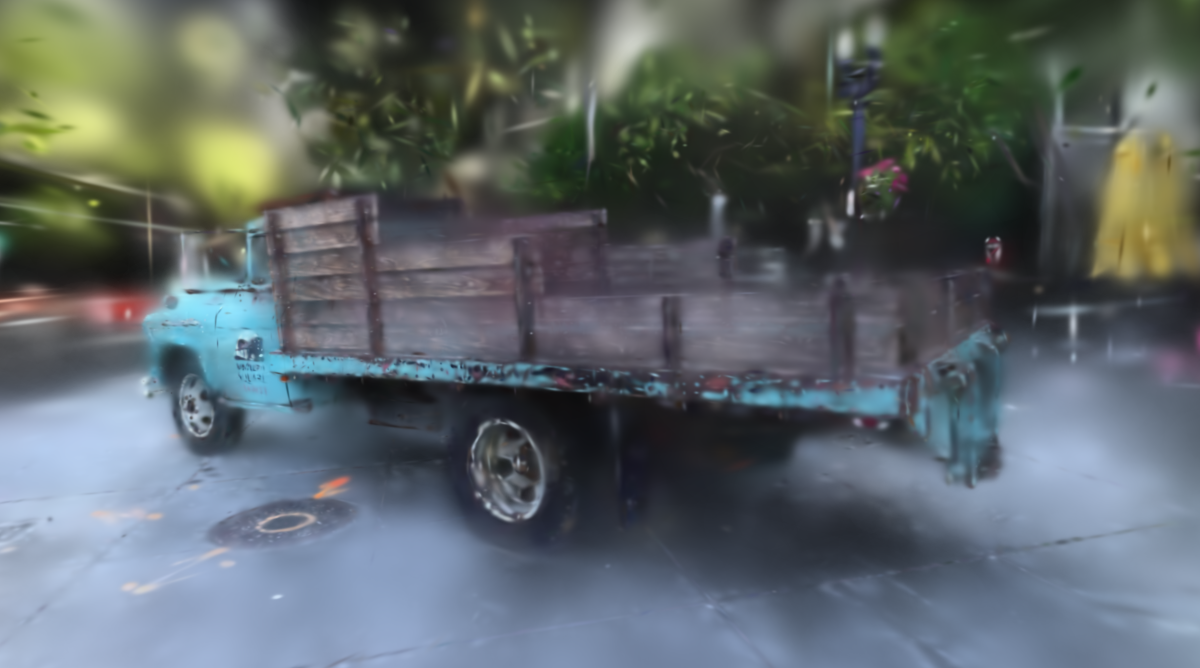}}
		\end{minipage}
		
		\\ \hline
		
		ls & \begin{minipage}[b]{0.15\columnwidth}
			\centering
			\raisebox{-.2\height}{\includegraphics[width=\linewidth]{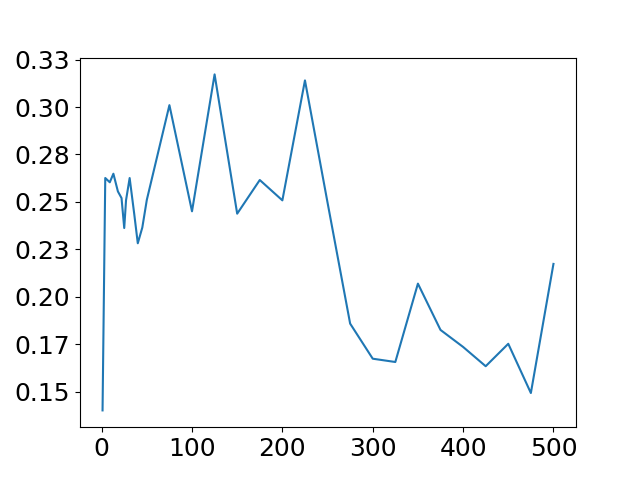}}
		\end{minipage}
		& \begin{minipage}[b]{0.15\columnwidth}
			\centering
			\raisebox{-.2\height}{\includegraphics[width=\linewidth]{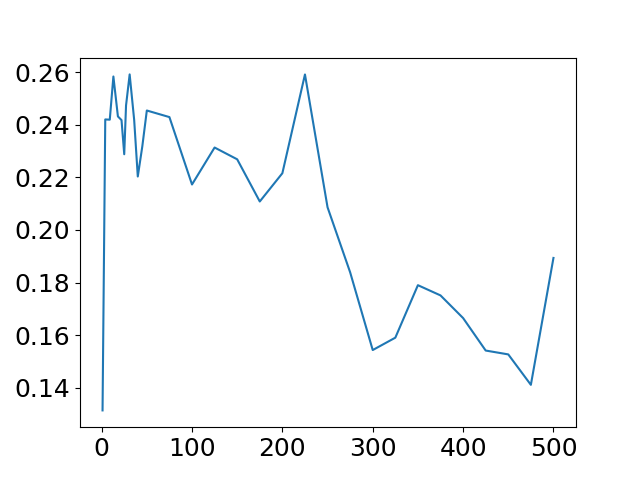}}
		\end{minipage}
		& \begin{minipage}[b]{0.15\columnwidth}
			\centering
			\raisebox{-.2\height}{\includegraphics[width=\linewidth]{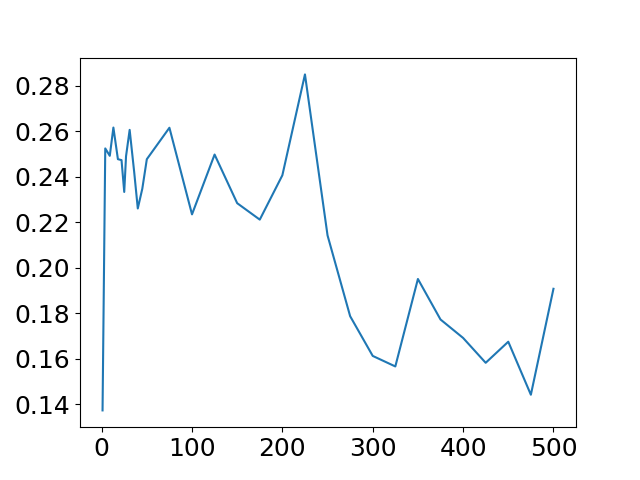}}
		\end{minipage} 
		& \begin{minipage}[b]{0.15\columnwidth}
			\centering
			\raisebox{-.2\height}{\includegraphics[width=\linewidth]{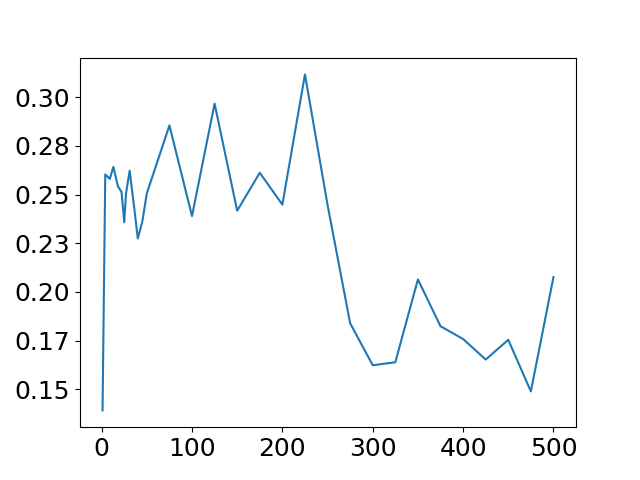}}
		\end{minipage} 
		& \begin{minipage}[b]{0.15\columnwidth}
			\centering
			\raisebox{-.2\height}{\includegraphics[width=\linewidth]{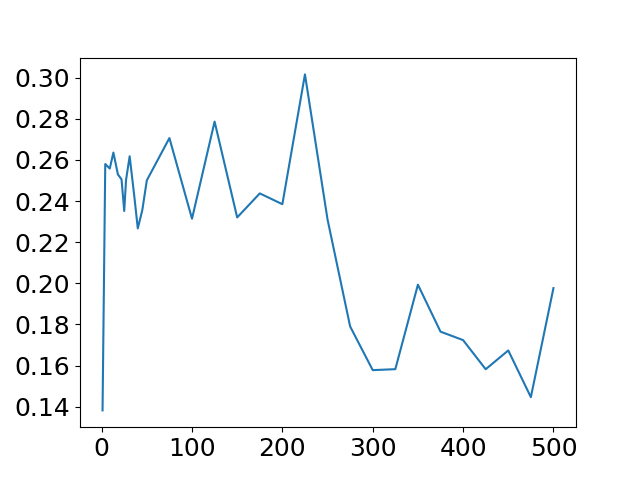}}
		\end{minipage} 
		& \begin{minipage}[b]{0.15\columnwidth}
			\centering
			\raisebox{-.2\height}{\includegraphics[width=\linewidth]{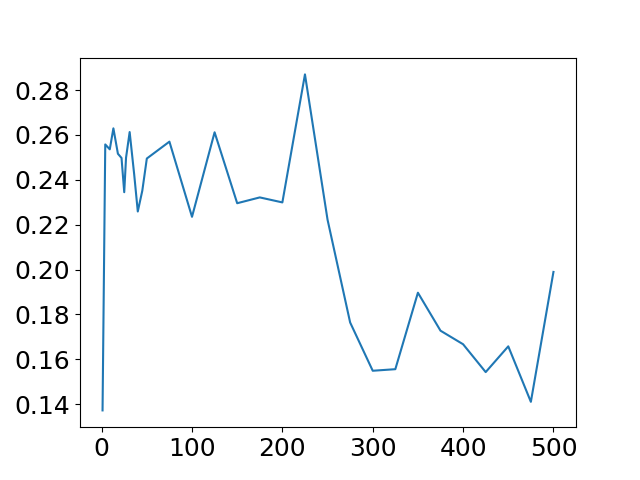}}
		\end{minipage}
		
		\\ \hline
		
		l1 &  0.14668
		& 0.14792
		& 0.15033
		& 0.14449
		& 0.14506
		& 0.14339
		
		
		
		\\ \hline
		
		p1 & 14.47329
		& 14.45789
		& 14.25963
		& 14.48396
		& 14.43938
		& 14.4773
		
		
		
		\\ \hline
		
		pt & 3.46875
		& 3.77812
		& 3.35938
		& 3.32188
		& 2.11875
		& 2.44062
		
		\\ \hline
		
		pc & 11.4901
		& 11.92386
		& 12.29908
		& 10.56013
		& 9.76496
		& 9.36328
		
		\\ \hline
		\textbf{it} & \textbf{0.4} & \textbf{0.5} & \textbf{0.6} & \textbf{0.7} & \textbf{0.8} & \textbf{0.9}
		\\ \hline
		ri & \begin{minipage}[b]{0.15\columnwidth}
			\centering
			\raisebox{-.2\height}{\includegraphics[width=\linewidth]{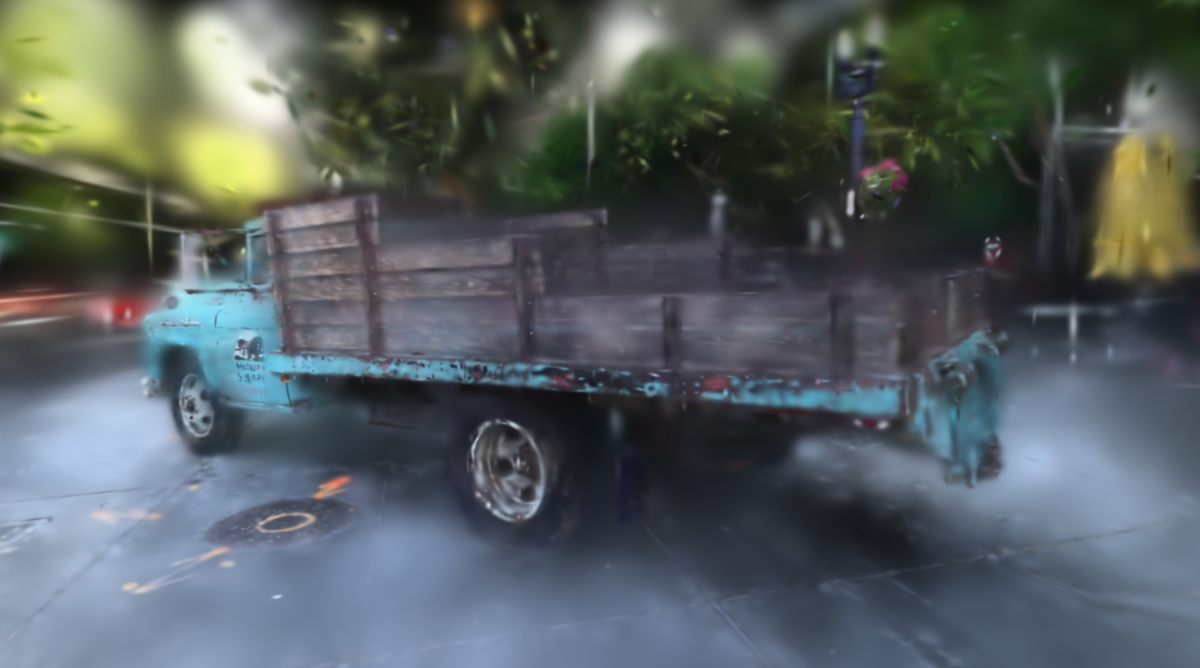}}
		\end{minipage}
		& \begin{minipage}[b]{0.15\columnwidth}
			\centering
			\raisebox{-.2\height}{\includegraphics[width=\linewidth]{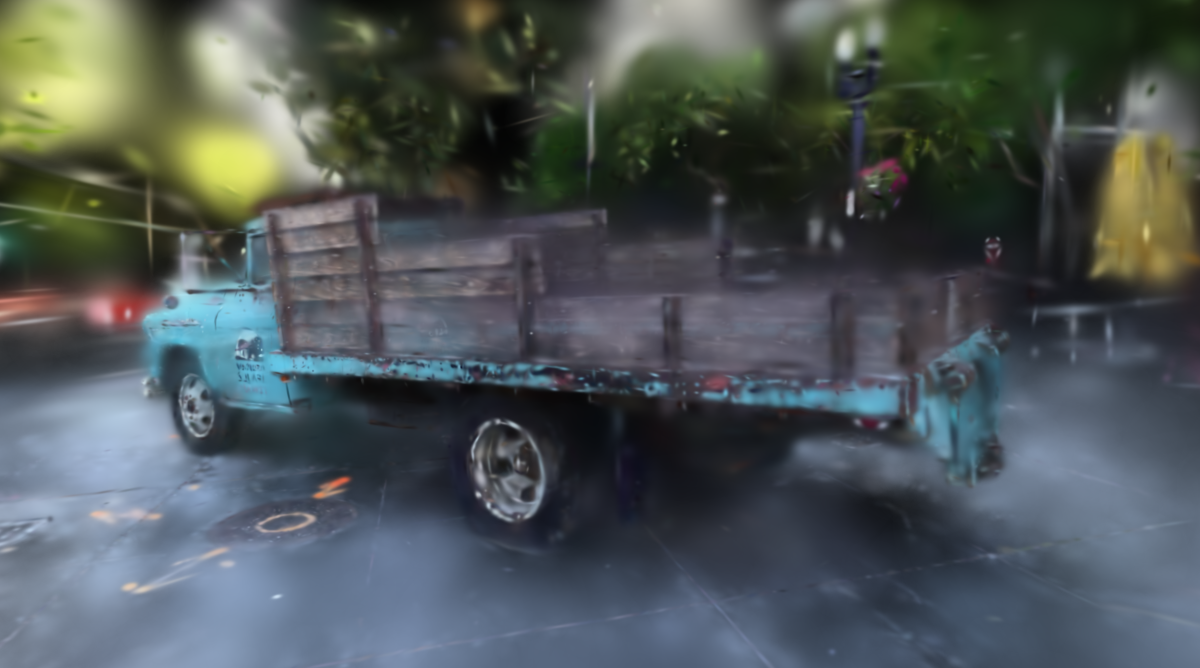}}
		\end{minipage}
		& \begin{minipage}[b]{0.15\columnwidth}
			\centering
			\raisebox{-.2\height}{\includegraphics[width=\linewidth]{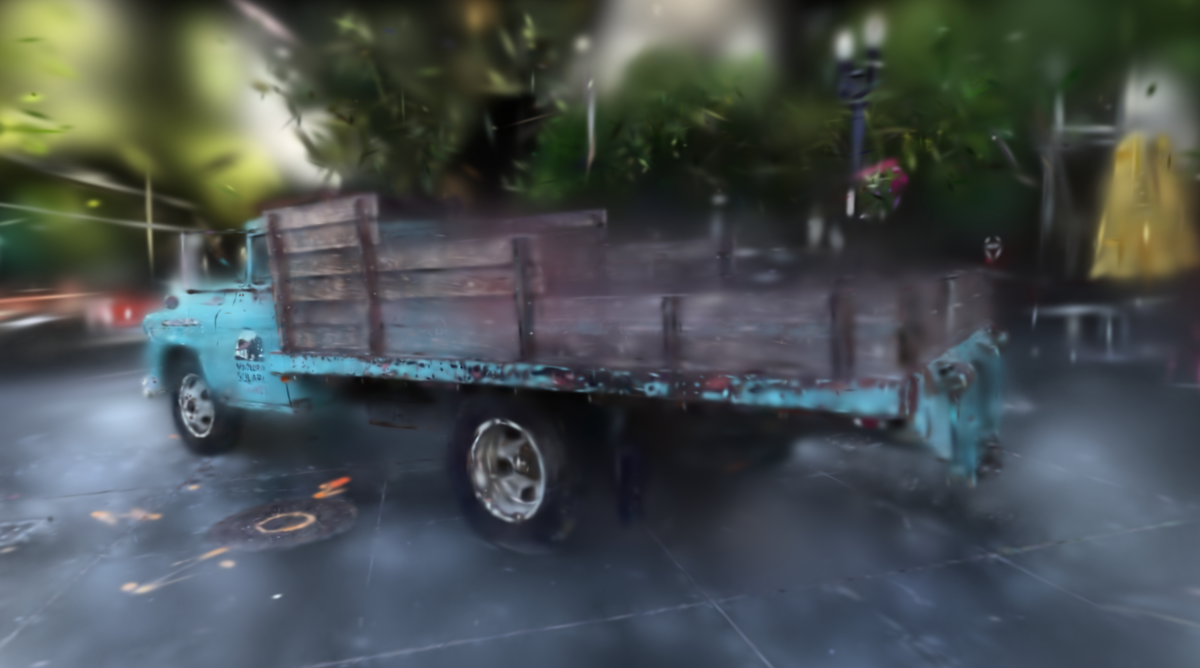}}
		\end{minipage} 
		& \begin{minipage}[b]{0.15\columnwidth}
			\centering
			\raisebox{-.2\height}{\includegraphics[width=\linewidth]{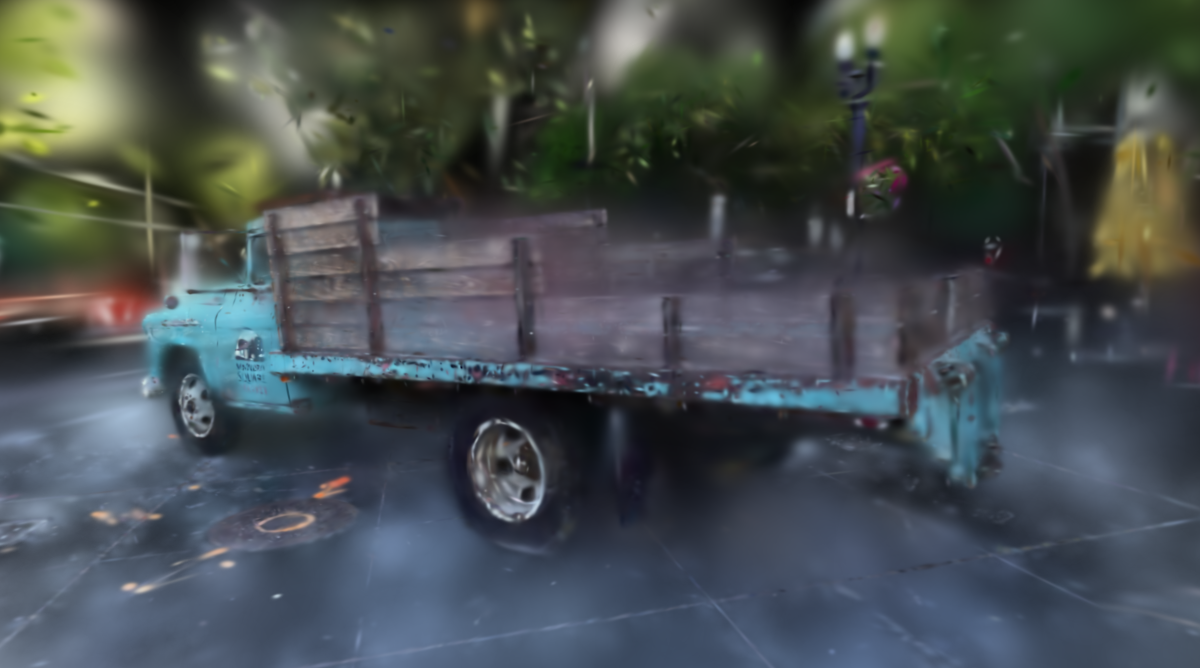}}
		\end{minipage} 
		& \begin{minipage}[b]{0.15\columnwidth}
			\centering
			\raisebox{-.2\height}{\includegraphics[width=\linewidth]{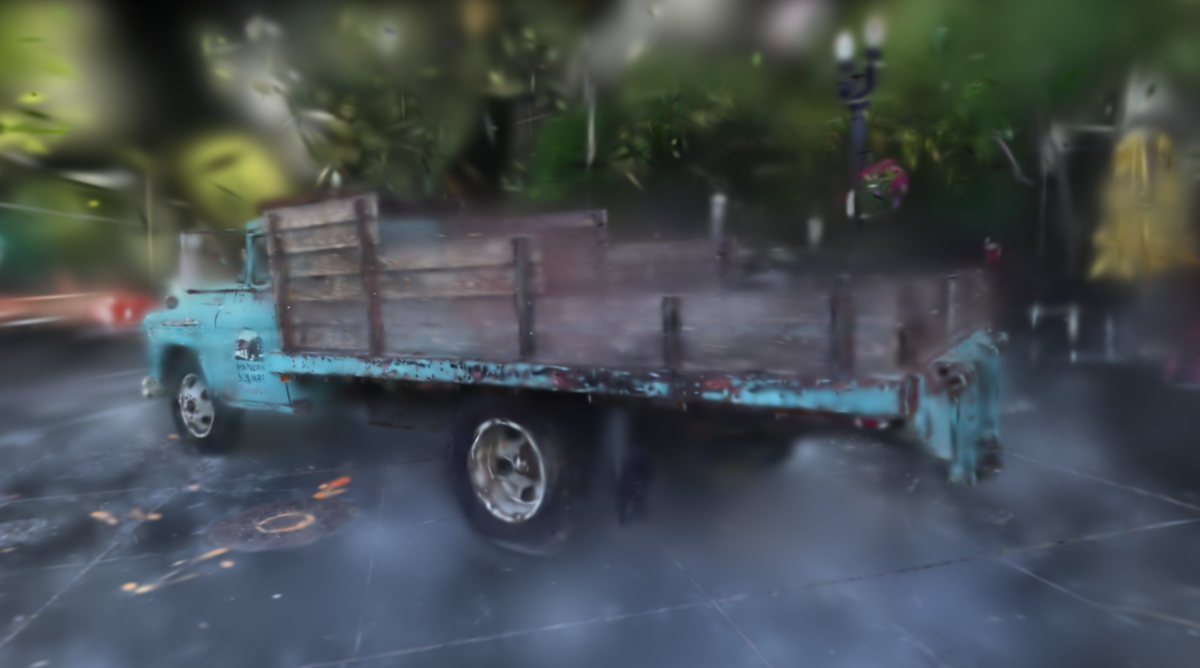}}
		\end{minipage} 
		& \begin{minipage}[b]{0.15\columnwidth}
			\centering
			\raisebox{-.2\height}{\includegraphics[width=\linewidth]{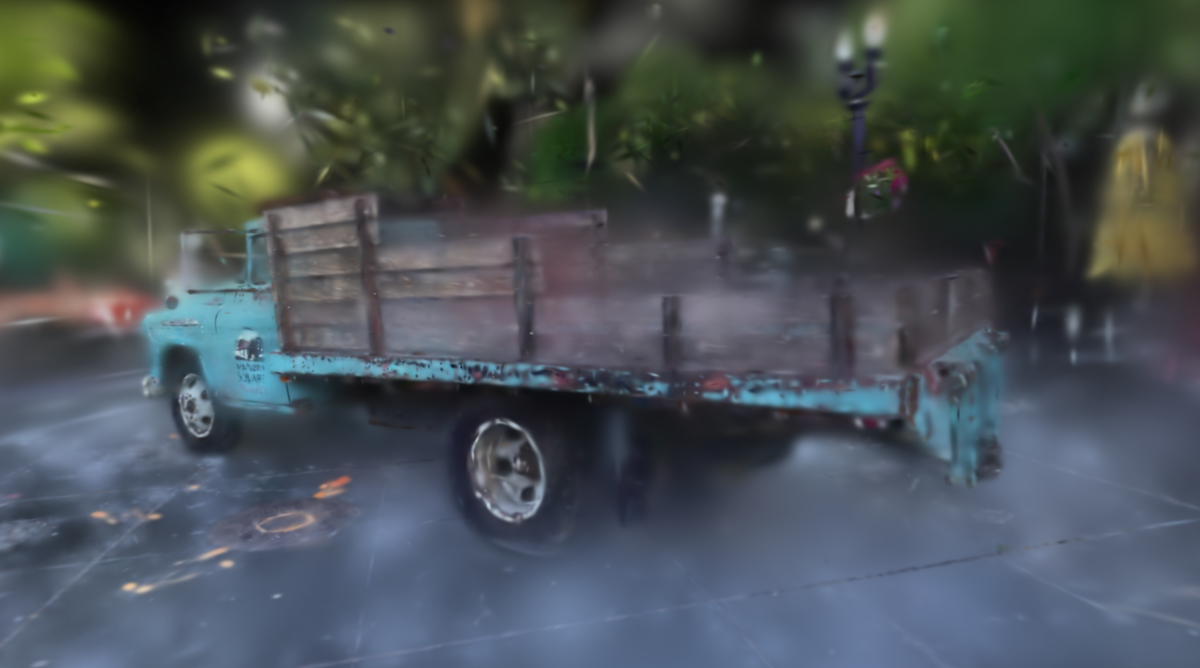}}
		\end{minipage}
		
		\\ \hline
		
		ls & \begin{minipage}[b]{0.15\columnwidth}
			\centering
			\raisebox{-.2\height}{\includegraphics[width=\linewidth]{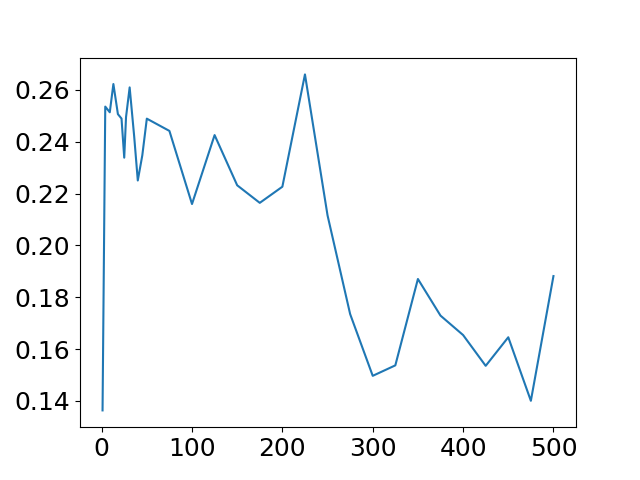}}
		\end{minipage}
		& \begin{minipage}[b]{0.15\columnwidth}
			\centering
			\raisebox{-.2\height}{\includegraphics[width=\linewidth]{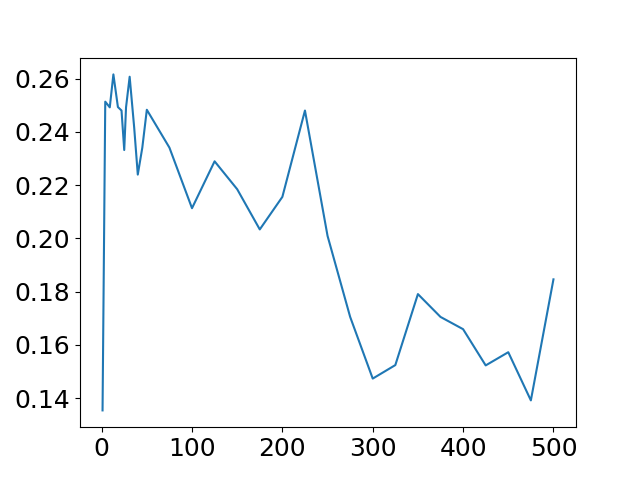}}
		\end{minipage}
		& \begin{minipage}[b]{0.15\columnwidth}
			\centering
			\raisebox{-.2\height}{\includegraphics[width=\linewidth]{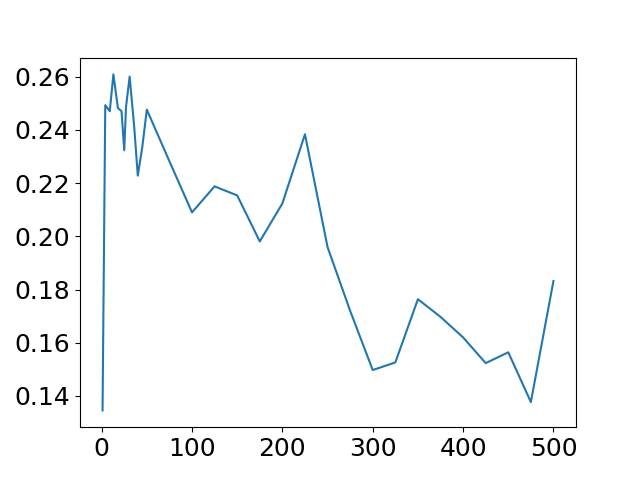}}
		\end{minipage} 
		& \begin{minipage}[b]{0.15\columnwidth}
			\centering
			\raisebox{-.2\height}{\includegraphics[width=\linewidth]{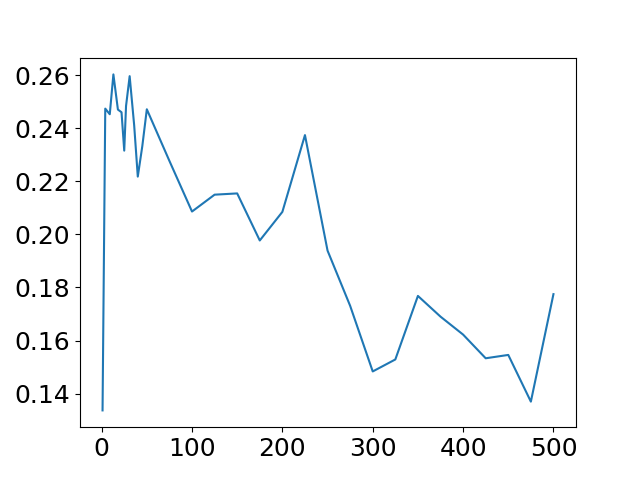}}
		\end{minipage} 
		& \begin{minipage}[b]{0.15\columnwidth}
			\centering
			\raisebox{-.2\height}{\includegraphics[width=\linewidth]{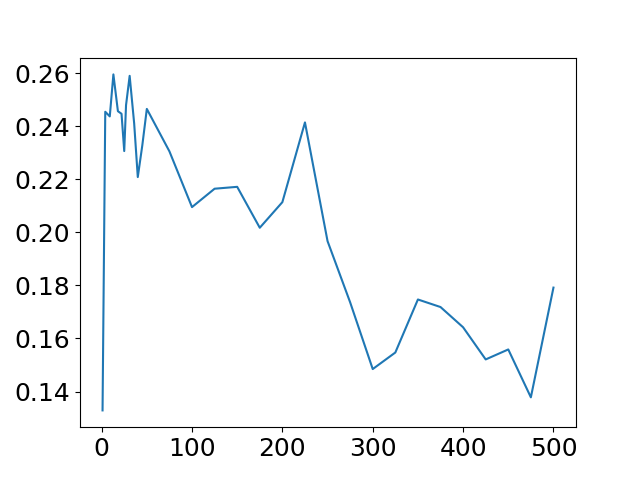}}
		\end{minipage} 
		& \begin{minipage}[b]{0.15\columnwidth}
			\centering
			\raisebox{-.2\height}{\includegraphics[width=\linewidth]{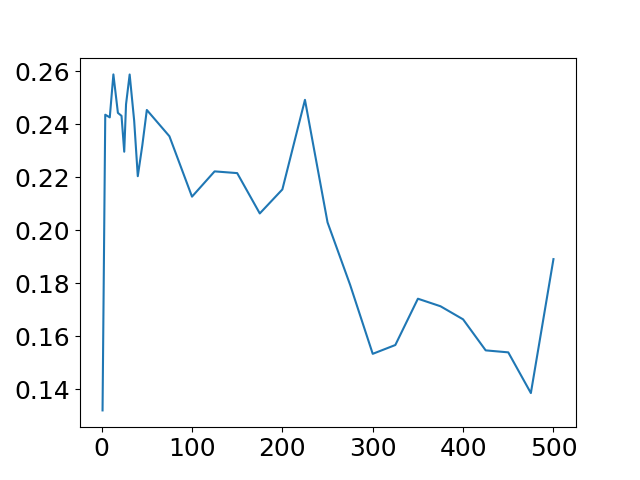}}
		\end{minipage}
		
		\\ \hline
		
		l1 &  0.14037
		& 0.13457
		& 0.13877
		& 0.14155
		& 0.14224
		& 0.14546
		
		
		
		\\ \hline
		
		p1 & 14.67536
		& 15.03836
		& 14.84794
		& 14.77118
		& 14.78867
		& 14.60945
		
		
		
		\\ \hline
		
		pt & 2.15937
		& 1.98125
		& 2.99375
		& 2.35
		& 2.87812
		& 3.5125
		
		\\ \hline
		
		pc & 8.97939
		& 9.04309
		& 9.2999
		& 9.84195
		& 10.46174
		& 11.08121
		
		\\ \hline

	\end{tabular}
\end{table}

The composition of the background in the 3D Gaussian Splatting model \cite{kerbl20233d} is also an important influence factor for the training and rendering speed. Without other improvement strategies, only changing the background. The training steps is 500, the training object is the truck dataset. As shown in Table \ref{tb4}, the default background of the model is set as white, which means that the three values of the three channels respectively of RGB of the background are all set as 1.0. We set the three values of the background as the same other values and made the corresponding experiments. With different datasets for test and evaluations, we found that a better valuing range for the background is usually $0.3\le bg \le0.5$, $bg$ represents the three same values of background color. As shown in Table \ref{tb4}, the $pc$ time when $bg=0.4$ has about $21.85\%$ than the default one (i.e.: $bg$=white). It can be also found that, during the changing of the background, that the transformations of the learning rate, training loss and PSNR are remain a good state. Therefore, changing the background is a feasible improvement strategy.

\subsection{Learning and Structure}

Subsequently, we will discuss about the feasible improvement strategies on the level of learning and structure of the 3D Gaussian Splatting model. Because the 3D Gaussian Splatting model has built an individual 3D Gaussian expression for each point of the input data of the model, the density of 3D Gaussian points will impact the rendering speed naturally, which means the sampling rate will influence the training speed. Normally, reducing the sampling rate will increasing the training speed. As shown in Table \ref{tb5}, we controlled the sampling rate by controlling the value of the sampling gap (i.e.:2, 4, 8, 20, 50, 500). The training steps is 3000. The bigger the sampling gap is, the lower the the sampling rate is, which is a  inversely proportional relationship. As the sampling gap being bigger, the $pc$ time has become increased, but  the rendering result has becoming blurrier. Considering the rendering performance and change, the improvement brought by this strategy in training speed should be careful considered.

\begin{table}[h]
	\caption{ Comparison of Impacts Caused by Different Points Sampling Gap.}
	\label{tb5}
	\centering
	\begin{tabular}{  c | c | c | c | c | c | c  }
		\hline
		\textbf{it} & \textbf{2} & \textbf{4} & \textbf{8} & \textbf{20} & \textbf{50} & \textbf{500}
		\\ \hline
		ri & \begin{minipage}[b]{0.15\columnwidth}
			\centering
			\raisebox{-.2\height}{\includegraphics[width=\linewidth]{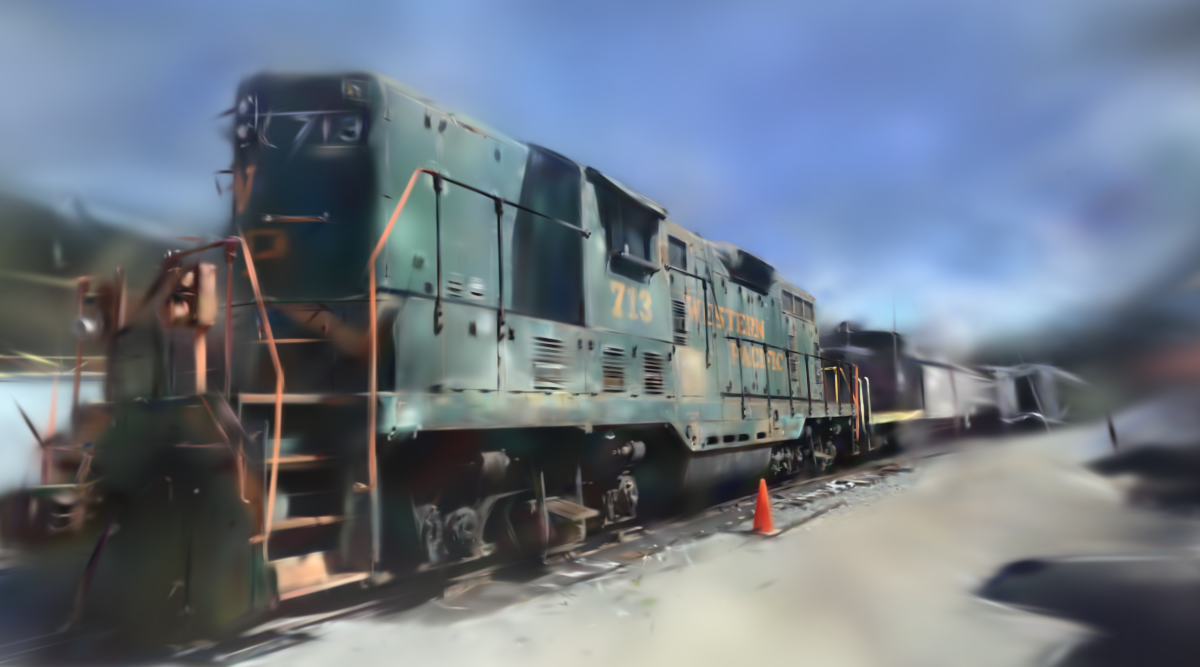}}
		\end{minipage}
		& \begin{minipage}[b]{0.15\columnwidth}
			\centering
			\raisebox{-.2\height}{\includegraphics[width=\linewidth]{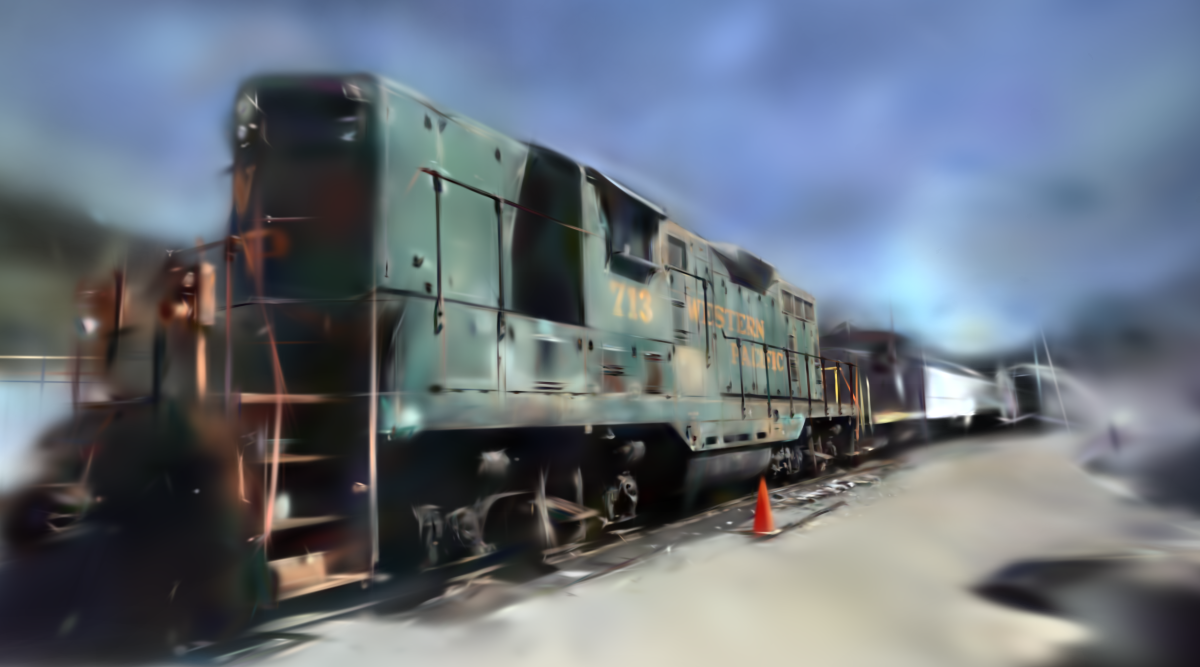}}
		\end{minipage}
		& \begin{minipage}[b]{0.15\columnwidth}
			\centering
			\raisebox{-.2\height}{\includegraphics[width=\linewidth]{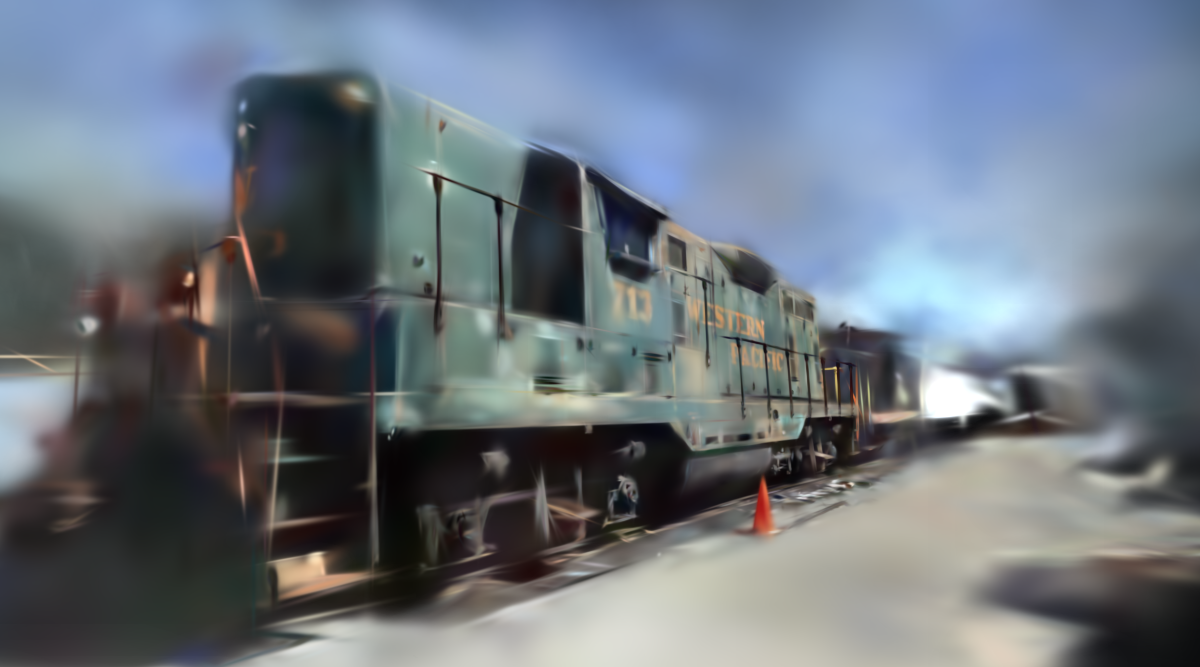}}
		\end{minipage} 
		& \begin{minipage}[b]{0.15\columnwidth}
			\centering
			\raisebox{-.2\height}{\includegraphics[width=\linewidth]{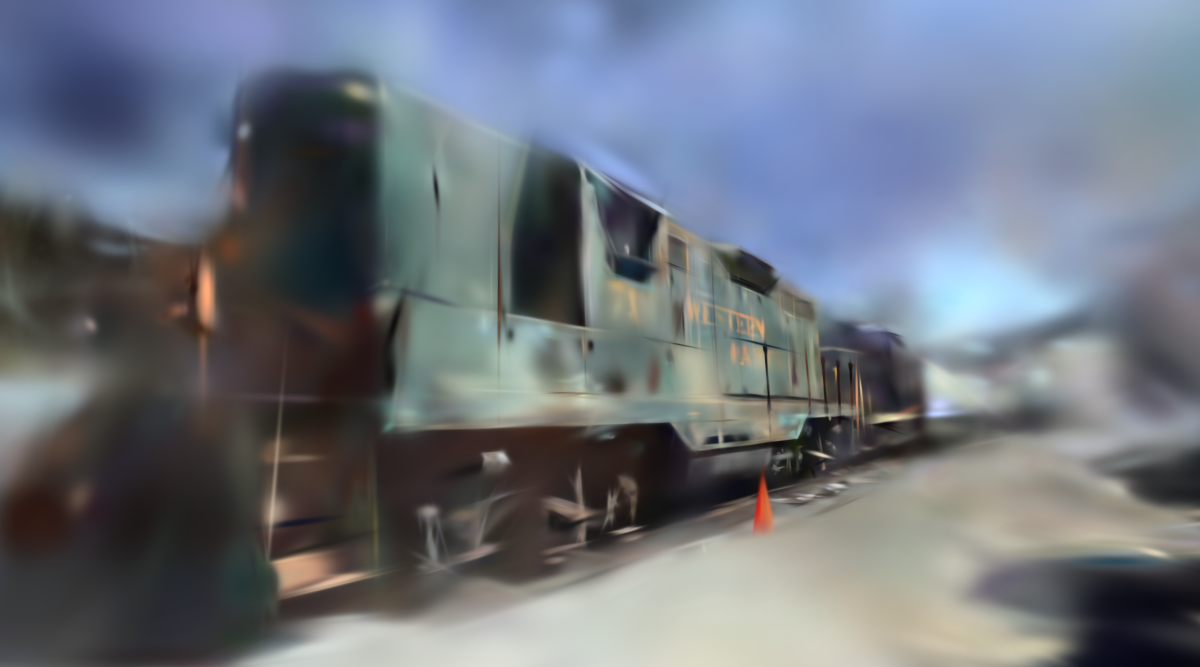}}
		\end{minipage} 
		& \begin{minipage}[b]{0.15\columnwidth}
			\centering
			\raisebox{-.2\height}{\includegraphics[width=\linewidth]{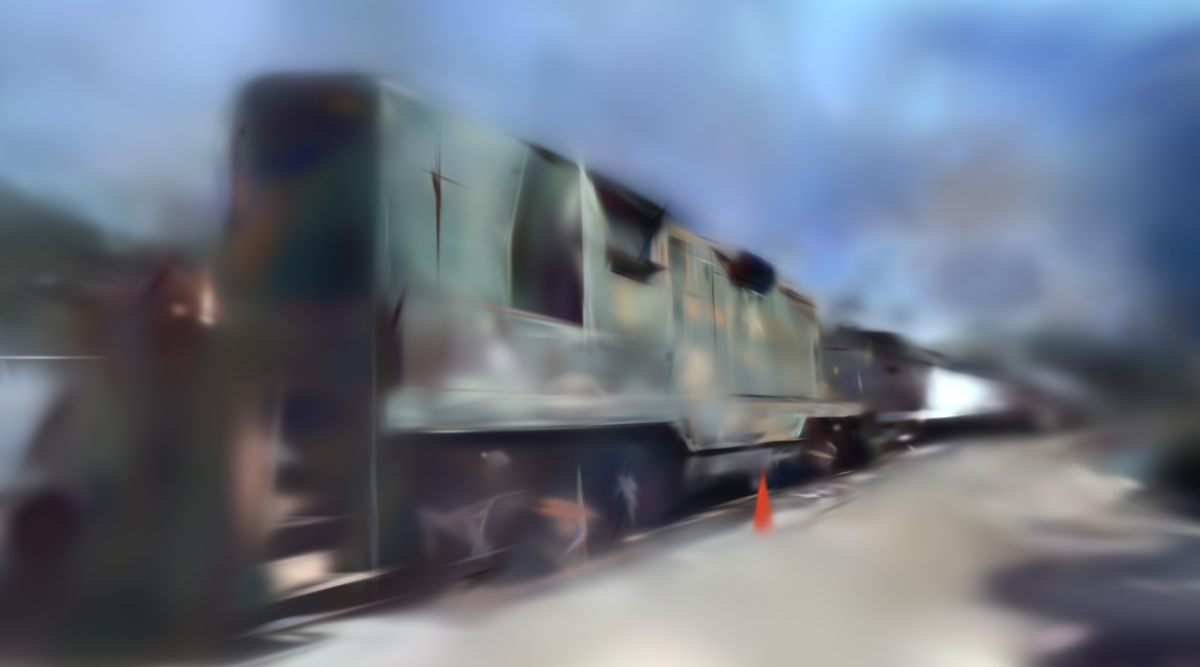}}
		\end{minipage} 
		& \begin{minipage}[b]{0.15\columnwidth}
			\centering
			\raisebox{-.2\height}{\includegraphics[width=\linewidth]{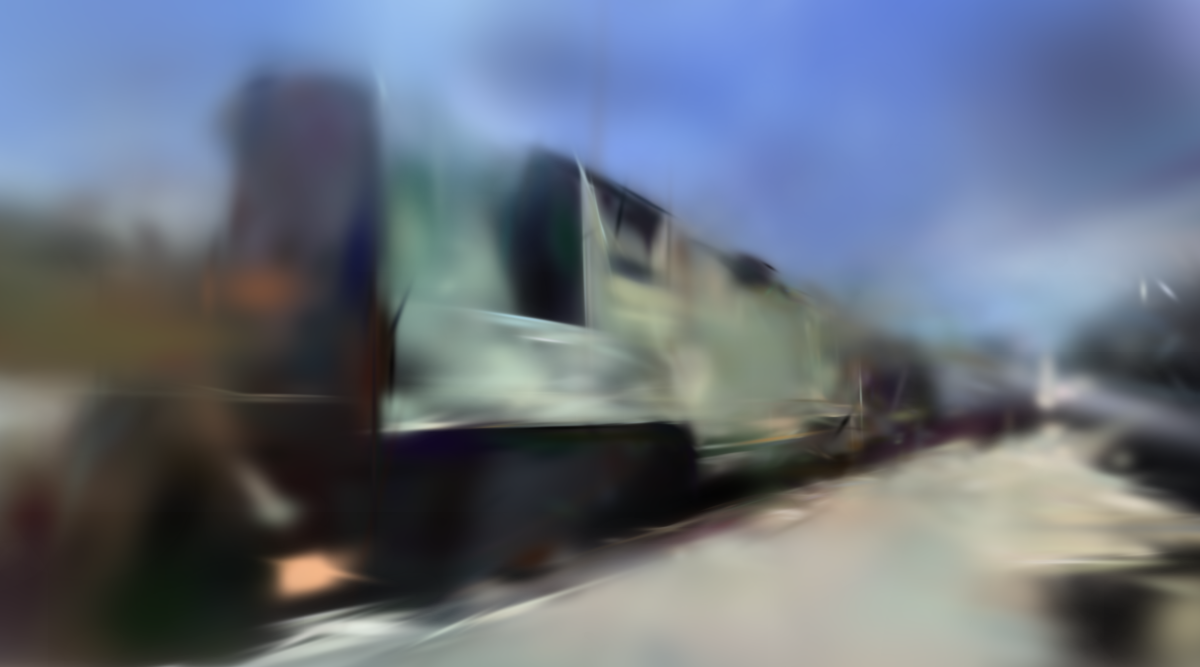}}
		\end{minipage}
		
		\\ \hline
		
		ls & \begin{minipage}[b]{0.15\columnwidth}
			\centering
			\raisebox{-.2\height}{\includegraphics[width=\linewidth]{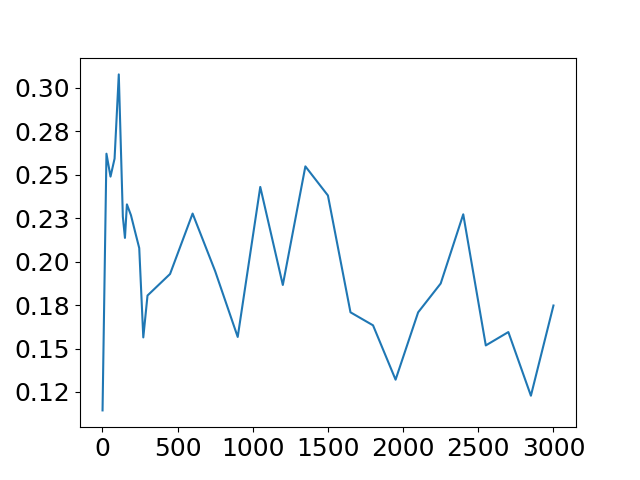}}
		\end{minipage}
		& \begin{minipage}[b]{0.15\columnwidth}
			\centering
			\raisebox{-.2\height}{\includegraphics[width=\linewidth]{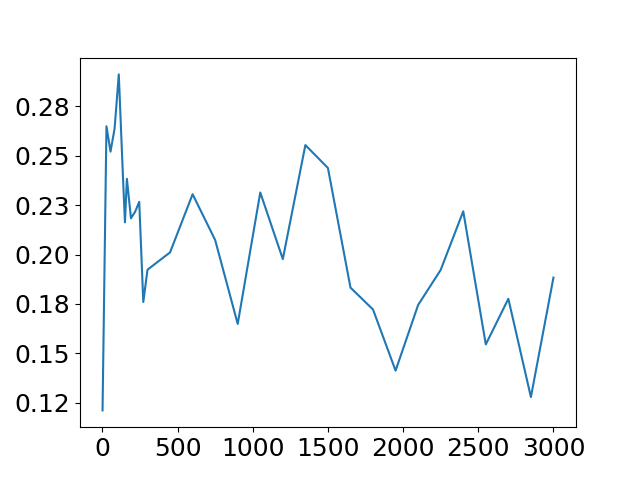}}
		\end{minipage}
		& \begin{minipage}[b]{0.15\columnwidth}
			\centering
			\raisebox{-.2\height}{\includegraphics[width=\linewidth]{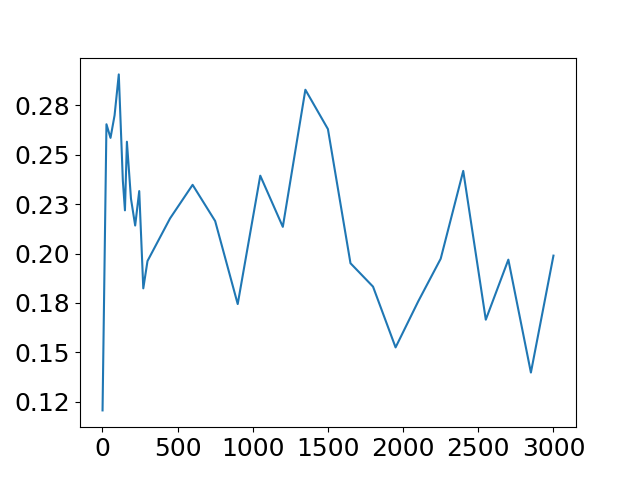}}
		\end{minipage} 
		& \begin{minipage}[b]{0.15\columnwidth}
			\centering
			\raisebox{-.2\height}{\includegraphics[width=\linewidth]{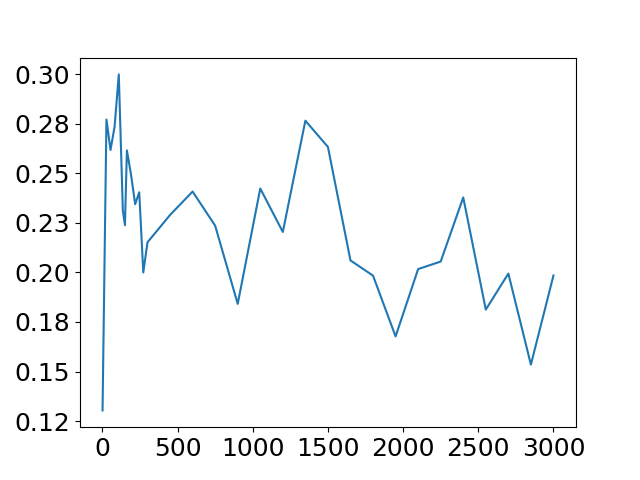}}
		\end{minipage} 
		& \begin{minipage}[b]{0.15\columnwidth}
			\centering
			\raisebox{-.2\height}{\includegraphics[width=\linewidth]{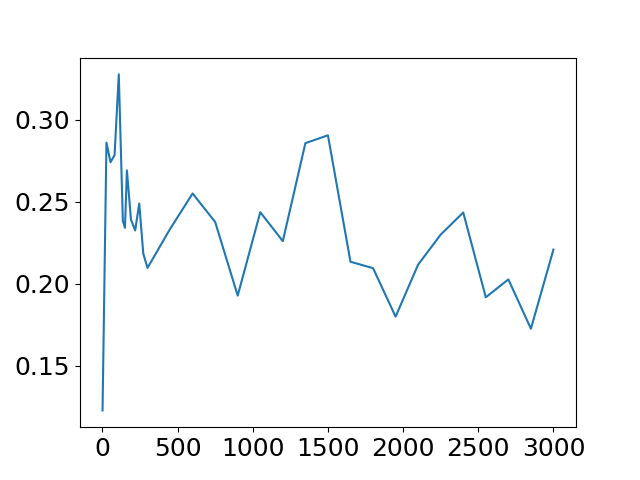}}
		\end{minipage} 
		& \begin{minipage}[b]{0.15\columnwidth}
			\centering
			\raisebox{-.2\height}{\includegraphics[width=\linewidth]{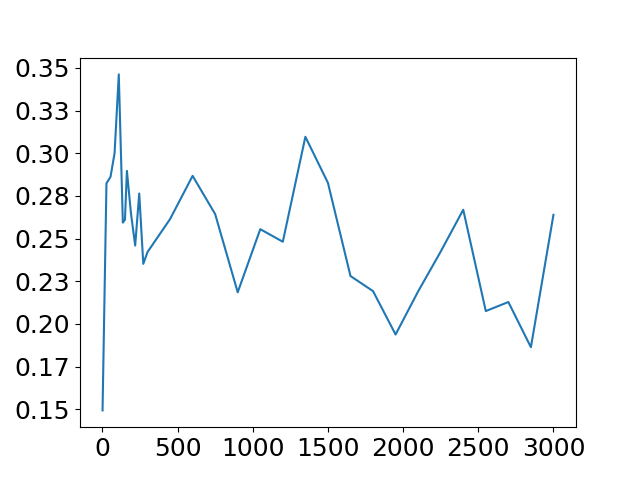}}
		\end{minipage}
		
		\\ \hline
		
		l1 &  0.17739
		& 0.17766
		& 0.18196
		& 0.185
		& 0.19744
		& 0.20525
		
		
		
		\\ \hline
		
		p1 & 12.88219
		& 12.85933
		& 12.66082
		& 12.50156
		& 12.00996
		& 11.66958
		
		
		
		\\ \hline
		
		pt & 12.36562
		& 10.06563
		& 9.48125
		& 9.69688
		& 9.29688
		& 8.8125
		
		\\ \hline
		
		pc & 53.26753
		& 51.09732
		& 48.89431
		& 47.34717
		& 46.46798
		& 45.53414
		\\ \hline
	\end{tabular}
\end{table}

The learning rate of the space location of 3D Gaussian points is also an important part in the original model \cite{kerbl20233d}. As shown in Table \ref{tb6}, without other improvement strategies, we used the a series of fixed $xyz$ learning rate (e.g.: 1e-i, i=1, .., 10), to test the training speed and found that, when the learning rate $lr$ meets the range $lr \in [1e-2, 1]$, the training speed will be improved in a big level but the rendering quality will reduce quickly, which means the adjustment here requires a balanced consideration. As shown in Table \ref{tb6}, without other improvement strategies, let training steps as 500, with the calculation formula that are shown as Equation \ref{eq11}, implemented the experiments for other types of changeable learning rates.

\begin{equation}  
	\label{eq11}
	\left\{  
	\begin{array}{lr}  
		rw(x) = \left\{  
		\begin{array}{lr}  
			a1, x \in [0, p1]    \\
			a2, x \in [p1, p2]   \\
			a3, x \in [p2, 1]   \\
		\end{array}  
		\right.   \\
		s(t) = \frac{a_{s}}{2}sin(b_{s}\frac{t}{T}2\pi + \frac{\pi}{2}) + 1   \\
		c(t) = \frac{a_{c}}{2}cos(b_{c}\frac{t}{T}2\pi + {\pi}) + 1   \\
		\frac{l(t)-p_{2}}{p_{1}-p_{2}} = \frac{t-t_{2}}{t_{1}-t_{2}}; E(t) = \varepsilon e^{\delta t}+\eta   \\
	\end{array}  
	\right.  
\end{equation}

As shown in Equation \ref{eq11} and Table \ref{tb6}, the $rw0-2$ represents a two-stage rectangular wave transformation for the $xyz$ learning rate. The first stage includes $0\%-20\%$ of the iterations, the second stage contains the $20\%-100\%$ iterations, etc. Here, $a_{1}=1e-3, a_{2}=1e-6, a_{3}=1e-9$. The function $s(t)$ represents the learning rate is being changing following the iteration $t$ based on the sine function, which has the similar function $c(t)$ to define the changing that is based on the cosine function. In the $it$ row, the parameter of $s(t)$ means $a_{s}, b_{s}$, the parameter of $c(t)$ means $a_{c},b_{c}$. Parameters $a_{s}, b_{s}, a_{c}, b_{c}$ represent the corresponding fining coefficients. $T$ represents the number of all of the iterations, which is fixed and different for different set of input model data. We used the two-point formula for a linear equation to describe and define the linear changing of $xyz$ learning, which passed two points $(t_{1}, p_{1})$ and $(t_{2}, p_{2})$, which has the parameter $(p_{1}, p_{2})$ and meets $t_{1}=1, t_{2}=iterations$. $E(t)$ is used to present the exponential changing learning rate. $\varepsilon, \delta, \eta$ are all fining coefficients. $E(t)$ passes the points $(t_{1}, E(t_{1}))$ and $(t_{2}, E(t_{2}))$, which can gains $\varepsilon = \frac{exp(t_{1})-exp(t_{2})}{e^{t_{1}}-e^{t_{2}}}$. $it$ row box means $(E(t_{1}), E(t_{2}))$ and meets $t_{1}=1, t_{2}=iterations$.

\begin{table}[h]
	\caption{ Impacts by Different Alterable xyz Learning Rates with Kinds of Change.}
	\label{tb6}
	\centering
	\begin{tabular}{  c | c | c | c | c | c | c  }
		\hline
		\textbf{it} & \textbf{original} & \textbf{1e-1} & \textbf{1e-2} & \textbf{rw0-2} & \textbf{rw0-2-4} & \textbf{rw0-3-6}
		\\ \hline
		ri & \begin{minipage}[b]{0.15\columnwidth}
			\centering
			\raisebox{-.2\height}{\includegraphics[width=\linewidth]{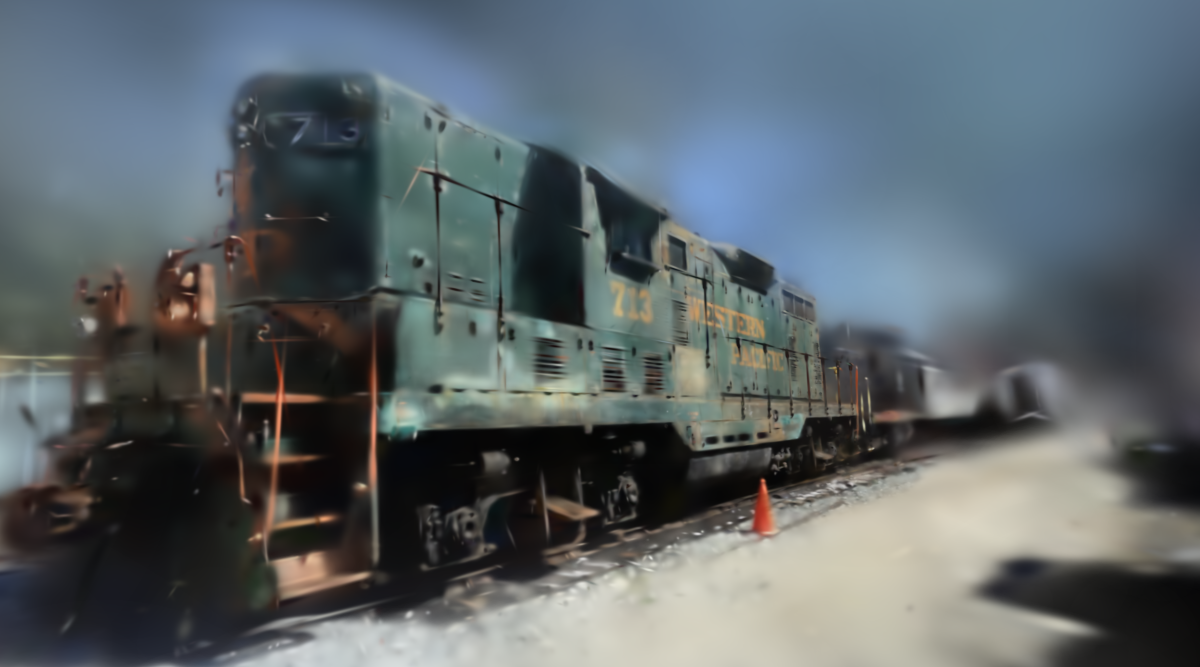}}
		\end{minipage}
		& \begin{minipage}[b]{0.15\columnwidth}
			\centering
			\raisebox{-.2\height}{\includegraphics[width=\linewidth]{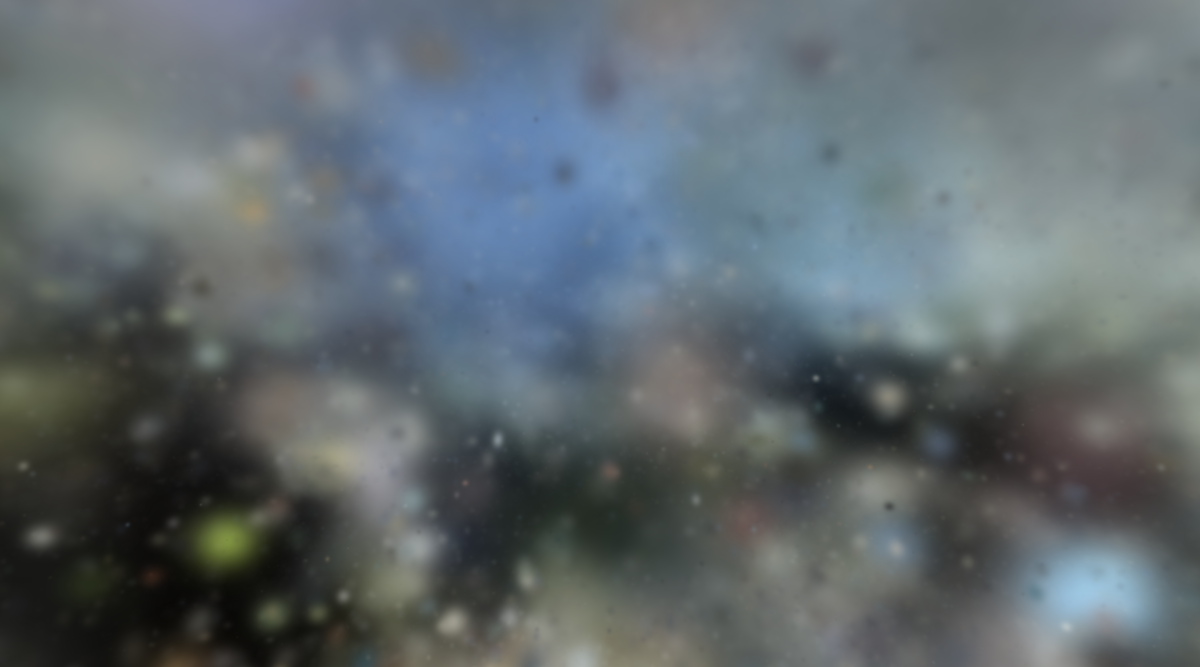}}
		\end{minipage}
		& \begin{minipage}[b]{0.15\columnwidth}
			\centering
			\raisebox{-.2\height}{\includegraphics[width=\linewidth]{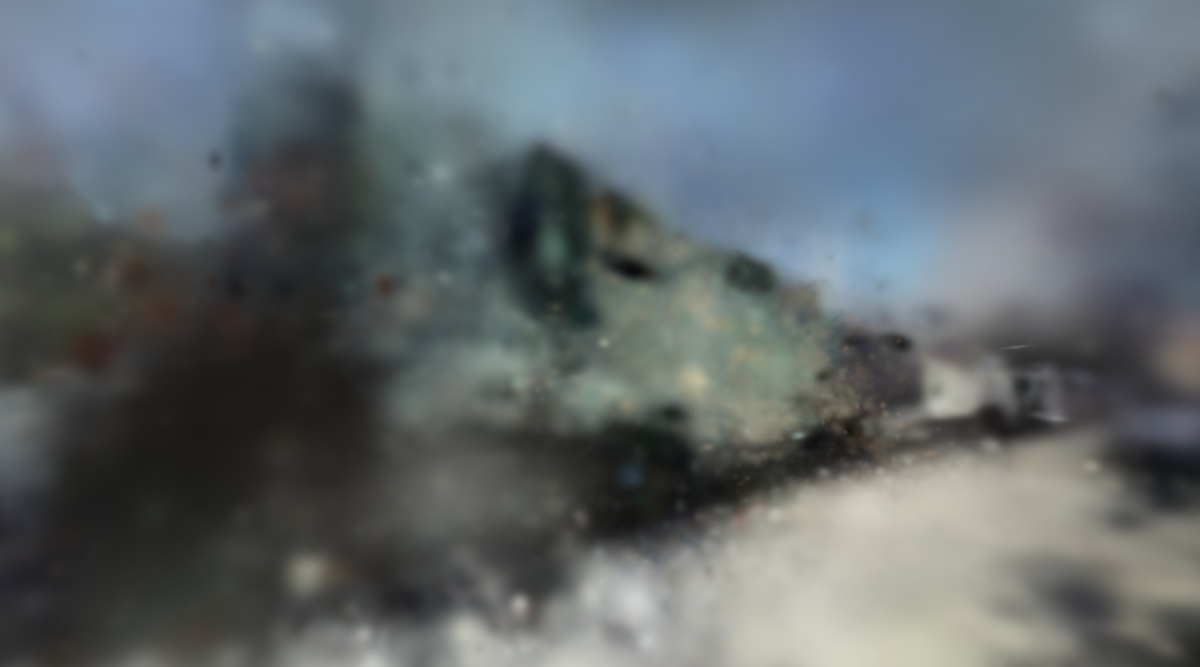}}
		\end{minipage} 
		& \begin{minipage}[b]{0.15\columnwidth}
			\centering
			\raisebox{-.2\height}{\includegraphics[width=\linewidth]{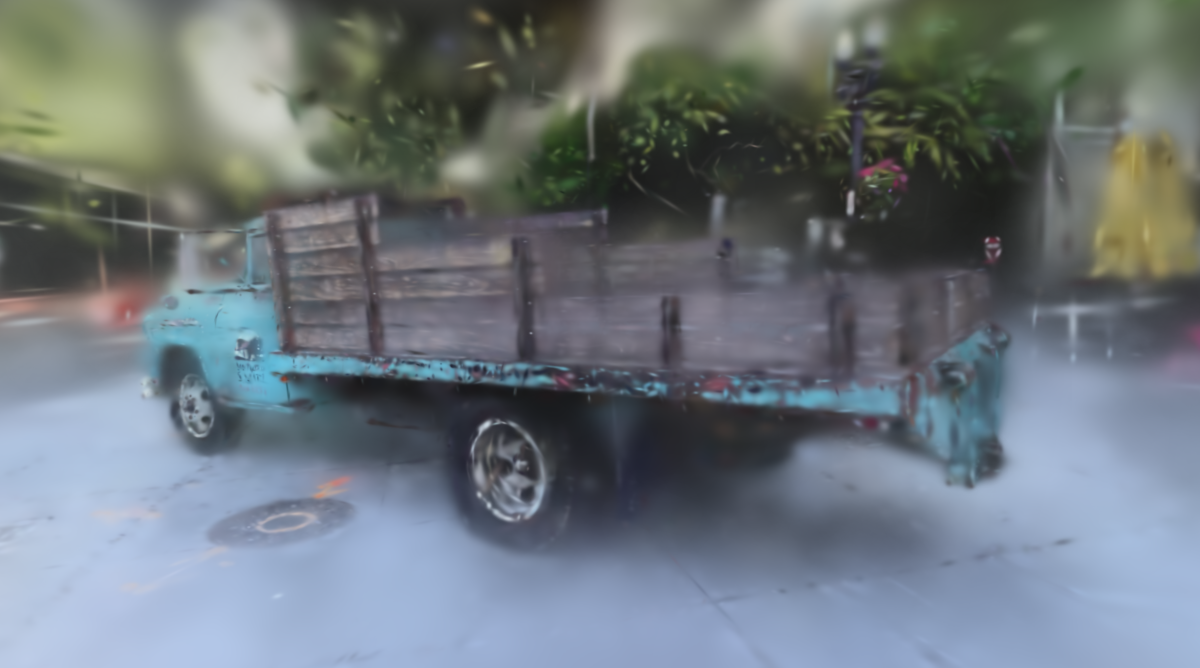}}
		\end{minipage} 
		& \begin{minipage}[b]{0.15\columnwidth}
			\centering
			\raisebox{-.2\height}{\includegraphics[width=\linewidth]{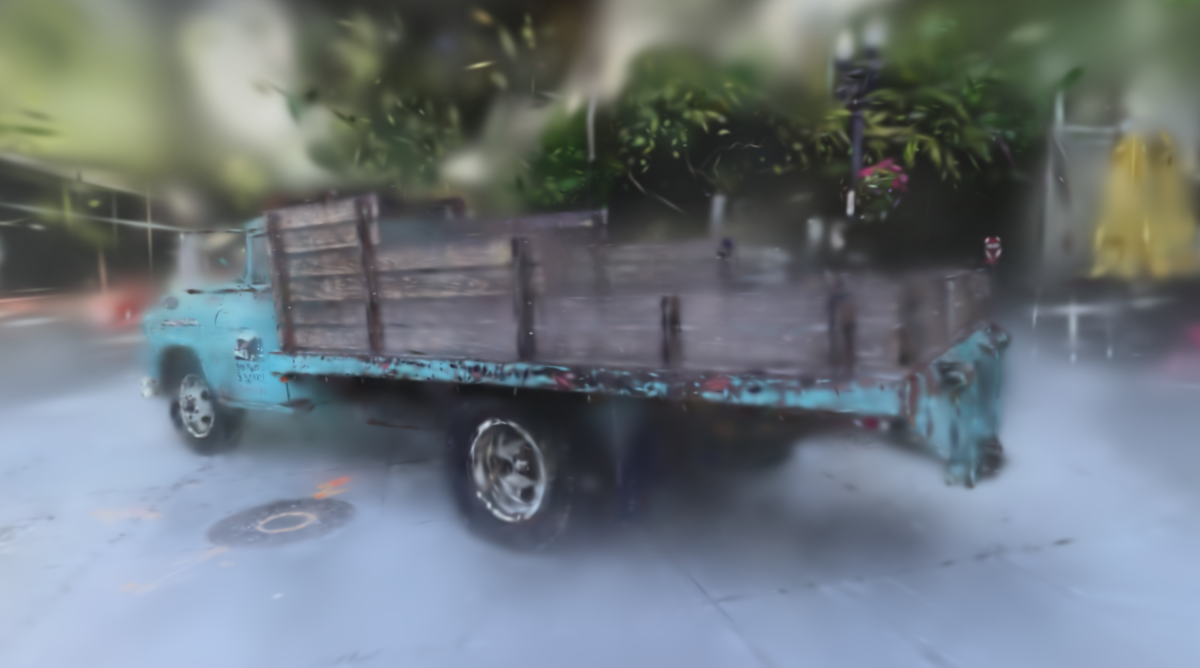}}
		\end{minipage} 
		& \begin{minipage}[b]{0.15\columnwidth}
			\centering
			\raisebox{-.2\height}{\includegraphics[width=\linewidth]{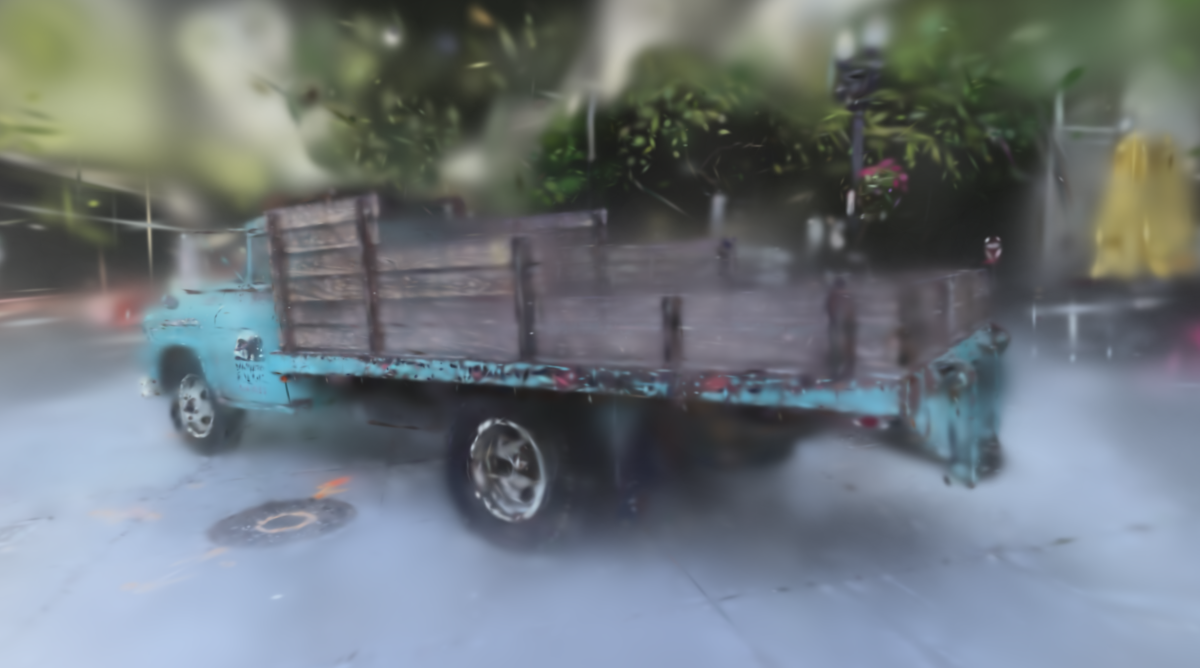}}
		\end{minipage}

		\\ \hline
		
		l1 &  0.1825
		& 0.23512
		& 0.19083
		& 0.14797
		& 0.1481
		& 0.14971
		
		
		
		\\ \hline
		
		p1 & 12.77277
		& 10.71311
		& 12.39797
		& 14.39041
		& 14.38418
		& 14.33256
		
		
		
		\\ \hline
		
		pt & 9.78125
		& 4.75625
		& 9.70625
		& 5.25938
		& 5.68438
		& 5.3
		
		\\ \hline
		
		pc & 22.48456
		& 15.13546
		& 22.45052
		& 11.53933
		& 11.43065
		& 11.62806
		
		\\ \hline
		
		st & 1000
		& 1000
		& 1000
		& 500
		& 500
		& 500
		
		\\ \hline
		
		\textbf{it} & \textbf{1e-2, 1} & \textbf{1e-3, 100} & \textbf{1e-3, 1} & \textbf{1e-3, 100} & \textbf{1e-1, 1e-9} & \textbf{1e-6, 1e-9}
		
		\\ \hline

		ri & \begin{minipage}[b]{0.15\columnwidth}
			\centering
			\raisebox{-.2\height}{\includegraphics[width=\linewidth]{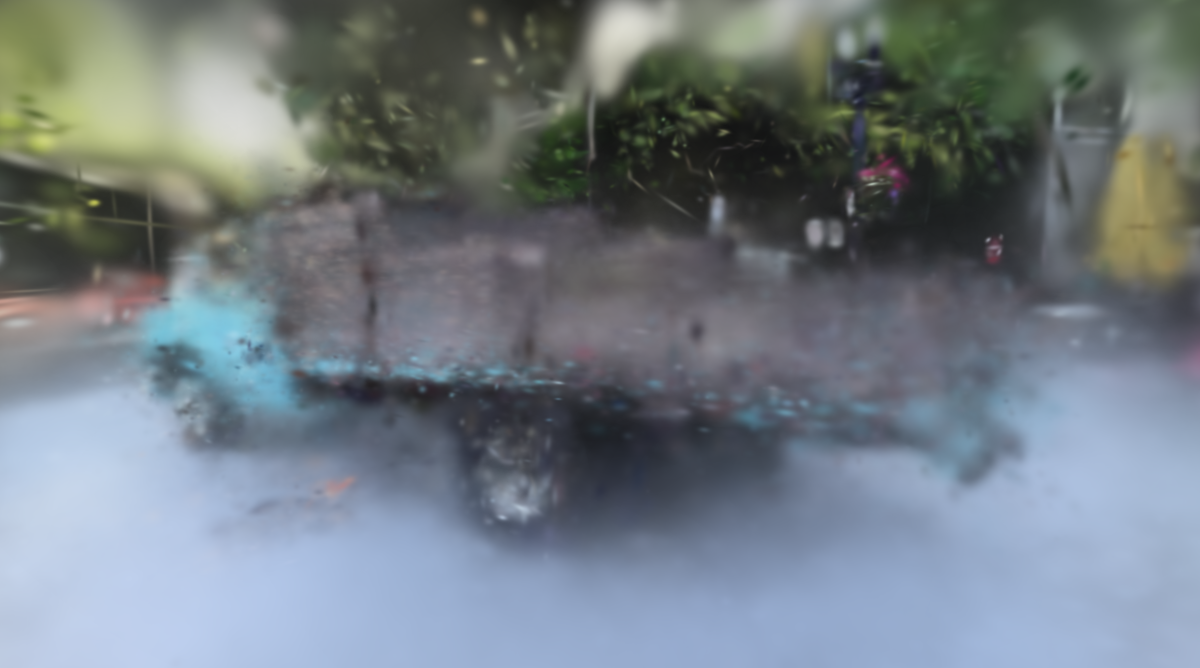}}
		\end{minipage}
		& \begin{minipage}[b]{0.15\columnwidth}
			\centering
			\raisebox{-.2\height}{\includegraphics[width=\linewidth]{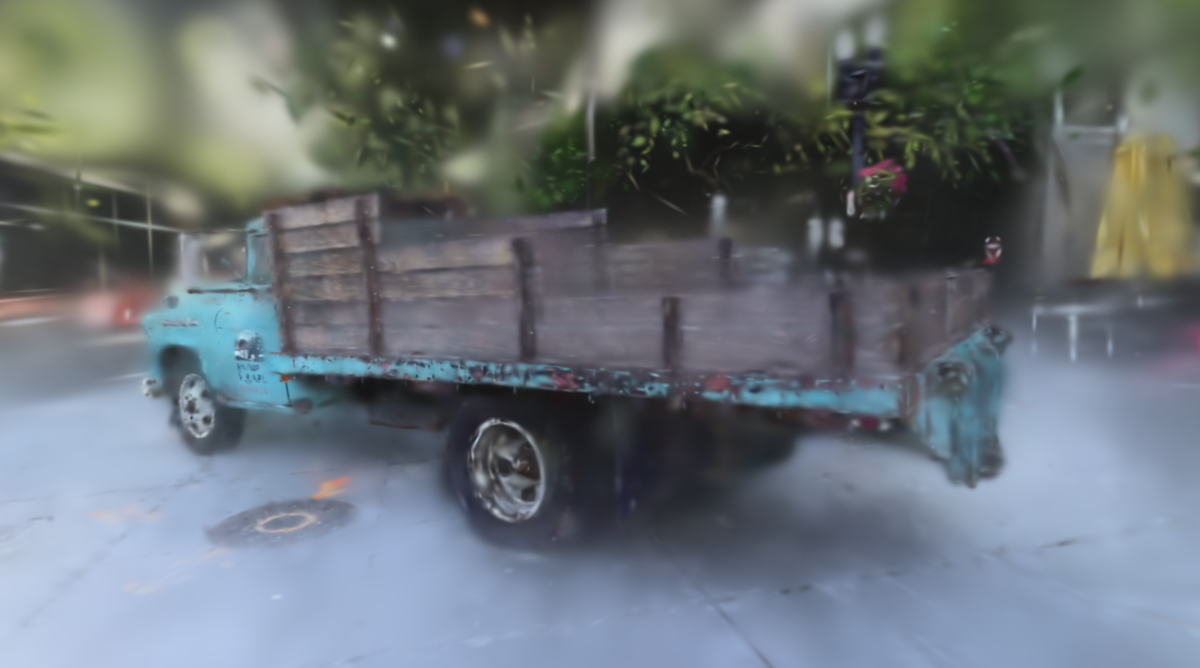}}
		\end{minipage}
		& \begin{minipage}[b]{0.15\columnwidth}
			\centering
			\raisebox{-.2\height}{\includegraphics[width=\linewidth]{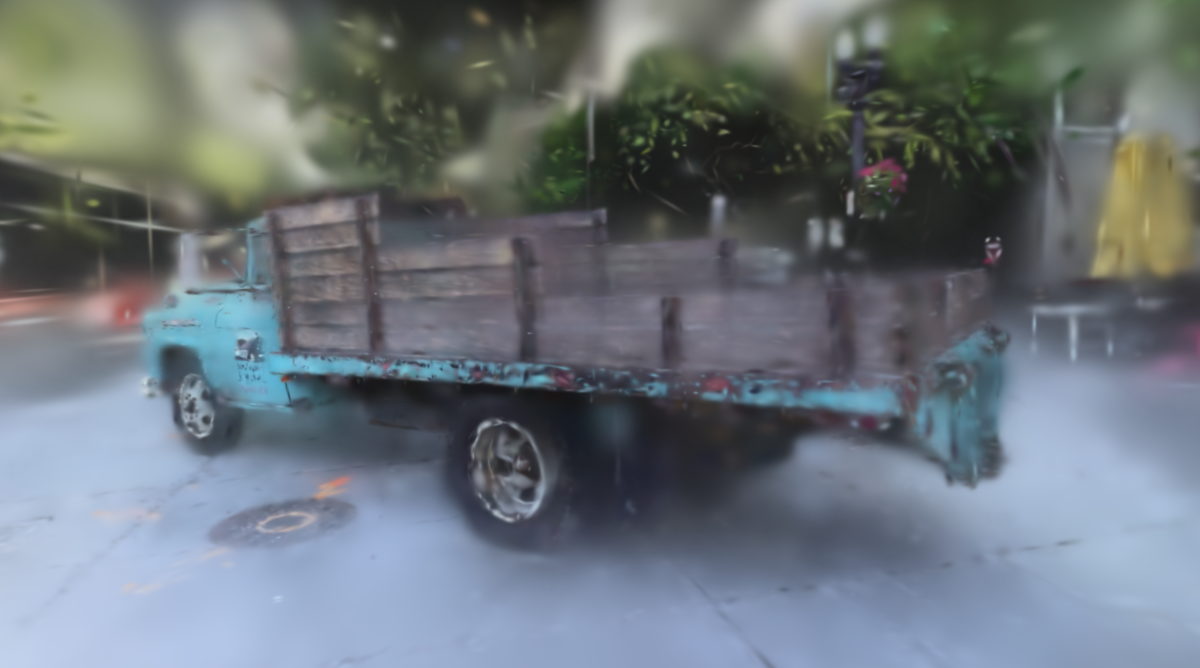}}
		\end{minipage} 
		& \begin{minipage}[b]{0.15\columnwidth}
			\centering
			\raisebox{-.2\height}{\includegraphics[width=\linewidth]{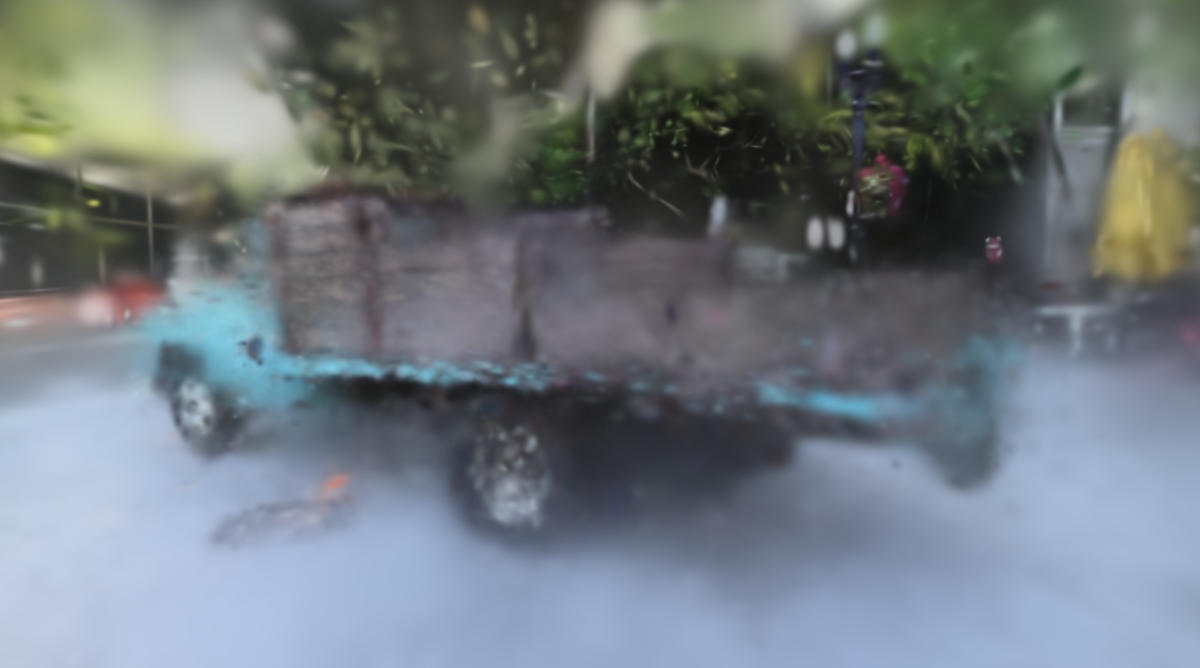}}
		\end{minipage} 
		& \begin{minipage}[b]{0.15\columnwidth}
			\centering
			\raisebox{-.2\height}{\includegraphics[width=\linewidth]{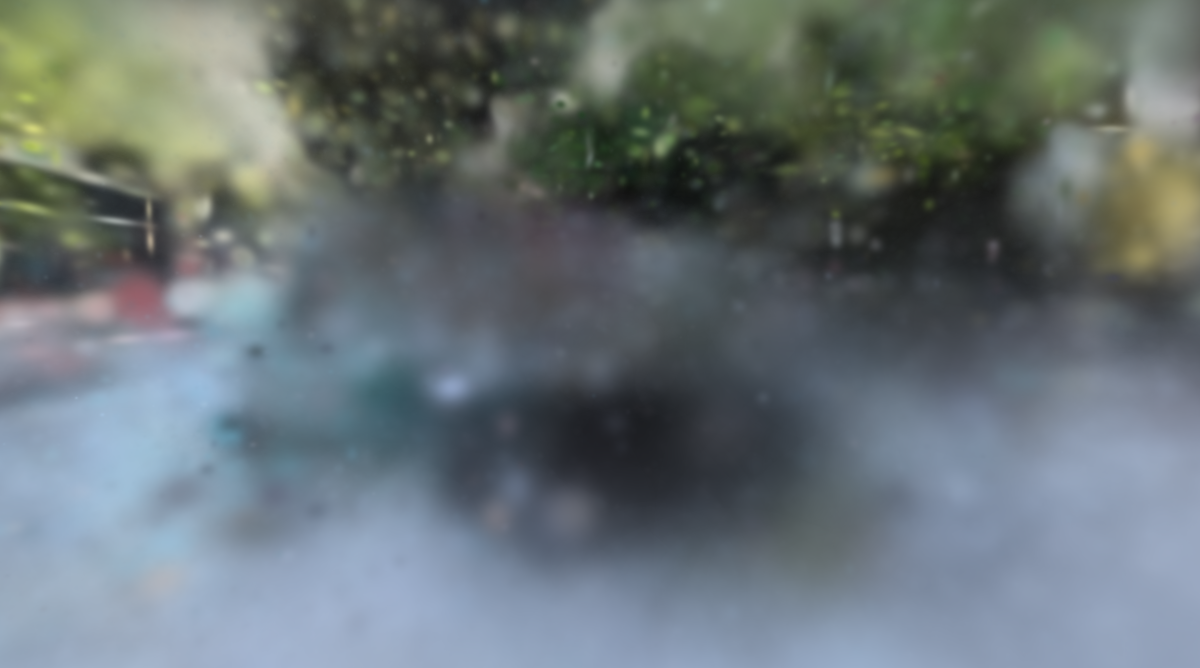}}
		\end{minipage} 
		& \begin{minipage}[b]{0.15\columnwidth}
			\centering
			\raisebox{-.2\height}{\includegraphics[width=\linewidth]{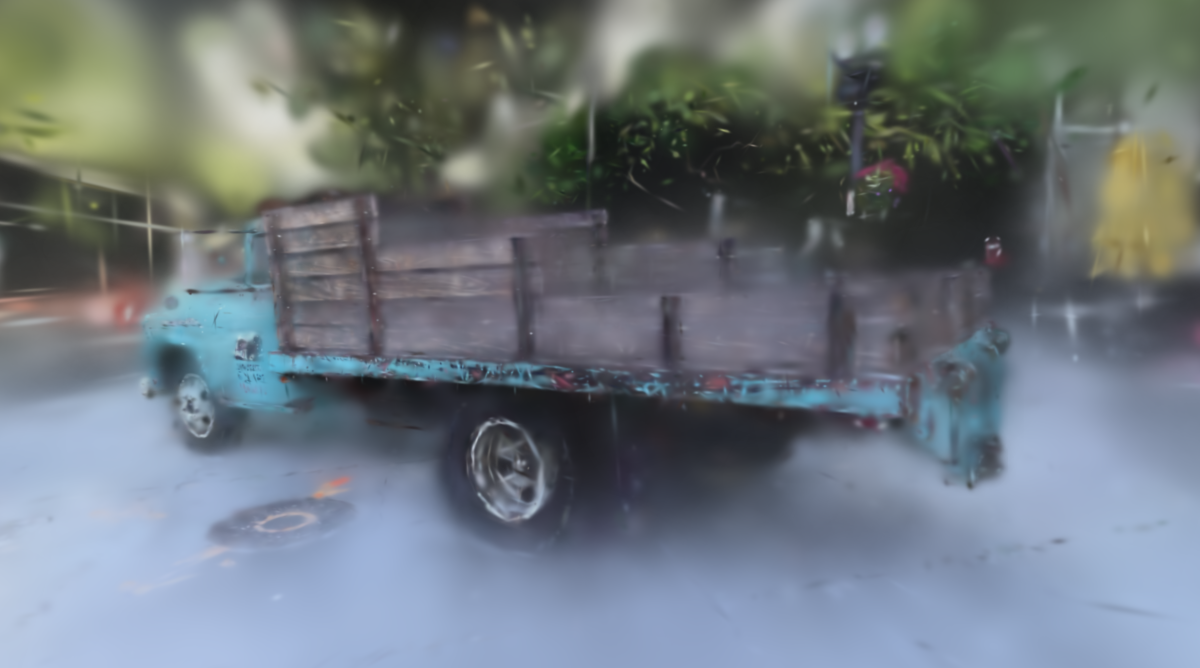}}
		\end{minipage}
		
		\\ \hline
		
		l1 &  0.15447
		& 0.14617
		& 0.14719
		& 0.14888
		& 0.15938
		& 0.1529
		
		
		
		\\ \hline
		
		p1 & 14.04064
		& 14.53539
		& 14.47097
		& 14.36201
		& 13.73271
		& 14.14589
		
		
		
		\\ \hline
		
		pt & 3.43438
		& 8.30312
		& 4.27812
		& 6.60312
		& 4.49375
		& 3.05
		
		\\ \hline
		
		pc & 11.79453
		& 21.30002
		& 11.51888
		& 12.47354
		& 10.59778
		& 11.48526
		
		\\ \hline
		
		ty & sin
		& sin
		& cos
		& cos
		& linear
		& exponential
		
		\\ \hline
	\end{tabular}
\end{table}

Not only the $xyz$ learning rate can be changed customarily, as shown in Table \ref{tb7}, each kind of the main learning rate can be also changed in each training iteration, e.g.: $scalling\ lr$, $opacity\ lr$, $rotation\ lr$ and $feature\ lr$, etc. After enough experiments, we found that changing the $scalling\ lr$ will gain an explicit change in the final rendering result and speed. $scalling\ lr$ has an optimal value that is mainly locate in the range of $1e-3, 1e-6$. The other learning rates has less impact on the improve of quality and training speed but are still useful. $feature\ lr=0$ is also interesting which means it will not change a lot the speed.

\begin{table}[h]
	\caption{ Comparison of Impacts by Other Different Types of Learning Rates.}
	\label{tb7}
	\centering
	\begin{tabular}{  c | c | c | c | c | c | c  }
		\hline
		\textbf{it} & \textbf{original} & \textbf{1e-3} & \textbf{1e-2} & \textbf{3e-1} & \textbf{5e-2} & \textbf{0}
		\\ \hline
		ri & \begin{minipage}[b]{0.15\columnwidth}
			\centering
			\raisebox{-.2\height}{\includegraphics[width=\linewidth]{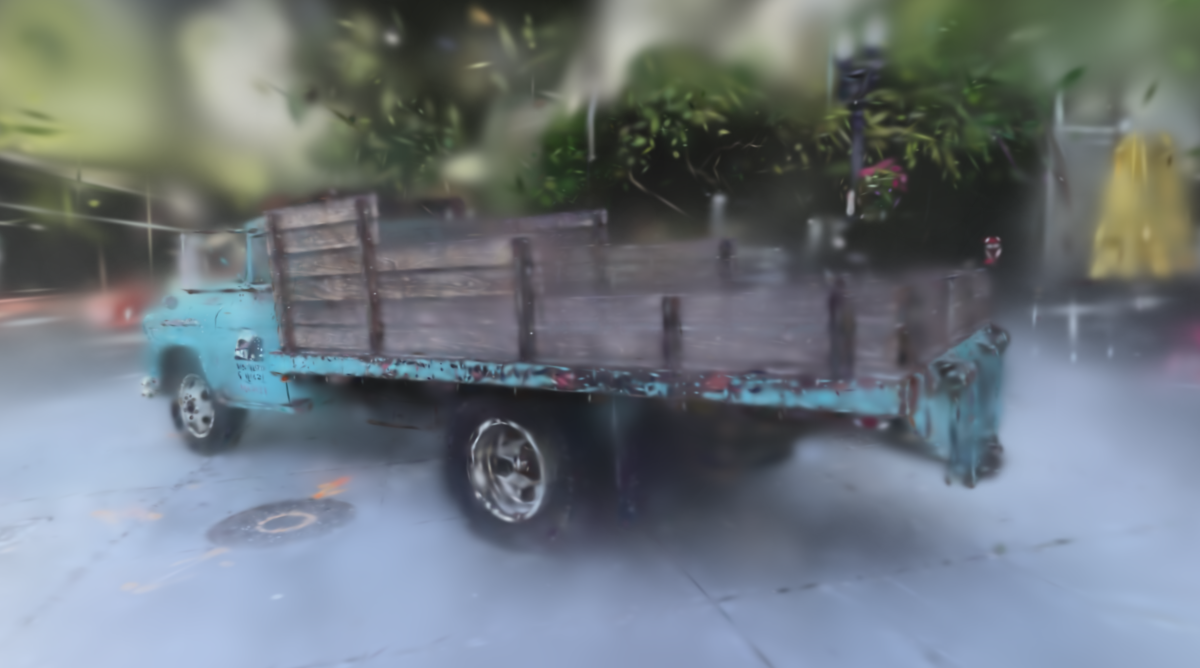}}
		\end{minipage}
		& \begin{minipage}[b]{0.15\columnwidth}
			\centering
			\raisebox{-.2\height}{\includegraphics[width=\linewidth]{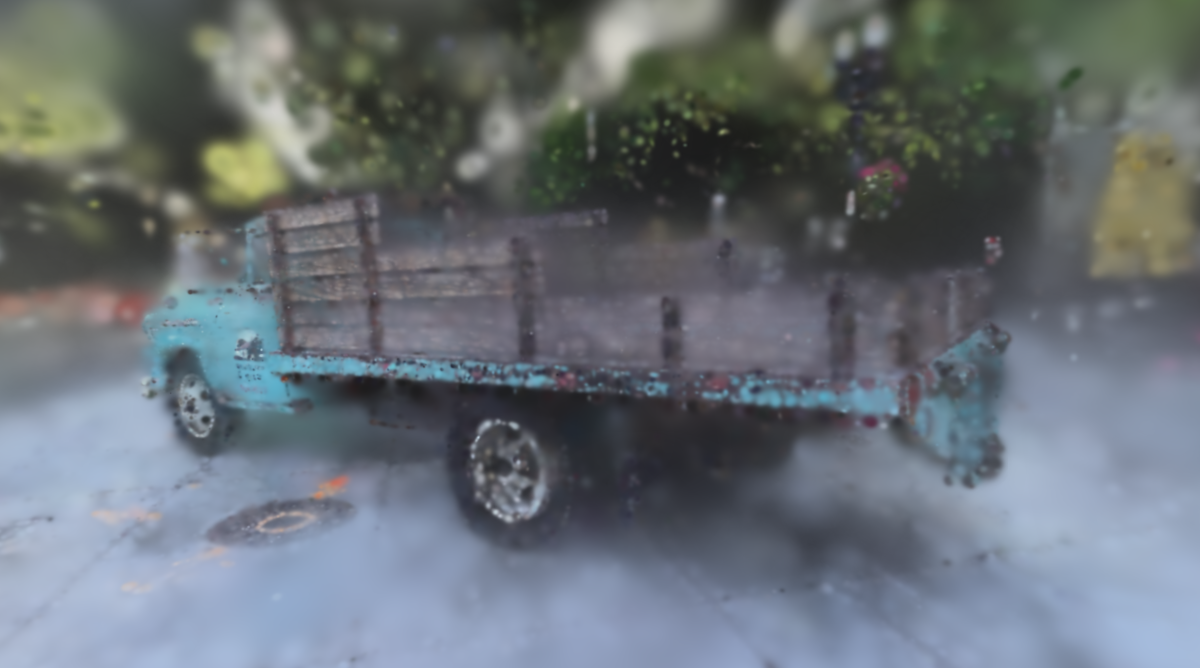}}
		\end{minipage}
		& \begin{minipage}[b]{0.15\columnwidth}
			\centering
			\raisebox{-.2\height}{\includegraphics[width=\linewidth]{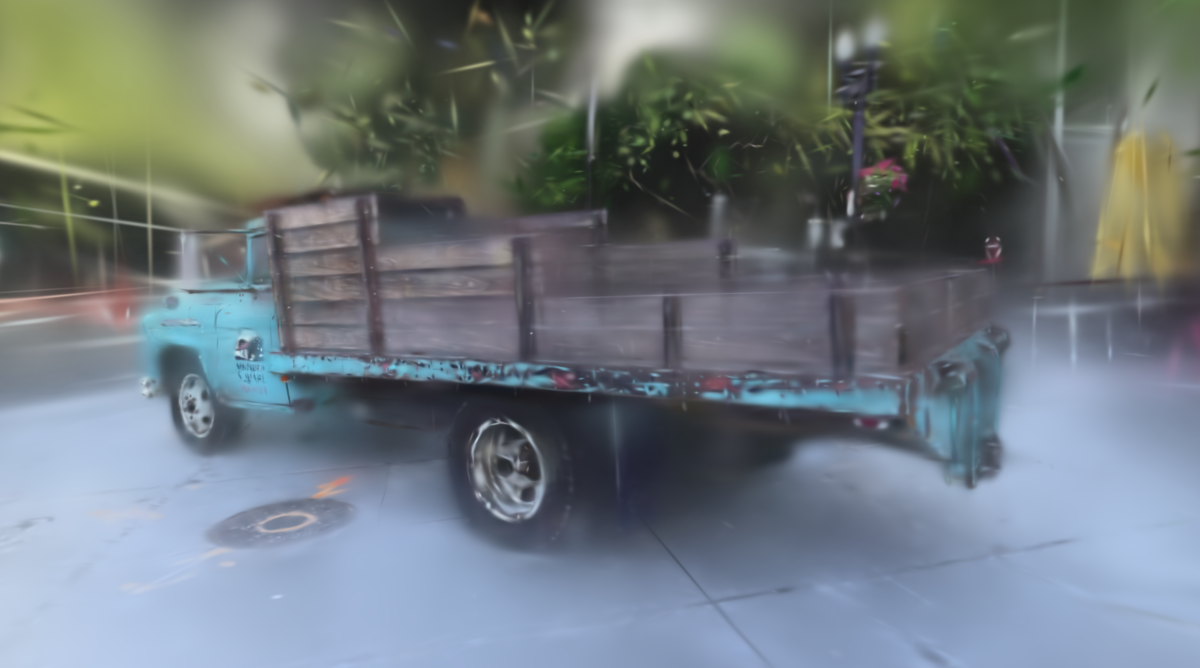}}
		\end{minipage} 
		& \begin{minipage}[b]{0.15\columnwidth}
			\centering
			\raisebox{-.2\height}{\includegraphics[width=\linewidth]{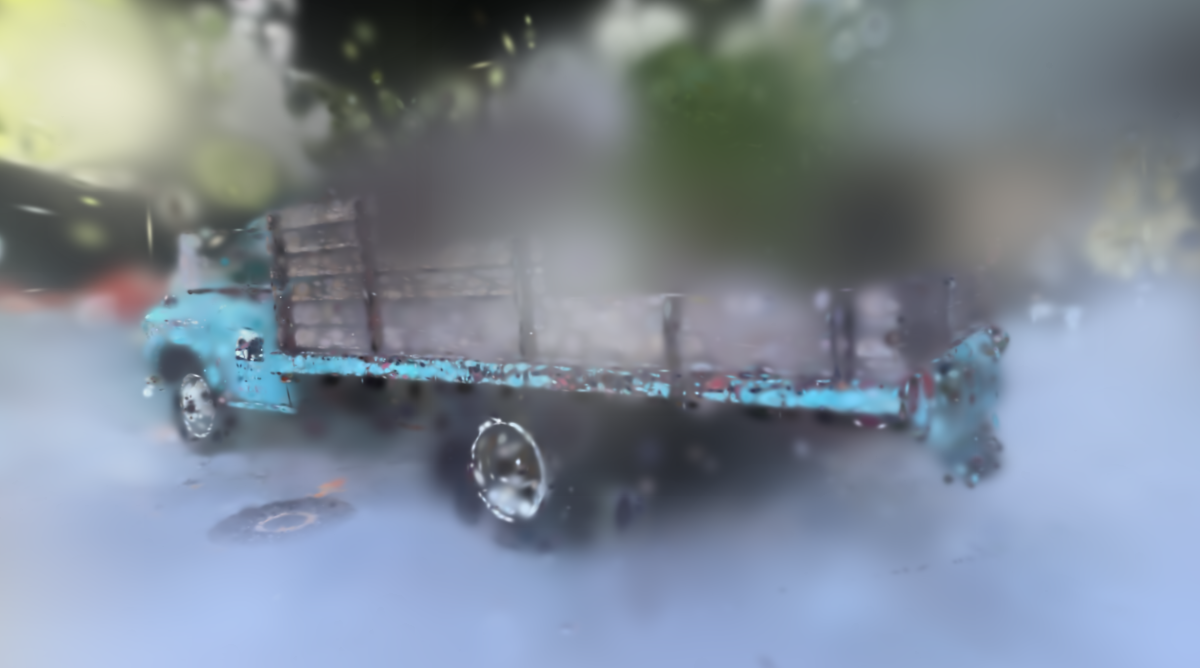}}
		\end{minipage} 
		& \begin{minipage}[b]{0.15\columnwidth}
			\centering
			\raisebox{-.2\height}{\includegraphics[width=\linewidth]{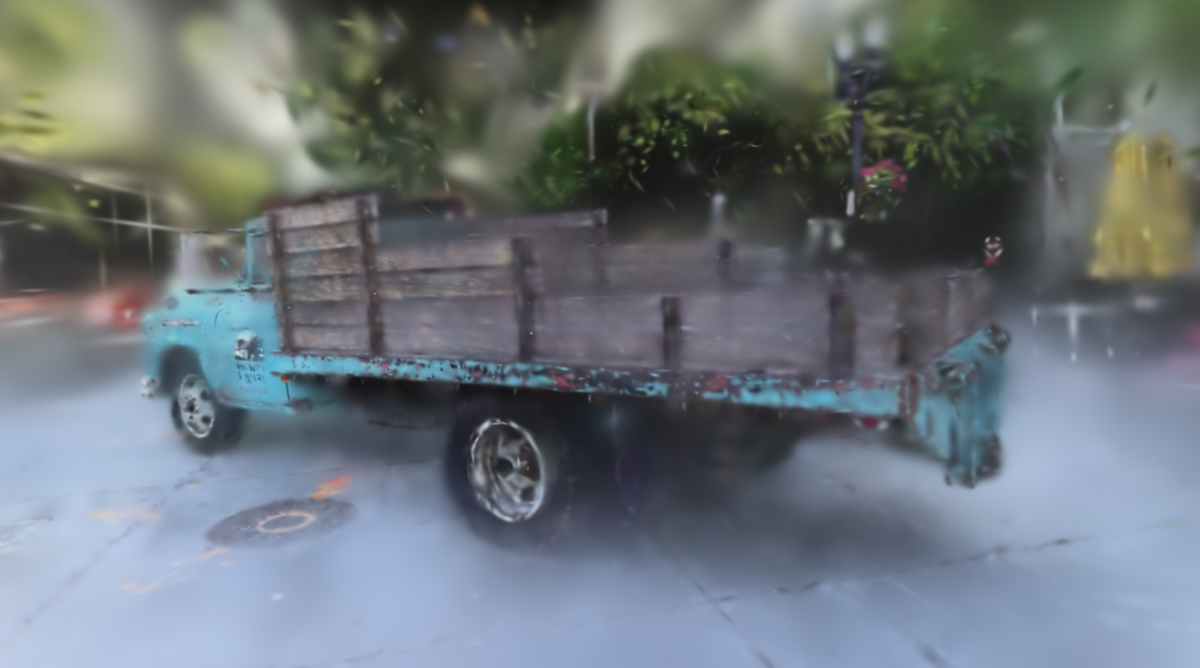}}
		\end{minipage} 
		& \begin{minipage}[b]{0.15\columnwidth}
			\centering
			\raisebox{-.2\height}{\includegraphics[width=\linewidth]{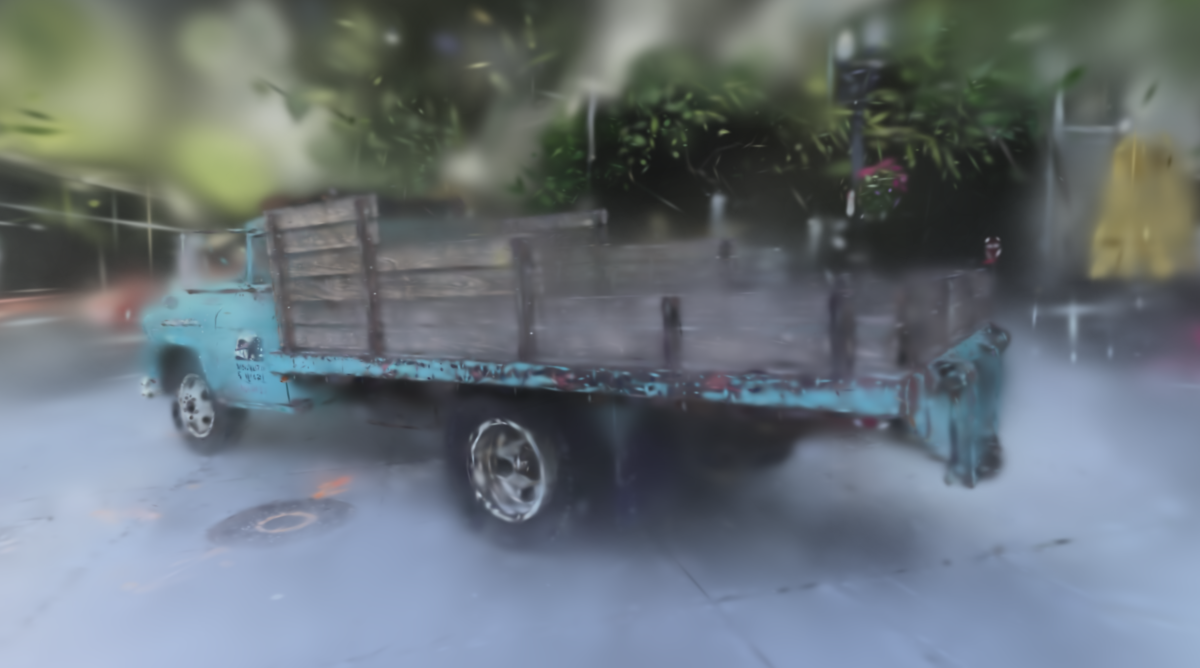}}
		\end{minipage}
		
		\\ \hline
		
		ls & \begin{minipage}[b]{0.15\columnwidth}
			\centering
			\raisebox{-.2\height}{\includegraphics[width=\linewidth]{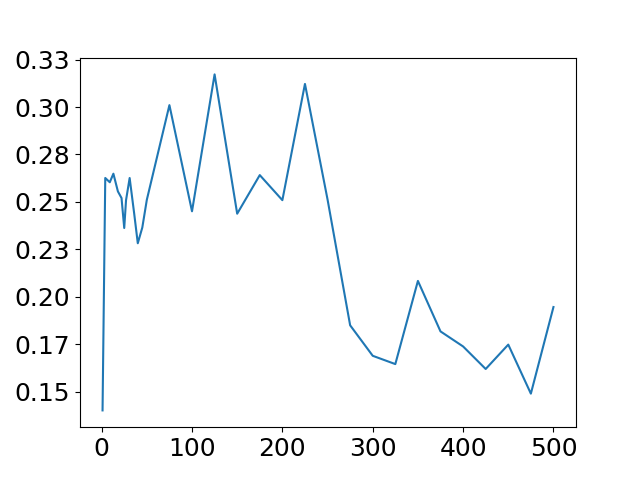}}
		\end{minipage}
		& \begin{minipage}[b]{0.15\columnwidth}
			\centering
			\raisebox{-.2\height}{\includegraphics[width=\linewidth]{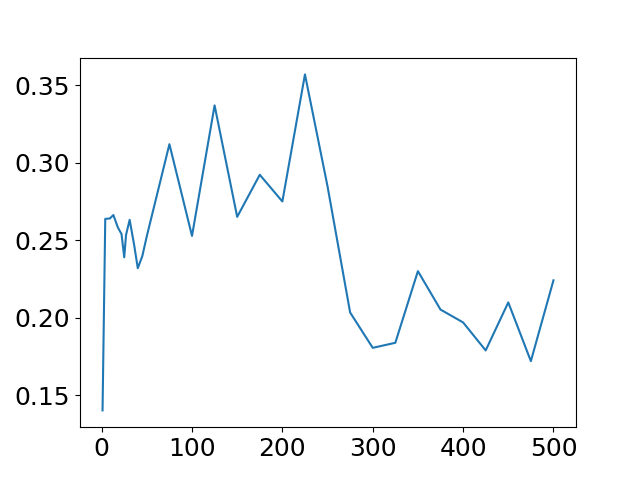}}
		\end{minipage}
		& \begin{minipage}[b]{0.15\columnwidth}
			\centering
			\raisebox{-.2\height}{\includegraphics[width=\linewidth]{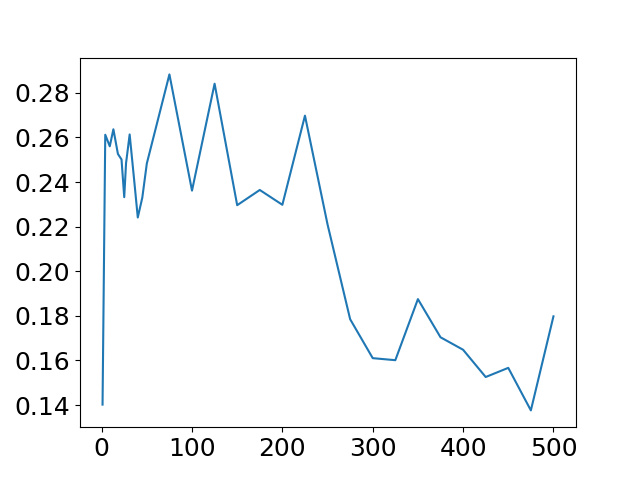}}
		\end{minipage} 
		& \begin{minipage}[b]{0.15\columnwidth}
			\centering
			\raisebox{-.2\height}{\includegraphics[width=\linewidth]{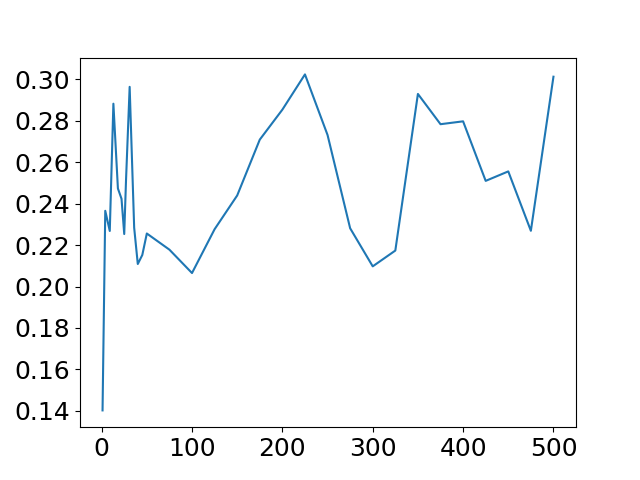}}
		\end{minipage} 
		& \begin{minipage}[b]{0.15\columnwidth}
			\centering
			\raisebox{-.2\height}{\includegraphics[width=\linewidth]{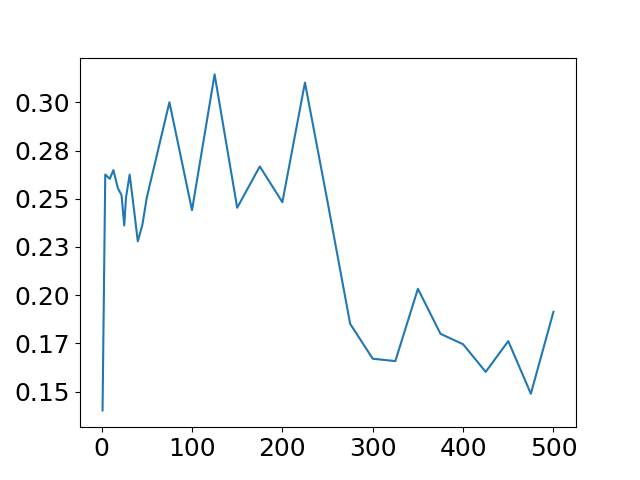}}
		\end{minipage} 
		& \begin{minipage}[b]{0.15\columnwidth}
			\centering
			\raisebox{-.2\height}{\includegraphics[width=\linewidth]{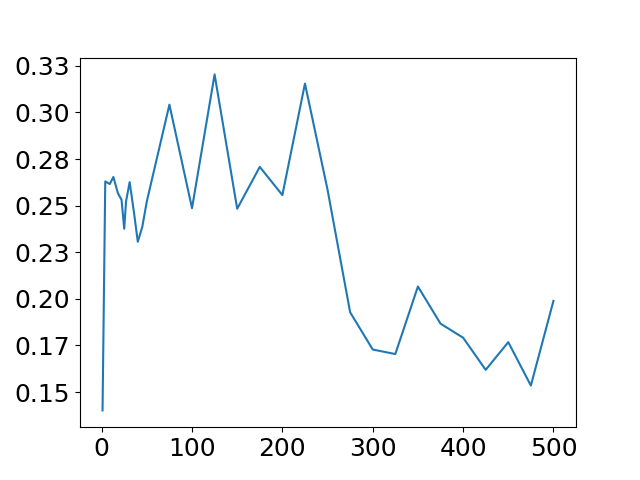}}
		\end{minipage}
		
		\\ \hline
		
		l1 &  0.14915
		& 0.17142
		& 0.13879
		& 0.22442
		& 0.15083
		& 0.15039
		
		\\ \hline
		
		pc & 12.41618
		& 10.95238
		& 14.26314
		& 8.49328
		& 12.43801
		& 12.42292
		
		\\ \hline
		
		ch & none
		& scaling lr
		& scaling lr
		& opacity lr
		& rotation lr
		& feature lr
		
		\\ \hline
	\end{tabular}
\end{table}

\subsection{Encoding and Decoding}

\begin{figure}[t]
	\centering
	\includegraphics[width=0.96\columnwidth]{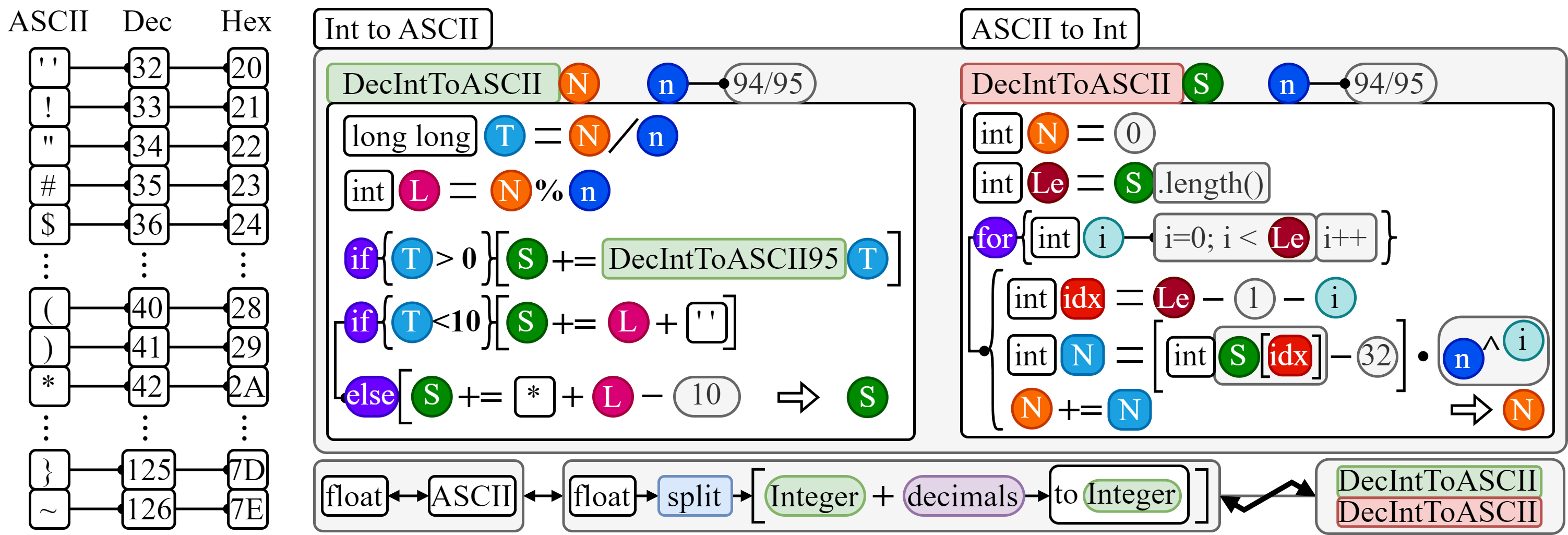}
	\caption{Running and design architecture of Anywhere.}
	\label{fig1}
\end{figure}

Data transmission is an important process that will impact the training speed. Therefore, improve the speed of data transmission will improve the training speed in some levels. To address this issue, we proposed a special integer compression in base 95 and a floating-point compression in base 94 with the ASCII encoding and decoding mechanism. The basic principle of this compression method is as shown in Figure \ref{fig1}, for the  algorithm design of integer compression in base 95, we selected No.32 to No.126 ASCII characters to make the one character encoding for the number range of $[0, 95]$, for the floating-point compression in base 94, we selected No.32 to No.125 ASCII characters to make the one character encoding for the number range of $[0, 94]$ for the integer component of the floating-point number, character ASCII(126) is used to connect the ASCII strings of the integer part and fractional part of the original floating-point number. In the processing of decoding, the integer ASCII strings will be transformed to decimal integer number directly, while the floating-point ASCII strings will be split into two part, one part is the ASCII string for the integer part of floating-point number $I$, the other part is the ASCII string for the decimal places $d$. The final floating-point value $f$ will be gained by $f=I10^{-d}$.

Due to there is not many places of the model can be optimized by using this ASCII encoding-decoding compression algorithm, with enough experiments, we found that the training speed has gained about $6\%$ or less. It is not easy to compress the color matrix into ASCII string in data transmission, the big difficulty is taking traversal of a big matrix will take a lot of process time. We use internal functions (e.g.: $torch::where$ and $torch::index\_select$) to replace the for loop to improve matrix traversal speed, but it is still not fast enough to support compression replacement. More optimization will be held in the future.

We also found that it will not make an explicit impact for the final rendering result if we compress each item of the out color matrix of the model to only one decimal places, which means using the number in the range of $[0.0, 2.0]$ with only one decimal places (only 21 alternative numbers), combining with the ASCII string compression algorithm, the color matrix data size will be reduced in a great level. Using the line break character will add one more operator in ASCII string writing and decoding. In the processing of $densify\_and\_prune$, appropriate adjustment and removing the $densify\_and\_split$ can also improve the speed quickly without explicit impact on the rendering quality.

\section{Conclusion}

The research is focusing on improving the training speed of the 3D Gaussian Splatting model. From color, learning, structure design and compression, we made various explorations and experiments to find the feasible improving methods. Our result can make some enough scientific and useful references from the relative research, which is meaningful for many research directions. We will make more exploration works on improving the 3D Gaussian Splatting in the future.

%
%
\bibliographystyle{splncs04}
\bibliography{main}
\end{document}